\pdfoutput=1

\documentclass[10pt,twocolumn,letterpaper]{article}
\usepackage[pagenumbers]{cvpr}

\usepackage{cuted} %

\usepackage{currfile} %

\usepackage{caption} %

\newcommand{\pcov}{p_{\mathrm{cov}}}  %

\usepackage[dvipsnames]{xcolor}
\usepackage{colortbl}  %
\usepackage{amsmath,amssymb,amsthm,bm,mathtools}
\newcommand{\cmark}{\ensuremath{\checkmark}}  %
\newcommand{\xmark}{\ensuremath{\times}}
\usepackage{pifont}

\usepackage{graphicx,booktabs,multirow,makecell,tabularx}
\usepackage{xspace}
\usepackage{tikz}
\usepackage{placeins} %
\usepackage{wrapfig}  %
\usetikzlibrary{arrows.meta,positioning,fit,backgrounds,calc}
\newtheorem{proposition}{Proposition}
\newtheorem{corollary}{Corollary}
\newtheorem{lemma}{Lemma}
\theoremstyle{definition}

\newtheorem{remark}{Remark}

\newcolumntype{C}{>{\centering\arraybackslash}X}  %
\newcommand{\myparagraph}[1]{\smallskip\noindent\textbf{#1}}

\newcommand{\method}{TriSegAgent\xspace}
\newcommand{\floc}{f_{\mathrm{loc}}}   %
\newcommand{\fseg}{f_{\mathrm{seg}}}   %
\definecolor{BaselineBlue}{RGB}{33,102,172}
\definecolor{OursRed}{RGB}{180,15,32}
\newcommand{\circled}[1]{\tikz[baseline=(c.base)]\node[circle,draw,inner sep=0.6pt,font=\scriptsize](c){#1};}

\makeatletter
\@ifpackageloaded{lineno}{%
  \@ifpackagelater{lineno}{2026/02/11}{%
    \RemoveFromHook{build/column/before}[lineno]%
    \AddToHook{build/column/before}{%
      \protected@write\@auxout{}{%
        \string\def\string\@LN@column{\if@firstcolumn2\else1\fi}}}%
  }{}%
}{}
\makeatother

\newcommand{\maketitlesupplonecolumn}{%
  \onecolumn
  \begin{center}
    {\Large\textbf{\thetitle}\\[0.5em]Supplementary Material}
  \end{center}
  \vspace{0.5em}%
}

\graphicspath{{./}{../}}
\newskip\arxivTFS \arxivTFS=\textfloatsep \newskip\arxivDTFS \arxivDTFS=\dbltextfloatsep
\newskip\arxivFS \arxivFS=\floatsep \newskip\arxivDFS \arxivDFS=\dblfloatsep \newskip\arxivITS \arxivITS=\intextsep
\usepackage[most]{tcolorbox} %
\usepackage[T1,OT1]{fontenc} %
\usepackage{arydshln} %

\definecolor{cvprblue}{rgb}{0.21,0.49,0.74}
\usepackage[pagebackref,breaklinks,colorlinks,allcolors=cvprblue]{hyperref}

\def\confName{CVPR}
\def\confYear{2027}

\title{TriSegAgent: Context-Efficient 3D Annotation with MLLMs via Tri-Planes}

\author{
Qingqiao Hu$^{1*}$\quad Jialin Li$^{2*}$\quad Meilong Xu$^{1}$\quad Kehan Qi$^{1}$\quad Yipeng Zhang$^{3}$\\
Lin Tian$^{4}$\quad Hongwei Bran Li$^{2}$\quad Klaus Mueller$^{1\dagger}$\quad Xiaoling Hu$^{5\dagger}$\\[0.25em]
{\normalsize $^{1}$Stony Brook University\qquad $^{2}$National University of Singapore\qquad $^{3}$University of California, Los Angeles}\\
{\normalsize $^{4}$Hong Kong Baptist University\qquad $^{5}$Massachusetts General Hospital and Harvard Medical School}\\
{\small\href{mailto:qingqiao.hu@stonybrook.edu}{\texttt{qingqiao.hu@stonybrook.edu}}}
}
\newcommand{\arxivfootnotes}{\begingroup\renewcommand{\thefootnote}{}\footnotetext{$^{*}$Equal contribution.\quad $^{\dagger}$Co-corresponding authors.}\endgroup}

\newif\ifwithsupp\withsupptrue
\DeclareRobustCommand{\supplref}[1]{\cref{#1}}
\newcommand{\mainref}[1]{\cref{#1}}

\begin{document}
\maketitle
\begin{strip}
\centering
\includegraphics[width=\textwidth]{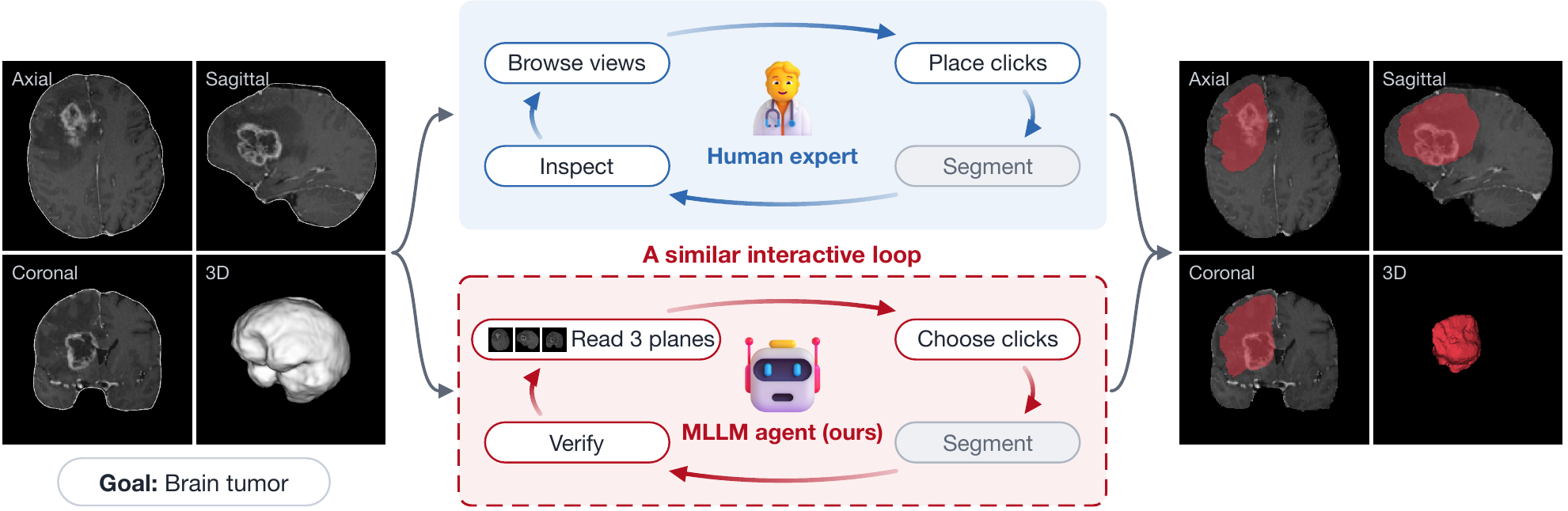}
\captionsetup{hypcap=false}
\captionof{figure}{\textbf{An MLLM agent annotates a 3D volume in an interactive loop.}
\emph{Middle-Top:} an expert annotates the target with an interactive segmentation model in a loop of browsing views, placing clicks, segmenting, and inspecting, from the input (\emph{left}) to a satisfactory prediction (\emph{right}).
\emph{Middle-Bottom:} in \method, an MLLM agent runs a similar loop with the same model: it looks at tri-planes, chooses clicks, and verifies the 3D prediction in several views before accepting it.}
\captionsetup{hypcap=true}
\label{fig:teaser}%
\end{strip}

\ifdefined\arxivfootnotes\arxivfootnotes\fi
\begin{abstract}
Interactive segmentation models turn a few clicks into a segmentation prediction across domains, so annotators only decide where the prediction still needs correction. Annotation thus becomes a loop of inspecting the prediction and clicking again. Each step of this loop is a visual judgment, which suits general-purpose multimodal large language model (MLLM) agents. In this work, we focus on letting these agents run this loop on 3D volumes. However, the MLLM in such an agent takes only 2D images as input, so the agent must observe the 3D volume in 2D. Since the loop calls the MLLM many times, this observation must be token-compact enough to keep every call cheap, yet informative enough to show error regions in the 3D prediction. We meet both needs with tri-planes, the three orthogonal planes that cross at one point in the volume. We show, both theoretically and empirically, that tri-planes are context-efficient yet keep sufficient detail to show error regions in the 3D prediction. Building on tri-planes, we introduce \method, an MLLM agent that refines the 3D prediction iteratively. Within the agent, the MLLM only judges the prediction on tri-planes, while a visual harness turns each judgment into a 3D action in code. To make the loop reliable, the harness places each click at an exact voxel in the volume and checks whether to accept the 3D prediction. To evaluate the agent, we build a diverse benchmark of 690 regions of interest (ROIs) from 35 public medical datasets across three imaging modalities. With frontier MLLMs, \method reaches an overall Dice about 10 points above the strongest baseline, and its lead is even larger on targets below 3\,cm. With flash-level MLLMs, it outperforms a recent 3D segmentation agent built on a frontier MLLM.
\end{abstract}
\newpage
\section{Introduction}
\label{sec:intro}

Interactive segmentation tools, which turn a few clicks or boxes into a segmentation prediction, have been widely used for annotation across domains, from natural images~\cite{sofiiuk2022ritm,kirillov2023sam} and microscopy~\cite{archit2025microsam} to 3D medical volumes~\cite{ma2025medsam2,isensee2025nninteractive,devente2026slicernninteractive}.
Since these tools draw most of the prediction, annotators only decide where it still needs correction~\cite{marinov2024interactive}.
Annotation thus becomes a loop of inspecting the prediction and clicking again~\cite{olabarriaga2001interaction}.
On a 3D volume (\cref{fig:teaser}, middle-top), guided by prior knowledge of where the target lies, an expert browses orthogonal planes of the volume, places clicks on one plane, lets the model segment, and inspects the result in different planes until the prediction is satisfactory~\cite{ma2025medsam2,isensee2025nninteractive}.
Each decision in this loop needs only a local judgment with a clear criterion: whether the prediction covers the target in the current view.
This perception--decision loop maps naturally to a multimodal large language model (MLLM) agent that looks at an image, acts through tools, and judges the outcome~\cite{yao2023react,hu2024sketchpad}. The planes displayed are the agent's observations; the clicks are its tool calls; and the prediction overlaid on the image is the feedback for its next decision.
This work therefore focuses on enabling general-purpose MLLM agents to annotate 3D volumes through such an interactive loop.
Given a volume and a text goal, the agent works with an interactive segmentation model over many steps and then returns a 3D prediction (\cref{fig:teaser}).

Running this loop on 3D volumes, however, exposes a gap between the data and the agent: the data are 3D voxel grids, whereas general-purpose MLLM agents take only 2D images as input.
In 2D, one image with the prediction overlaid makes each decision fully observable at little token cost~\cite{zhu2025segagent,liu2026medsamagent}.
Existing methods bridge this gap in three ways, none of which suits a long agent loop.
The first trains a native 3D model~\cite{bai2024m3d,wu2025radfm} at a high cost.
Since 3D training data are limited, such a model transfers poorly to unseen settings.
The second feeds a general-purpose MLLM the whole volume as a sequence of slices, like the frames of a video~\cite{bai2025qwen25vl,comanici2025gemini25,golden2026medgemma15}.
Each call then carries hundreds of images, although the current decision needs only a few of them.
Such long inputs make the token cost explode over the many calls of the loop and lower the accuracy of each call~\cite{levy2024sametask,hsieh2024ruler,wang2025mmneedle}.
The third shows only a few slices or projections at a time~\cite{wang2024votemi,callaghan2026alignment}, which keeps each call affordable but leaves the agent blind to error regions elsewhere.
The latter two ways expose a fundamental dilemma between a limited context budget and spatial observability, and the key challenge is to design an observation that is token-compact yet informative enough to show error regions in the 3D prediction.
\begin{figure}[t]
\centering
\includegraphics[width=\columnwidth]{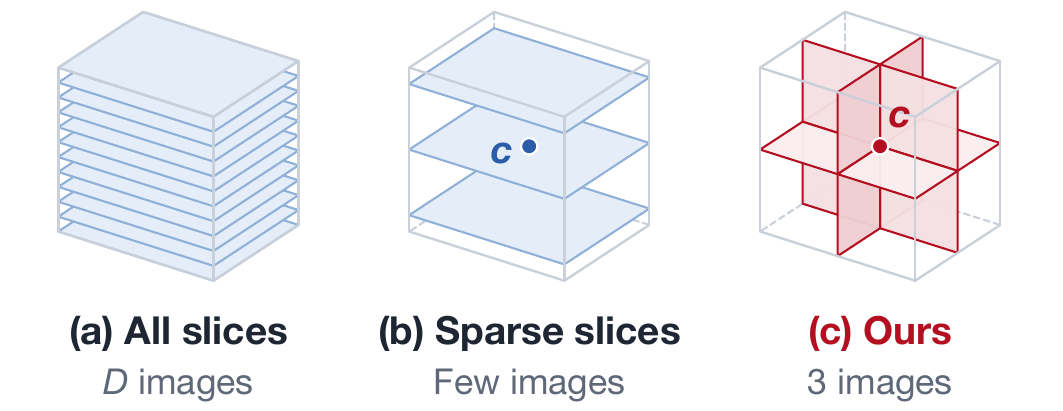}
\caption{\textbf{Visualization of different 3D inputs.}
At the same detail, a dense slice stack costs one image per slice ($D$ slices in the volume), whereas tri-planes need only three images; at the same pixel budget, a sparse stack leaves gaps between its slices.}
\label{fig:triview}
\end{figure}

We address this challenge with synchronized \emph{tri-planes}, three orthogonal planes that cross at one point (\cref{fig:triview}c).
Inspired by how engineering drawings~\cite{iso5456part2}, medical image viewers~\cite{yushkevich2006itksnap,fedorov2012slicer}, and connectomics tools~\cite{boergens2017webknossos} display or show 3D data, we make tri-planes the observation of the agent, since they are extremely context efficient yet keep sufficient detail.
To show the 3D context around the crossing point in the same detail, a slice stack needs every slice, whereas tri-planes need only three images: for a $512^{3}$ volume, Qwen2.5-VL~\cite{bai2025qwen25vl} uses about 1k tokens for one set of tri-planes, two orders of magnitude fewer than the 166k of the full slice stack.
Every MLLM call is therefore cheap, and the whole annotation loop remains affordable.
Tri-planes also keep sufficient detail: with the pixels of only three images, they cut the volume along all three axes, so an error region that slips between sparse parallel slices is still likely to cross one of the planes.
We prove that this advantage holds in the worst case, so a few sets of tri-planes are likely to show the error regions in a 3D prediction.
\begin{figure*}[t]
\centering
\includegraphics[width=0.98\textwidth]{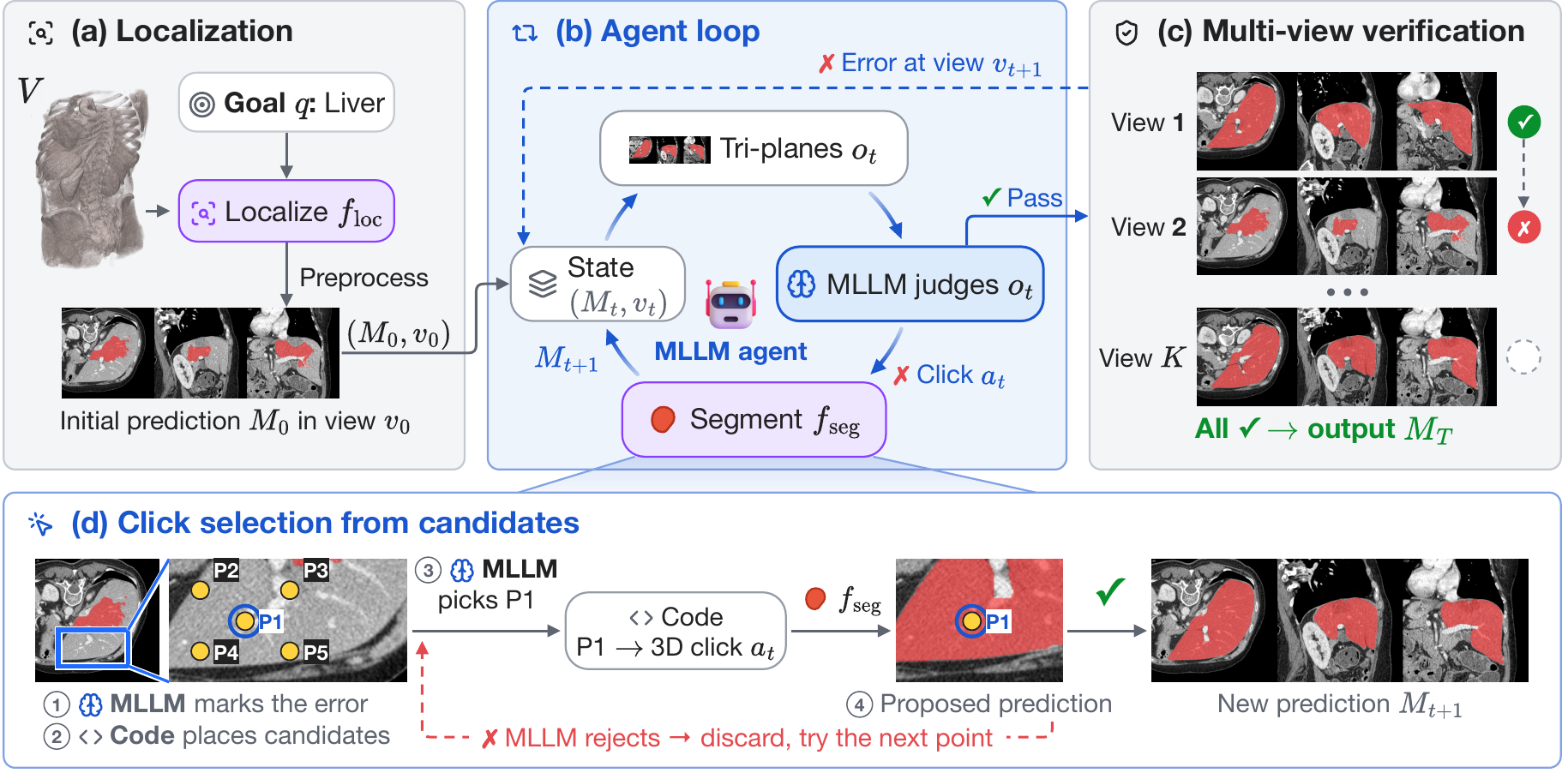}
\caption{\textbf{Overview of \method.}
Given a volume $V$ and a text goal $q$, the agent refines a 3D prediction until it passes multi-view verification.
\textbf{(a)}~The localization model $\floc$ locates the target, and preprocessing turns its prior into the initial prediction $M_0$ in view $v_0$.
\textbf{(b)}~The MLLM judges the tri-planes $o_t$ of view $v_t$ with $M_t$ as an overlay: if it reports an error, the interactive segmentation model $\fseg$ turns a click $a_t$ into $M_{t+1}$; otherwise, verification starts.
\textbf{(c)}~The MLLM judges up to $K$ verification views in turn; the first view with an error becomes $v_{t+1}$, and the agent outputs $M_T$ once all views pass.
\textbf{(d)}~Click selection from candidates: \protect\circled{1}~the MLLM marks the error with a box; \protect\circled{2}~the harness places candidates in it on the correct side of the prediction boundary; \protect\circled{3}~the MLLM picks one, which the harness maps to a 3D click $a_t$; \protect\circled{4}~$\fseg$ proposes a prediction, which becomes $M_{t+1}$ only if the MLLM confirms it; otherwise, the MLLM picks another point.}
\label{fig:workflow}
\end{figure*}

Building on such context-efficient layout, we introduce \method (\cref{fig:workflow}), an MLLM agent that pairs a general-purpose MLLM with a visual harness to refine a 3D prediction through a long interactive loop.
The harness does most of the volumetric computation, such as rendering the tri-planes, or moving the \emph{view}, the cubic sub-region that tri-planes cut through, so that the MLLM can focus on visual judgments.
Before the loop starts, the agent needs to know roughly where the target lies.
Since MLLMs cannot localize the target reliably~\cite{rahmanzadehgervi2024blind,yang2023setofmark,li2025screenspotpro}, a localization model provides a \emph{localization prior}, which helps to set the first tri-planes and the initial prediction.
At each step within the loop, the MLLM judges whether the prediction overlay on the tri-planes shows an error, and the harness turns this judgment into a 3D action, such as placing a click in the volume.

To make the loop reliable, we design mechanisms around two key aspects: where to click and whether to accept the prediction.
First, MLLMs judge images well but are less reliable at giving exact pixel positions~\cite{rahmanzadehgervi2024blind,yang2023setofmark,li2025screenspotpro}, so a click that the MLLM places itself can land a few pixels off, on the wrong side of the prediction boundary.
We therefore let the MLLM only mark the error and pick one of a few candidate points that the harness computes in code, so every click lands on an exact voxel on the correct side of the boundary.
Second, the tri-planes show only the area around their center, so a 3D prediction can look correct in the current view but still miss a distant part of the target.
The agent therefore accepts a prediction only after it also passes in several other views spread across the target, and otherwise corrects the errors that these views show.
We call this check \emph{multi-view verification}: each correction needs only a local view, while the decision to accept considers the whole prediction.
Under idealized assumptions, we prove that the chance of accepting a wrong prediction stays below a limit that falls geometrically with the number of views per check.

Since many annotated 3D volumes are publicly available in medical imaging, we build a \emph{diverse} benchmark of 690 regions of interest (ROIs) from 35 public datasets, covering three imaging modalities.
The ROIs include many small targets below 3\,cm and lesion types that the interactive segmentation model was never trained on, such as gallstones and pituitary tumors.
With frontier MLLMs~\cite{anthropic2026opus55,openai2026gpt6astra}, \method outperforms every baseline in overall Dice, by about 10 points over the strongest one and by more on small targets.
Even with flash-level MLLMs~\cite{google2026gemini38flash,deepseek2026v41flash,qiu2026qwen38next}, it outperforms nnInteractive Agent~\cite{isensee2026nninteractiveagent} running a frontier model~\cite{openai2026gpt61sol}.
With Gemini-3.8-Flash, tri-planes need 9.8$\times$ fewer input tokens per MLLM call than a stack of 16 parallel slices.
Multi-view verification makes the agent accept fewer wrong predictions.
On the localization model's unseen datasets, \method outperforms this model with every MLLM that we test.

Our contributions are as follows:
\begin{itemize}[topsep=0pt,partopsep=0pt,parsep=0pt,itemsep=0pt]
  \item We make tri-planes the observation of the agent and show that they are context efficient, with sufficient detail to show an error region.
  \item Building on tri-planes, we propose \method, an MLLM agent for 3D volume annotation whose visual harness handles most geometric computation and the multi-view verification, so that the MLLM makes only visual judgments.
  \item On 690 ROIs from 35 datasets, \method outperforms every baseline with frontier MLLMs, and with flash-level MLLMs it still outperforms nnInteractive Agent running a frontier model.
\end{itemize}
\section{Related Work}
\label{sec:related}

\myparagraph{Context management for MLLMs on 3D volumes.}
A video, like a 3D volume, is a long sequence of 2D images.
General-purpose MLLMs read it as frames sampled across the whole video~\cite{lin2024videollava,li2025llavaonevision,wang2024qwen2vl,georgiev2024gemini15} or within the time windows that a query needs~\cite{ye2026videoresearcher}.
Since visual tokens grow with the number of frames, many methods compress frame tokens inside the model~\cite{li2024llamavid,song2024moviechat,shu2025videoxl,shen2025longvu}, and accuracy still drops on longer videos~\cite{fu2025videomme,zhou2025mlvu}.
MLLMs read 3D volumes in the same ways: as a sequence of slices~\cite{bai2025qwen25vl,comanici2025gemini25,yang2024medgemini,golden2026medgemma15}, with tokens pruned inside the model~\cite{fang2026photon,liu2026medpruner}, or as a few slices, projections, or orthogonal planes through a fixed point~\cite{wang2024votemi,trinh2026spatialmed,callaghan2026alignment}.
In an annotation loop, the former makes every call costly, whereas the latter misses error regions elsewhere.
\method instead moves tri-planes across the volume as the observation of each step, which are context efficient yet keep sufficient detail.

\myparagraph{Agents for interactive segmentation.}
Interactive segmentation models predict a target mask from clicks or boxes~\cite{kirillov2023sam,ma2024medsam,ma2025medsam2,wang2023sammed3d,isensee2025nninteractive}.
Like humans, MLLMs are able to interact with these models.
In 2D, SegAgent~\cite{zhu2025segagent}, MedSAM-Agent~\cite{liu2026medsamagent}, and IBISAgent~\cite{jiang2026ibisagent} train MLLMs to output click or box coordinates, and SAM~3 agents refine text prompts~\cite{carion2026sam3,liu2025medsam3}.
Medical agents also call segmentation and analysis tools~\cite{hoopes2024voxelprompt,fallahpour2025medrax,li2024mmedagent,wang2026medagent3d,nath2025vilam3,roschewitz2026radagent,shen2026medopenclaw}, but they analyze images rather than refine a 3D prediction.
In 3D, nnInteractive Agent~\cite{isensee2026nninteractiveagent} lets coding agents operate nnInteractive by reading voxel coordinates off rendered views.
In \method, the MLLM only judges tri-planes, while the harness computes every click position in code, and the agent accepts a 3D prediction only after checking it in several views.
More is discussed in \supplref{sec:suppl_related}.

\myparagraph{End-to-end 3D models.}
Text-prompted segmentation models directly map a 3D volume and a text prompt to a segmentation mask~\cite{liu2023clipdriven,zhao2025sat,du2024segvol,rokuss2026voxtell,xin2026text3dsam,jiang2026medicalsam3}, while 3D MLLMs similarly rely on specialized volumetric encoders~\cite{bai2024m3d,wu2025radfm,hamamci2026ctclip}.
Although effective on in-domain targets, these approaches require additional fine-tuning to adapt to unseen domains, and lack the ability to iteratively refine their outputs.
Instead, \method iteratively verifies and corrects its predictions in a loop using a general-purpose MLLM and an open-set interactive segmentation model~\cite{kirillov2023sam,isensee2025nninteractive}, leveraging the superior transferability of frontier MLLMs to unseen settings over task-specific models~\cite{callaghan2026alignment}.

\section{Methodology}
\label{sec:method}
\subsection{Preliminary}
\label{sec:prelim}
\myparagraph{Viewing properties of 3D volumes.}
A 3D volume $V$ is defined on a regular voxel grid with spacing $h$.
To observe such a volume, image viewers~\cite{yushkevich2006itksnap,fedorov2012slicer,wolf2005mitk} often display 2D planes through a chosen point.
They render these planes from a view $v=(c,F)$, a cubic sub-region of $V$ with center $c$ and side length $F$.
Each plane is resampled onto an $n\times n$ pixel grid with pixel spacing $F/n$, so at a fixed grid size, $F$ sets both the visible region and the finest visible detail (see \supplref{sec:suppl_aniso} for anisotropic spacing).
Because a single view cannot provide both wide context and fine detail, where and at what scale to look must be balanced.

\myparagraph{Problem formulation.}
Given a 3D volume $V$ and a text goal $q$, the task is to predict the binary 3D mask of the target structure, whose ground truth $M^\ast$ is unknown to the agent.
We formulate this task as a loop in which an agent refines a 3D prediction $M_t$ at each step $t$ with an interactive segmentation model $\fseg$ (\cref{fig:teaser}, middle-bottom).
To keep each MLLM call affordable, the agent observes $V$ and $M_t$ only through a few 2D planes rendered from its current view $v_t=(c_t,F_t)$.
The agent sees only these planes, so we model this loop as a partially observable Markov decision process (POMDP) whose state includes the hidden $M^\ast$:
\begin{equation}
\underbrace{(M^\ast,M_t,v_t)}_{\text{state},\ s_t},\quad
\underbrace{R(V,M_t;v_t)}_{\text{observation},\ o_t},\quad
\underbrace{a_t\sim\pi(\cdot\mid o_t,q)}_{\text{policy}},
\label{eq:pomdp}
\end{equation}
where $R$ only renders: it draws planes of $V$ from $v_t$ with $M_t$ on them in red as an \emph{overlay} (in \method, these planes are the tri-planes; \cref{sec:triview}).
An action $a_t$ is a positive or negative \emph{click} at a voxel, a \emph{move} to a new view $v_{t+1}$, or an action that ends the loop at step $T$ and outputs the final prediction $M=M_T$.
Only a click changes the prediction, through $\fseg$: $M_{t+1}=\fseg(V,M_t,a_t)$.
The policy thus takes 2D images as input but must output actions in 3D.

\myparagraph{Overview.}
\method instantiates this POMDP as shown in \cref{fig:workflow}.
Its observation $o_t$ consists of the \emph{tri-planes} of $v_t$: the axial, coronal, and sagittal planes through $c_t$, each rendered as one image, with $M_t$ as an overlay (\cref{sec:triview}).
For brevity, we say that the MLLM judges a view when it judges the tri-planes of that view.
The agent that carries out the policy $\pi$ consists of an MLLM and a \emph{harness}, the program around the MLLM (\cref{sec:harness}).
In each step, the MLLM only makes visual judgments on the tri-planes, and the harness does the rest in code.

\subsection{Tri-planes as a context-efficient observation}
\label{sec:triview}

Once the target is located, each judgment in the loop is local, since an error region of the prediction can be recognized from its surroundings.
An MLLM call therefore needs only one or a few views, and showing the whole volume would spend tokens on regions that the call does not judge.
Its input should instead be compact yet likely to show the error regions in its view.
Tri-planes meet both needs with only three images (we drop the index $t$ in this section).

\myparagraph{Context efficiency.}
We compare tri-planes with stacks of parallel slices, the standard way to show a volume as 2D images, for the same view.
Since an MLLM turns each image into visual tokens in proportion to its pixels~\cite{bai2025qwen25vl} or into a roughly fixed number of tokens~\cite{georgiev2024gemini15,li2025llavaonevision}, we examine how many pixels and images a stack needs to show what tri-planes show around $c$ in the same detail.
We measure this detail by the sampling rate $r$, the number of samples per view width on each of the three planes, so that neighboring samples lie at most $F/r$ apart.
Tri-planes render each plane as one $n\times n$ image, so they reach $r=n$ with $3n^2$ pixels.
A stack of $N_{\mathrm{s}}$ slices of $n\times n$ pixels samples the two planes that cross its slices only at its slice positions, so its sampling rate is at most $\min(n,N_{\mathrm{s}})$.
Reaching sampling rate $r$ thus takes at least $r$ slices of $r\times r$ pixels, \ie, $r$ images and $r^3$ pixels, $r/3$ times as many as tri-planes in either count (\cref{fig:triview}a; \supplref{rem:suppl_budget}).

\myparagraph{Sufficient detail.}
Conversely, at the same pixel budget, tri-planes show more detail than any stack.
Let $r_{\text{tri}}(B)$ and $r_{\text{stack}}(B)$ denote the largest sampling rates at which tri-planes and a slice stack, respectively, show the three planes with at most $B\ge3$ pixels.
By \supplref{cor:suppl_budget}, these sampling rates satisfy:
\begin{equation}
\underbrace{r_{\text{tri}}(B)=\lfloor\sqrt{B/3}\rfloor}_{\text{tri-planar}}
\quad\text{versus}\quad
\underbrace{r_{\text{stack}}(B)\le\sqrt[3]{B}}_{\text{any slice stack}}.
\label{eq:detail}
\end{equation}
For example, with the pixels of tri-planes at $n=512$, a stack reaches a sampling rate of at most $92$.
A stack that instead keeps $512\times512$ slices within this budget becomes sparse: it has only three slices and leaves gaps between them (\cref{fig:triview}b).
These gaps can hide an error region that is thin along the stacking axis, whereas among the tri-planes, only the plane parallel to the slices can miss the region due to such thinness.
For randomly placed views and slices, we prove that tri-planes are far more likely than the same number of parallel slices to show an error region in the worst case (\supplref{sec:suppl_detect}).

\subsection{A harness for acting on tri-planes}
\label{sec:harness}

To separate visual judgments from exact geometry, the harness keeps the volume $V$, the 3D prediction $M_t$, and all clicks in code and shows the MLLM only rendered images.
It renders the tri-planes with the overlay and maps each box or point that the MLLM selects on a plane back to voxels.
This mapping needs no depth estimate: each plane is axis-aligned and passes through the known center $c_t$, so every pixel corresponds to exactly one voxel.
The harness has three parts: (i) initialization starts the loop, (ii) click selection from candidates refines the prediction, and (iii) multi-view verification decides whether to accept the prediction.

\myparagraph{Initialization.}
The agent starts the loop from prior knowledge of where the target lies.
This prior matters because a small target appears on only a few of the hundreds of slices, so an MLLM that searches the whole scan from the text alone often misses it (\supplref{sec:suppl_abl_loc}).
The agent therefore takes the localization prior, a coarse mask of the target, from a localization model $\floc$.
A preprocessing step then turns this prior into the initial state $(M_0,v_0)$ (\cref{fig:workflow}a; details in \supplref{sec:suppl_init}).
If the localization fails, the MLLM searches the whole scan on axial slices as a fallback, which recovers some of the targets that the localization misses.

\myparagraph{Click selection from candidates.}
The agent refines the prediction with clicks: a positive click adds a missing part of the target, and a negative click removes an extra part.
Each click must be exact, since a click only a few pixels off can land on the wrong side of the prediction boundary.
MLLMs, however, are less reliable at giving exact pixel positions~\cite{rahmanzadehgervi2024blind,yang2023setofmark}.
The harness therefore splits each click into four sub-steps (\cref{fig:workflow}d): the MLLM only locates the error roughly and picks among candidate points, and the harness computes every exact position in code.
\circled{1}~The MLLM draws a rough box around a missing or an extra part on one plane.
Because a click on a false alarm would damage a correct part of the prediction, the harness checks each box before clicking.
It first drops any box that contradicts the overlay, \eg, a box that the MLLM marks as a missing part although the overlay already covers most of it.
For each remaining box, the harness renders a new view centered on the box, so that the MLLM can examine the error in all three planes before confirming or withdrawing it.
\circled{2}~Let $b_t$ denote the set of voxels that the box covers on its plane.
Inside $b_t$, the harness samples $N\le5$ candidate points $x_1,\dots,x_N$ on the correct side of the prediction boundary:
\begin{equation}
x_j\in\begin{cases}b_t\setminus M_t & \text{missing part (positive click)},\\ b_t\cap M_t & \text{extra part (negative click)},\end{cases}
\label{eq:candidates}
\end{equation}
and draws them as rings with numbers on the plane of the box.
\circled{3}~The MLLM picks one number $j^\ast$, and the harness maps it to the exact voxel $x_{j^\ast}$.
The click label $\ell_t$ is $+1$ for a missing part and $-1$ for an extra part, so the action is $a_t=(x_{j^\ast},\ell_t)$.
\circled{4}~$\fseg$ turns this click into a proposed prediction, which becomes $M_{t+1}$ only if it passes basic checks and the MLLM confirms its overlay; otherwise, the harness undoes the click (\supplref{sec:suppl_click}).
\emph{The MLLM thus never outputs a click position, yet every click lands at an exact voxel on the correct side of the prediction boundary.}

\myparagraph{Multi-view verification.}
Tri-planes show the 3D prediction only as three cross-sections within one view, so the MLLM may accept the overlay in one view even when the prediction misses a distant part of the target or spills into a nearby organ.
The prediction must therefore also pass in other views, and the following property of a click shows where new error regions can appear:

\begin{remark}[Where new error regions can appear]
\label{rem:verify}
An \emph{error region} is a connected component of voxels where the prediction and $M^\ast$ disagree, away from the boundary of $M^\ast$, as long as the component covers at least a fixed fraction of the target volume (\supplref{sec:suppl_rem_verify}).
Let $\Delta_t$ denote the region where $M_{t+1}$ and $M_t$ differ.
Since a click changes the prediction only inside $\Delta_t$, every error region of $M_{t+1}$ that does not intersect $\Delta_t$ was already part of an error region of $M_t$.
For the same reason, if a view's planes do not intersect $\Delta_t$, its tri-planes remain unchanged, so the MLLM does not need to judge this view again.
\end{remark}

\noindent Since the harness knows $\Delta_t$ from the two predictions, it places one view on this region to check for new error regions.
Error regions that existed before the click can lie anywhere, so the other views cover the places where a prediction often goes wrong: its center, its surface, and its separate pieces.
Together, these views, at most $K$ of them, form the \emph{verification views} $\mathcal{V}(M_t)$ (\cref{fig:workflow}c), and the supplement describes how the harness places them, including the extra views after a large correction (\supplref{sec:suppl_verify}).
After each correction, the agent verifies the updated prediction again, and it accepts the prediction once all views of a verification pass.
Under the idealized assumptions below, the chance that the agent accepts a wrong prediction during the whole run stays below a limit that falls geometrically as each verification uses more views (proof in \supplref{sec:suppl_idealized}):
\looseness=1

\begin{corollary}[Reaching a prediction without error regions]
\label{cor:verify}
Assume that $M_0$ has $N_{\mathrm{err}}$ error regions and that three conditions hold:
(i)~\emph{correct judgments}, \ie, the MLLM judges every view correctly;
(ii)~\emph{coverage}, \ie, in every verification, at least $k$ views each show a remaining error region with probability at least $\pcov$, even after earlier views have missed it; and
(iii)~\emph{clean corrections}, \ie, each error that the MLLM confirms leads to a correction that clears at least one error region of $M_0$ without expanding an existing error region or introducing a new one.
Without budget limits, the agent then accepts a prediction without error regions within $N_{\mathrm{err}}$ corrections with probability at least $1-\mathbb{E}[N_{\mathrm{err}}](1-\pcov)^k$, where probabilities are taken over the unknown $M^\ast$ given what the agent has observed.
\end{corollary}

\noindent Keeping the failure bound $\mathbb{E}[N_{\mathrm{err}}](1-\pcov)^k$ below a target level $\alpha$ thus requires only $k\ge\log(\mathbb{E}[N_{\mathrm{err}}]/\alpha)/\log\frac{1}{1-\pcov}$ views, a number that grows logarithmically with $\mathbb{E}[N_{\mathrm{err}}]/\alpha$.
The agent knows neither $N_{\mathrm{err}}$ nor $\pcov$, so the harness instead uses a fixed rule with at most $K$ views (\supplref{sec:suppl_verify}).
The coverage $\pcov$ depends on the target and on the size and shape of each error region.
For error regions that cover at least 10\% of the target, the first view of a verification already shows one in more than 90\% of the verifications whose prediction has one (\supplref{sec:suppl_abl_harness}).
However, the real loop departs from assumptions (i) and (iii): the MLLM does not always judge correctly, and a correction can create new error regions (\cref{rem:verify}).
When the MLLM misses an error region that a view shows, the agent may accept a clearly wrong prediction, which happens in a small fraction of cases (\cref{sec:rq4}); when it stays unsure about a view after two retries, the agent stops without accepting the prediction (\supplref{sec:suppl_verify}).
\supplref{sec:suppl_idealized} gives a bound on the outcome when (i) or (iii) fails.

\section{Experiments}
\label{sec:exp}

We evaluate \method with three research questions:
\textbf{RQ1} (\cref{tab:main,fig:prior}): Can it annotate ROIs of different types and sizes automatically and reliably?
\textbf{RQ2} (\cref{fig:view}): Do tri-planes provide a more context-efficient observation?
\textbf{RQ3} (\cref{sec:rq4}): How does each design of the harness, namely the initial prediction, click selection from candidates, and multi-view verification, contribute to the annotation?

\begin{table*}[t]
\centering
\small
\setlength{\tabcolsep}{4pt}
\caption{
\textbf{Performance on the benchmark.}
The Lesion group covers lymph nodes and lesions (tumors, stones, cysts, and stroke lesions).
Oracle: $\fseg$ with five iterative clicks from the ground truth, shown as a reference.
SegVol scores zero on 365 of the 690 ROIs, so its median is zero.
The number in parentheses after each mean is the half-width of its 95\% confidence interval, from resampling the ROIs (\supplref{sec:suppl_stats}).
}
\label{tab:main}
\ADLnullwidehline %
\resizebox{\textwidth}{!}{%
\begin{tabular}{l ll cc cccc cc cc}
\toprule
& & & \multicolumn{2}{c}{\textbf{Overall}}
& \multicolumn{4}{c}{\textbf{Anatomy} (453)}
& \multicolumn{2}{c}{\textbf{Lesion} (237)}
& \multicolumn{2}{c}{\textbf{ROI size}} \\
\cmidrule(lr){4-5}\cmidrule(lr){6-9}\cmidrule(lr){10-11}\cmidrule(l){12-13}
\textbf{Category} & \textbf{Method} & \textbf{MLLM} & \textbf{Mean} & \textbf{Median}
& \textbf{Organ} & \textbf{Muscle} & \textbf{Vessel} & \textbf{Bone}
& \textbf{Lesion} & \textbf{Lymph node}
& \textbf{Small} & \textbf{Large} \\
\midrule
\rowcolor{black!6} Reference & \multicolumn{2}{l}{Oracle (5 clicks)} & 78.9\phantom{\ (0.0)} & 84.5 & 81.2 & 82.6 & 80.1 & 35.7 & 76.7 & 73.9 & 71.7 & 83.2 \\
\midrule
\multirow{5}{*}{End-to-end} & \multicolumn{2}{l}{SegVol~\cite{du2024segvol}} & 26.3 (2.8) & 0.0 & 37.0 & 1.5 & 53.2 & 0.0 & 7.0 & 0.0 & 8.0 & 37.3 \\
& M3D-LaMed~\cite{bai2024m3d} & Phi-3-4B~\cite{abdin2024phi3} & 26.8 (2.5) & 4.2 & 38.9 & 16.2 & 50.6 & 0.0 & 5.0 & 2.9 & 8.4 & 38.0 \\
& \multicolumn{2}{l}{Text3DSAM~\cite{xin2026text3dsam}} & 44.8 (2.3) & 47.9 & 52.3 & 55.9 & 52.3 & \underline{33.7} & 31.0 & 32.9 & 27.4 & 55.3 \\
& \multicolumn{2}{l}{SAT-Pro~\cite{zhao2025sat}} & 50.1 (2.8) & 65.2 & 66.1 & 10.6 & 72.4 & 0.2 & 25.6 & 0.0 & 31.0 & 61.7 \\
& \multicolumn{2}{l}{VoxTell~\cite{rokuss2026voxtell}} & 58.4 (2.5) & 71.2 & 69.3 & 60.8 & 71.2 & 30.8 & 41.2 & 12.3 & 41.5 & 68.7 \\
\midrule
2D agent & MedSAM-Agent~\cite{liu2026medsamagent} & Qwen3-VL-8B~\cite{bai2025qwen3vl} & 30.1 (2.5) & 14.2 & 41.4 & 0.0 & 33.9 & 0.9 & 13.8 & 1.2 & 7.4 & 43.8 \\
\midrule
\multirow{3}{*}{3D agent} & \multirow{3}{*}{nnInteractive Agent~\cite{isensee2026nninteractiveagent}} & GPT-5.6-Sol~\cite{openai2026gpt56} & 36.7 (2.9) & 21.0 & 34.6 & 46.1 & 25.8 & 28.5 & 43.5 & 17.8 & 20.2 & 46.8 \\
& & Claude-Sonnet-5.5~\cite{anthropic2026sonnet55} & 44.3 (2.9) & 49.0 & 45.8 & 65.1 & 33.9 & \underline{33.7} & 44.0 & 24.0 & 23.7 & 56.7 \\
& & GPT-6.1-Sol~\cite{openai2026gpt61sol} & 59.7 (2.5) & 72.4 & 64.5 & 42.3 & 62.6 & \textbf{37.0} & 54.1 & 36.3 & 42.3 & 70.3 \\
\midrule
\rowcolor{blue!5} & & GPT-5.6-Sol~\cite{openai2026gpt56} & 64.2 (2.3) & 74.9 & 70.8 & 68.4 & 69.6 & 20.7 & 54.4 & 47.9 & 51.3 & 72.0 \\
\rowcolor{blue!5} & & Claude-Sonnet-5.5~\cite{anthropic2026sonnet55} & 63.2 (2.5) & 75.7 & 72.3 & 75.3 & 71.5 & 29.2 & 47.5 & 48.4 & 47.4 & 72.8 \\
\rowcolor{blue!5} & & \rule[-1.1ex]{0pt}{0pt}GPT-6.1-Sol~\cite{openai2026gpt61sol} & 68.8 (2.1) & 80.0 & 74.1 & \underline{78.0} & 74.6 & 23.9 & \underline{61.2} & \textbf{51.6} & \underline{54.3} & 77.6 \\
\cdashline{3-13}
\rowcolor{blue!5} & & \rule{0pt}{2.6ex}DeepSeek-V4.1-Flash~\cite{deepseek2026v41flash} & 64.7 (2.5) & 77.9 & 74.0 & 75.3 & 74.1 & 22.4 & 49.3 & 47.5 & 46.5 & 75.7 \\
\rowcolor{blue!5} & & Gemini-3.8-Flash~\cite{google2026gemini38flash} & 66.3 (2.3) & 78.7 & 75.1 & 77.8 & 70.9 & 24.5 & 52.4 & 48.1 & 51.5 & 75.3 \\
\rowcolor{blue!5} & & Claude-Opus-5.5~\cite{anthropic2026opus55} & \underline{69.2} (2.2) & \textbf{80.5} & \textbf{76.0} & \textbf{78.8} & \underline{75.0} & 28.9 & 58.8 & 49.0 & 52.8 & \textbf{79.0} \\
\rowcolor{blue!5} \multirow{-7}{*}{Ours} & \multirow{-7}{*}{\method} & GPT-6-Astra~\cite{openai2026gpt6astra} & \textbf{70.0} (2.1) & \underline{80.1} & \underline{75.2} & \underline{78.0} & \textbf{75.4} & 29.0 & \textbf{62.7} & \underline{50.9} & \textbf{55.6} & \underline{78.7} \\
\bottomrule
\end{tabular}
}
\end{table*}

\subsection{Experiment Setup}
\label{sec:setup}

\myparagraph{Benchmark.}
We prioritize diversity of ROI types over the number of cases per type (\supplref{sec:suppl_bench}).
The benchmark comprises 690 ROIs in 262 volumes from 35 public CT, MR, and PET datasets, covering organs, vessels, bones, muscles, tumors, and lymph nodes.
We call an ROI small when the equivalent spherical diameter of its ground-truth mask is below 3\,cm (260 ROIs) and large otherwise (\supplref{sec:suppl_bench}).
It also includes lesion types from datasets that neither $\floc$ nor $\fseg$ saw in training~\cite{wu2025mswal,huang2026dolchid,pandit2025pituitary,liang2025utendomri}.
For in-distribution datasets, we take only validation or test cases (\supplref{sec:suppl_bench}).
To limit API costs, we use a 190-ROI subset for all experiments except \cref{tab:main,fig:prior} (\supplref{sec:suppl_bench}).
This subset still covers every target of the benchmark, since it keeps at least one ROI for each target of each dataset.
Outside \cref{tab:main}, \method runs with Gemini-3.8-Flash unless noted otherwise, and \supplref{sec:suppl_abl_layout,sec:suppl_abl_harness} repeat the input-layout comparison and the ablation with DeepSeek-V4.1-Flash.

\myparagraph{Baselines and metrics.}
We compare \method with five end-to-end 3D segmentation models, VoxTell~\cite{rokuss2026voxtell}, SAT-Pro~\cite{zhao2025sat}, Text3DSAM~\cite{xin2026text3dsam}, SegVol~\cite{du2024segvol}, and M3D-LaMed~\cite{bai2024m3d}, and the 2D agent MedSAM-Agent~\cite{liu2026medsamagent}, using weights from their official repositories.
The five end-to-end segmentation models receive only the target name, whereas MedSAM-Agent segments slice by slice within the slice range of the $\floc$ box.
As a 3D agent baseline, we run nnInteractive Agent~\cite{isensee2026nninteractiveagent} in its default configuration, using Codex~\cite{openai2025codex} with GPT-6.1-Sol or GPT-5.6-Sol, and Claude Code~\cite{anthropic2025claudecode} with Claude-Sonnet-5.5.
\supplref{sec:suppl_setup} gives more detailed settings of all systems.
To test whether our designs can be helpful for general coding agents, we evaluate Codex~\cite{openai2025codex} with GPT-6.1-Sol~\cite{openai2026gpt61sol}, which drives the same $\fseg$ (\supplref{sec:suppl_codex}).
We evaluate every system with 3D Dice (\%).
For the agents, we also report MLLM calls per ROI and the number of false accepts (FA).
In all tables, the best and second-best results are bolded and underlined.

\myparagraph{Implementation details.}
We implement \method as a finite-state machine in LangGraph~\cite{langchain2024langgraph}, with BiomedParse~v2~\cite{zhao2025biomedparse,zhao2025boltzformer} as $\floc$ and nnInteractive~\cite{isensee2025nninteractive} as $\fseg$. In practice, the harness places the three planes of each view side by side in one image. All MLLMs in \cref{tab:main} use identical prompts and their default reasoning effort.
We tuned the prompts and parameters of the agent, such as the number of verification views $K$, on a separate 74-ROI development set that shares no cases with the benchmark (\supplref{sec:suppl_bench}).
The agent makes at most 26 MLLM calls for each ROI (\supplref{sec:suppl_ablation} varies this budget).
All experiments run on a single NVIDIA RTX 4090 GPU.

\subsection{Results}
\begin{figure}[t]
\centering
\includegraphics[width=\columnwidth]{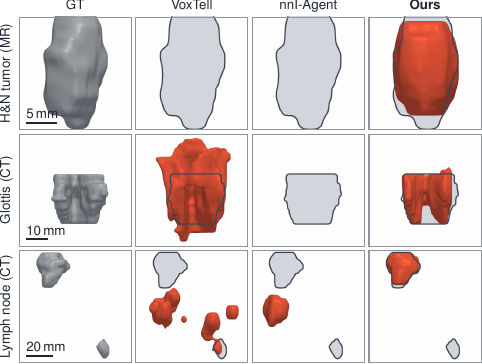}
\caption{\textbf{Comparison of small-target segmentation.}
The three targets measure 13, 21, and 22\,mm from top to bottom; gray is the ground truth and orange the 3D prediction, and a panel without orange has no prediction in view.
nnI-Agent: nnInteractive Agent~\cite{isensee2026nninteractiveagent} with GPT-6.1-Sol~\cite{openai2026gpt61sol}.
\supplref{sec:suppl_qualitative} shows more cases.}
\label{fig:qualitative}
\end{figure}

\myparagraph{Performance on our benchmark.}
With each MLLM, \method achieves a higher overall Dice than every baseline, and frontier MLLMs give it the highest Dice (\cref{tab:main}).
With the same MLLM, it outperforms nnInteractive Agent for each of the three MLLMs that both agents run, and even with a flash-level MLLM, it outperforms nnInteractive Agent running a frontier model.
The other baselines learn their targets from training data and generalize poorly: on rare lesion types such as gallstones and pituitary tumors, they score far lower than both agents.
With frontier MLLMs, the lead of \method over the strongest baseline is even larger on small targets (\cref{fig:qualitative}).
Unlike nnInteractive Agent, \method's harness computes click position through code to avoid missing small targets.
With frontier MLLMs, \method also leads on every ROI type except bones.
All ten bone ROIs come from MR scans, and even $\fseg$ with clicks from the ground truth separates them poorly from nearby tissue.
For \method, a box around a thin or curved part of a bone also covers mostly other tissue, so its candidate points may not land on the bone.

\begin{table}[t]
\centering
\footnotesize
\setlength{\tabcolsep}{1.9pt}
\renewcommand{\arraystretch}{1.0}
\caption{
\textbf{Generalization of the harness design to other agents.}
Prior: the first view comes from $\floc$.
Codex accepts every prediction that it submits, and its calls are the turns that ran a command; --: not recorded.
Codex runs GPT-6.1-Sol and nnInteractive Agent runs GPT-5.6-Sol.
}
\label{tab:codex}
\resizebox{\columnwidth}{!}{%
\begin{tabular}{@{}l cc cc cc cc@{}}
\toprule
\multirow{2}{*}{\textbf{Method}}
& \multicolumn{2}{c}{\textbf{Observation}}
& \multicolumn{4}{c}{\textbf{Dice}}
& \multicolumn{2}{c}{\textbf{Behavior}} \\
\cmidrule(lr){2-3}\cmidrule(lr){4-7}\cmidrule(l){8-9}
& \textbf{Input} & \textbf{Prior}
& \textbf{Mean} & \textbf{Median} & \textbf{Small} & \textbf{Large}
& \textbf{FA}$\downarrow$ & \textbf{Calls}$\downarrow$ \\
\midrule
\multirow{4}{*}{Codex~\cite{openai2025codex}} & volume & \xmark & 62.3 & 74.6 & 43.6 & \underline{73.0} & 28 & 27.8 \\
 & axial & \xmark & 55.2 & 68.3 & 40.7 & 63.4 & 42 & 29.9 \\
 & tri-planes & \xmark & 53.8 & 66.4 & 37.7 & 63.0 & 46 & \underline{24.2} \\
 & tri-planes & \cmark & \underline{63.0} & 74.7 & \underline{46.9} & 72.3 & \underline{19} & 25.1 \\
\midrule
\multirow{2}{*}{\makecell[l]{nnInteractive\\Agent~\cite{isensee2026nninteractiveagent}}} & default & \xmark & 34.2 & 14.3 & 13.6 & 45.9 & -- & -- \\
 & default & \cmark & 58.9 & \underline{75.4} & 44.5 & 67.1 & -- & -- \\
\midrule
\rowcolor{blue!5} \method & tri-planes & \cmark & \textbf{65.8} & \textbf{79.3} & \textbf{49.3} & \textbf{75.3} & \textbf{12} & \textbf{11.7} \\
\bottomrule
\end{tabular}
}
\end{table}
\begin{figure}[t]
\centering
\includegraphics{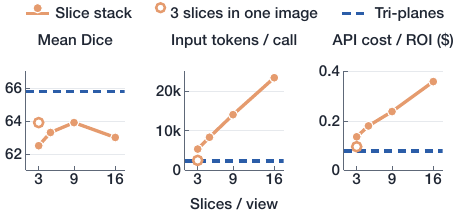}
\caption{
\textbf{Context efficiency of the input layout.}
Only the input layout differs between the lines.
Slice stack: parallel axial slices around the view center, one image per slice; tri-planes always show three planes, so they appear as a dashed line.
}
\label{fig:view}
\end{figure}

\myparagraph{Context efficiency of tri-planes.}
\label{sec:rq2}
Tri-planes are the most context-efficient input layout while preserving performance (\cref{fig:view}; \supplref{sec:suppl_abl_layout}).
For a slice stack, the input tokens per call increase with the number of slices, whereas its Dice does not improve and stays below that of tri-planes.
Tri-planes use the fewest input tokens per MLLM call (up to $9.8\times$ fewer than a 16-slice stack), yet their detail is sufficient for the annotation.
They also need the fewest reasoning tokens and calls, which gives them the lowest API cost per ROI.

\begin{figure}[t]
\centering
\includegraphics{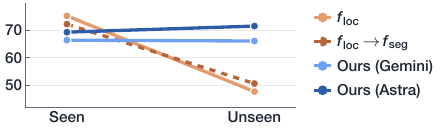}
\caption{
\textbf{Generalization to unseen datasets.}
Dice of $\floc$, of $\floc{\rightarrow}\fseg$ (no MLLM), and of \method with Gemini-3.8-Flash and GPT-6-Astra on the seen (480 ROIs) and unseen (210 ROIs) datasets of $\floc$.
}
\label{fig:prior}
\end{figure}
\myparagraph{Generalization to unseen datasets.}
On the unseen datasets of $\floc$, \method outperforms both $\floc$ and $\floc{\rightarrow}\fseg$, in which $\fseg$ segments the target from the mask of $\floc$ without an MLLM (\cref{fig:prior}; \supplref{sec:suppl_abl_loc}).
The Dice of both drops sharply from the seen to the unseen datasets, whereas that of \method with Gemini-3.8-Flash or GPT-6-Astra remains stable across the two.
Since $\fseg$ raises the Dice of $\floc$ only slightly on the unseen datasets, the stable Dice of \method comes from the agent rather than from $\fseg$.
\method does not simply accept the prior: it checks the initial prediction from the prior in tri-planes and corrects it through clicks, and when $\floc$ finds nothing, it searches the whole scan for the target on its own.
With frontier MLLMs, \method also outperforms $\floc$ over all ROIs (\supplref{sec:suppl_abl_loc}).

\myparagraph{Generalizability to other agents.} Tri-planes and the localization prior help other agents (\cref{tab:codex}).
Compared with axial slices, tri-planes let Codex reach a similar Dice with fewer calls, due to their sufficient detail.
The prior brings the largest gain to both Codex and nnInteractive Agent, because it starts both agents at the target.
Even with both designs, Codex with a frontier MLLM scores below \method with a flash-level MLLM and needs 2.1$\times$ as many calls, because without our harness, its MLLM must give every click position itself.

\subsection{Ablation Study}
\label{sec:rq4}

\myparagraph{Prior mask for initialization.}
Initializing $\fseg$ with the prior mask instead of a single click raises the mean Dice by 12.8 points and halves the false accepts (\cref{tab:ablation}).
Since an interactive model such as $\fseg$ is sensitive to its first prompt, and later clicks only refine the prediction that this prompt produces~\cite{lin2020fca,isensee2025nninteractive}, outlining the whole target gives a much better start than marking one point.

\begin{table}[t]
\centering
\footnotesize
\setlength{\tabcolsep}{4pt}
\renewcommand{\arraystretch}{1.0}
\caption{
\textbf{Ablation of the harness.}
\xmark\ marks a removed design, and the shaded row is the full \method.
Without the prior mask, $\fseg$ starts from one click that the harness places on the same view; without click selection, the MLLM outputs the coordinates of every click itself.
}
\label{tab:ablation}
\resizebox{\columnwidth}{!}{%
\begin{tabular}{@{}ccc cc cc c@{}}
\toprule
\multirow{2}{*}{\makecell{\textbf{Prior}\\\textbf{mask}}} & \multirow{2}{*}{\makecell{\textbf{Click}\\\textbf{selection}}} & \multirow{2}{*}{\makecell{\textbf{Multi-view}\\\textbf{verification}}}
& \multicolumn{4}{c}{\textbf{Dice}}
& \multirow{2}{*}{\textbf{FA}$\downarrow$} \\
\cmidrule(lr){4-7}
& &
& \textbf{Mean} & \textbf{Median} & \textbf{Small} & \textbf{Large} & \\
\midrule
\xmark & \cmark & \cmark & 53.0 & 63.8 & 37.1 & 62.1 & 25 \\
\cmark & \xmark & \cmark & 61.4 & 77.9 & 44.6 & 71.0 & \underline{15} \\
\cmark & \cmark & \xmark & \underline{65.1} & \underline{79.1} & \underline{48.5} & \underline{74.6} & 16 \\
\rowcolor{blue!5} \cmark & \cmark & \cmark & \textbf{65.8} & \textbf{79.3} & \textbf{49.3} & \textbf{75.3} & \textbf{12} \\
\bottomrule
\end{tabular}
}
\end{table}

\begin{figure*}[t]
\centering
\includegraphics[width=\textwidth]{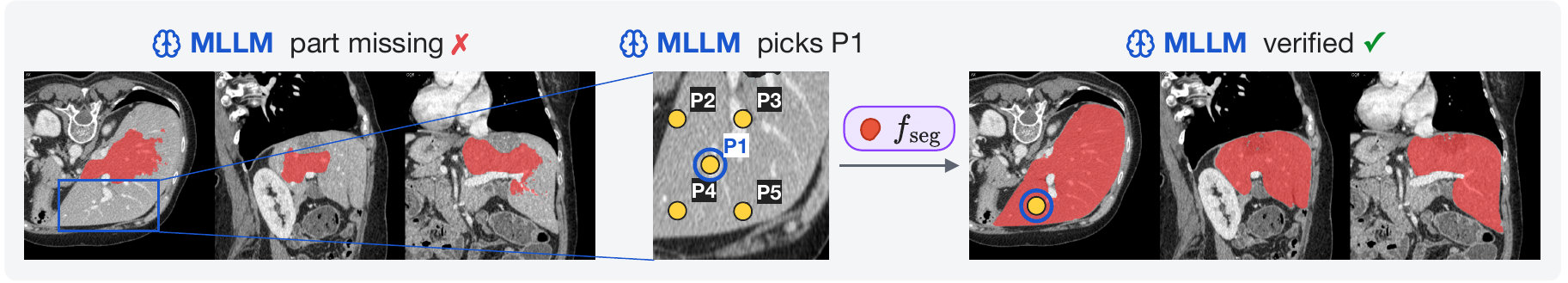}
\caption{\textbf{One click of \method on the liver} (CT, AbdomenAtlas).
The red overlay is the 3D prediction before (\emph{left}) and after (\emph{right}) the click; the middle panel enlarges the blue box with the candidate points.}
\label{fig:casestudy}
\end{figure*}

\myparagraph{Click selection from candidates.}
Selecting among candidate points, rather than letting the MLLM output raw coordinates, raises the mean Dice on both small and large targets (\cref{tab:ablation}).
The gain comes from exact clicks: the harness places every candidate on an exact voxel on the correct side of the prediction boundary (\cref{sec:harness}).
For example, \cref{fig:casestudy} shows how the MLLM picks candidates on the liver. 

\begin{figure}[t]
\centering
\includegraphics{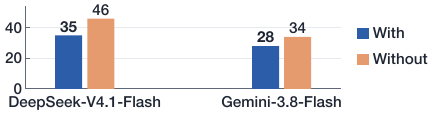}
\caption{
\textbf{Accepted ROIs with Dice below 50\%}, with and without multi-view verification.
}
\label{fig:verification}
\end{figure}
\myparagraph{Multi-view verification.}
Multi-view verification reduces the accepted ROIs with Dice below 50\% for both MLLMs (\cref{fig:verification}); with Gemini-3.8-Flash, it also lowers the false accepts and raises the mean Dice (\cref{tab:ablation}).
Without it, the agent accepts a prediction as soon as a single view passes, so it accepts more wrong predictions (\cref{tab:ablation}).
It helps both compact ROIs and ROIs with complex shapes such as the tubular aorta (\supplref{sec:suppl_ablation}).

\pagebreak
\section{Conclusion}
\label{sec:conclusion}
We presented \method, an MLLM agent that annotates 3D volumes through an interactive loop with a segmentation model.
The agent observes the volume through tri-planes, which keep each MLLM call compact yet likely to show error regions in the 3D prediction.
Within the agent, the MLLM only judges the prediction on tri-planes, while the harness turns each judgment into a 3D action in code and checks each 3D prediction from several views before the agent accepts it.
With frontier MLLMs, \method achieves a higher overall Dice than every baseline, and with flash-level MLLMs, it still outperforms nnInteractive Agent running a frontier model.
Tri-planes use up to 9.8$\times$ fewer input tokens per MLLM call than a 16-slice stack, without loss in performance.

\section*{LLM Usage Statement}
The MLLMs and coding agents in \cref{tab:main,tab:codex} are the subject of our experiments and run as part of the evaluated systems.
Separately, we used generative AI tools (Claude Code and Codex) to draft and edit this paper, including the LaTeX of its tables and the source code of its figures.
We also used them when debugging the research code and managing the experiments.
For the related work, these tools helped us find references, and we manually checked that every citation is correct.

{
    \small
    \bibliographystyle{ieeenat_fullname}
    \bibliography{main}
}

\ifwithsupp
  \setlength{\textfloatsep}{\arxivTFS}\setlength{\dbltextfloatsep}{\arxivDTFS}
\setlength{\floatsep}{\arxivFS}\setlength{\dblfloatsep}{\arxivDFS}\setlength{\intextsep}{\arxivITS}
\captionsetup{skip=10pt}\flushbottom
\maketitlesupplonecolumn
  \appendix
  \setcounter{equation}{0}\renewcommand{\theequation}{S\arabic{equation}}\renewcommand{\theHequation}{S.\arabic{equation}}
  \setcounter{figure}{0}\renewcommand{\thefigure}{S\arabic{figure}}\renewcommand{\theHfigure}{S.\arabic{figure}}
  \setcounter{table}{0}\renewcommand{\thetable}{S\arabic{table}}\renewcommand{\theHtable}{S.\arabic{table}}
  \setcounter{proposition}{0}\def\theHproposition{S.\arabic{proposition}}\renewcommand{\theproposition}{S\arabic{proposition}}
  \setcounter{lemma}{0}\def\theHlemma{S.\arabic{lemma}}\renewcommand{\thelemma}{S\arabic{lemma}}
  \setcounter{corollary}{0}\def\theHcorollary{S.\arabic{corollary}}\renewcommand{\thecorollary}{S\arabic{corollary}}
  \setcounter{remark}{0}\def\theHremark{S.\arabic{remark}}\renewcommand{\theremark}{S\arabic{remark}}
  \setlength{\parindent}{0pt}
\setlength{\parskip}{0.4em}
\clubpenalty=10000
\widowpenalty=10000
\newcolumntype{Y}[1]{>{\columncolor{white}[0pt][\tabcolsep]}#1}
\newcolumntype{Z}[1]{>{\columncolor{white}[\tabcolsep][0pt]}#1}

\begingroup
\section*{Contents}
Each section of this supplement expands one part of the main paper.
The table below lists what each section contains, the part of the main paper that it supports, and the page where it starts.

\vspace{0.4em}
\begin{center}
\footnotesize
\setlength{\tabcolsep}{5pt}
\renewcommand{\arraystretch}{1.3}
\begin{tabular}{@{}c l p{0.44\textwidth} l r@{}}
\toprule
\textbf{Sec.} & \textbf{Title} & \textbf{What it contains} & \textbf{Main paper} & \textbf{Page} \\
\midrule
\ref{sec:suppl_notation} & Notation and Terminology & Terms with a fixed meaning and all symbols & All & \pageref{sec:suppl_notation} \\
\ref{sec:suppl_related} & Extended Related Work & Interactive segmentation tools and simulated annotators & \mainref{sec:related} & \pageref{sec:suppl_related} \\
\ref{sec:suppl_theory} & Theory of Tri-Planes & Error detection by tri-planes, the pixel budget of tri-planes and slice stacks, near-minimality, and anisotropic voxel spacing & \mainref{sec:triview} & \pageref{sec:suppl_theory} \\
\ref{sec:suppl_agent} & Agent Details & Initialization, click selection from candidates, verification views, and the acceptance rule & \mainref{sec:harness} & \pageref{sec:suppl_agent} \\
\ref{sec:suppl_verify_theory} & Theory of Multi-View Verification & Where new error regions can appear, the chance of missing an error region, and the outcome and cost of an idealized run & \mainref{rem:verify}, \mainref{cor:verify} & \pageref{sec:suppl_verify_theory} \\
\ref{sec:suppl_bench} & Benchmark Composition & Datasets and their statistics, training exposure, data splits, the development set, and the ablation subset & \mainref{sec:setup} & \pageref{sec:suppl_bench} \\
\ref{sec:suppl_setup} & Experimental Setup & MLLMs, nnInteractive Agent, and other models & \mainref{sec:setup} & \pageref{sec:suppl_setup} \\
\ref{sec:suppl_codex} & Coding-Agent Evaluation & Codex under four observation conditions & \mainref{sec:rq2} & \pageref{sec:suppl_codex} \\
\ref{sec:suppl_ablation} & Extended Experiment Results & Results with Qwen3.8-Flash, localization and generalization, input layout, harness design ablation and the coverage of verification views, and model choices and hyperparameters & \mainref{sec:rq2}, \mainref{sec:rq4} & \pageref{sec:suppl_ablation} \\
\ref{sec:suppl_stats} & Statistical Analysis & Bootstrap intervals and paired tests for the main results & \mainref{tab:main} & \pageref{sec:suppl_stats} \\
\ref{sec:suppl_qualitative} & Qualitative Results & 3D predictions compared across methods, and single clicks shown step by step & \mainref{sec:rq2} & \pageref{sec:suppl_qualitative} \\
\ref{sec:suppl_prompts} & MLLM Prompts & Every prompt that the agent sends to the MLLM & \mainref{sec:harness} & \pageref{sec:suppl_prompts} \\
\bottomrule
\end{tabular}
\end{center}
\endgroup
\clearpage

\section{Notation and Terminology}
\label{sec:suppl_notation}

\cref{tab:suppl_notation} lists the terms that the paper uses with a fixed meaning (left) and the symbols of the main paper and of \cref{sec:suppl_agent} (right).
The agent consists of an MLLM for visual judgments and a harness for rendering, tool calls, and the geometry of views and candidate points.
Symbols that appear only inside a proof in \cref{sec:suppl_theory} or \cref{sec:suppl_verify_theory} are defined in that proof.

\vspace{0.4em}
\begin{minipage}{\textwidth}
\centering
\footnotesize
\setlength{\tabcolsep}{4pt}
\renewcommand{\arraystretch}{1.15}
\captionof{table}{\textbf{Notation and terminology.} (a)~Terms that the paper uses with a fixed meaning. (b)~Symbols of the main paper and of \cref{sec:suppl_agent}.}
\label{tab:suppl_notation}
\begin{tabular}{@{}l@{}p{0.04\textwidth}@{}l@{}}
\multicolumn{1}{@{}c@{}}{\textbf{(a)} Terms} & & \multicolumn{1}{@{}c@{}}{\textbf{(b)} Symbols} \\[2pt]
\cmidrule[\heavyrulewidth]{1-1}\cmidrule[\heavyrulewidth]{3-3}
\begin{tabular}[t]{@{}>{\raggedright\arraybackslash}p{0.12\textwidth}>{\raggedright\arraybackslash}p{\dimexpr0.45\textwidth-8pt\relax}@{}}
\textbf{Term} & \textbf{Meaning} \\
\midrule
MLLM & A general-purpose multimodal large language model; it sees only rendered images and returns visual judgments, \eg, a box, a candidate number, or a decision to accept. \\
Harness & The program around the MLLM; it keeps $V$ and $M_t$, renders views, calls $\floc$ and $\fseg$, and computes the geometry of views and candidates in code. \\
Agent & The MLLM and the harness together; it clicks, moves to the next view, and either accepts the prediction or stops without accepting it. \\
MLLM agent & Any agent whose decisions come from an MLLM, \eg, \method, a 2D segmentation agent, or a coding agent. \\
\method & Our agent, with tri-planes as observations, click selection from candidates, and multi-view verification. \\
\addlinespace[3pt]
View & A cubic region $v=(c,F)$ of $V$ with center $c$ and side length $F$. \\
Tri-planes & The axial, coronal, and sagittal planes through the center $c$ of a view, each rendered as one image, with $M_t$ drawn as an overlay. \\
Overlay & The 3D prediction $M_t$ drawn in red on the tri-planes. \\
Verification view & A view centered on another part of $M_t$, which the MLLM judges before the agent accepts $M_t$. \\
\addlinespace[3pt]
Localization prior & The coarse 3D prediction $P$ that $\floc$ outputs for $q$; it centers $v_0$ and seeds $M_0$. \\
Initial prediction & $M_0$, which the MLLM chooses from two proposals that $\fseg$ grows from the main component of $P$ and from a point. \\
Proposal & A new prediction, for $M_0$ or from a click, which the harness commits only if it passes the basic checks and the MLLM confirms it. \\
Click & A positive or negative point prompt to $\fseg$ at one voxel. \\
Candidate & A point inside the box of an error that the MLLM confirms; the harness samples it and draws it as a numbered ring. \\
Error region & A connected component of voxels where a prediction and $M^\ast$ disagree, away from the boundary of $M^\ast$ and covering at least a fixed fraction of the target volume (\cref{sec:suppl_rem_verify}). \\
Confirm an error & The MLLM marks an error with a box and keeps the box through the box check and the confirmation view. \\
\addlinespace[3pt]
Accept & The agent returns $M_T$ after every verification view passes. \\
Stop without acceptance & The agent stops without accepting the prediction and records the reason and the views that it could not confirm. \\
\end{tabular}
& &
\begin{tabular}[t]{@{}>{\raggedright\arraybackslash}p{0.085\textwidth}>{\raggedright\arraybackslash}p{\dimexpr0.305\textwidth-8pt\relax}@{}}
\textbf{Symbol} & \textbf{Meaning} \\
\midrule
$V$, $h$ & Volume; its voxel spacing \\
$v=(c,F)$ & View with center $c$ and side length $F$ \\
$n$ & Image side length in pixels \\
$r$ & \leavevmode Sampling rate: samples per view width on each plane \\
$B$ & Pixel budget of an observation \\
$D$ & Total number of slices in the volume \\
$r_{\text{tri}}$, $r_{\text{stack}}$ & Largest integer sampling rate with at most $B$ pixels: tri-planes; stack \\
\addlinespace[3pt]
$q$ & Text goal \\
$M^\ast$, $M$ & Ground-truth 3D mask; final prediction \\
$M_t$, $M_0$ & 3D prediction at step $t$; initial prediction \\
$s_t$, $o_t$, $a_t$ & State, observation, and action at step $t$ \\
$\pi$, $R$ & Policy; renderer of planes with the overlay \\
$T$, $T_{\mathrm{acc}}$ & Actual stopping time; acceptance time \\
\addlinespace[3pt]
$\floc$, $P$ & Localization model; localization prior \\
$\fseg$ & Interactive segmentation model \\
$b_t$ & Voxel set of the MLLM's box on its plane \\
$x_1,\dots,x_N$ & Candidate points in $b_t$, with $N\le5$ \\
$j^\ast$, $\ell_t$ & Selected candidate number; click label ($\pm1$) \\
$\mathcal{V}(M_t)$, $K$ & Verification views of $M_t$; their largest number \\
$\chi_t$ & Whether every view in $\mathcal{V}(M_t)$ passes (1) or not (0) \\
$\mathcal{K}^+_t$, $\mathcal{K}^-_t$ & Voxel positions of positive and negative clicks up to step $t$ \\
$\Delta_t$ & Voxels that step $t$ changes, $M_{t+1}\triangle M_t$ \\
$N_{\mathrm{err}}$ & Number of error regions of $M_0$ \\
$P_0$ & Main component of $P$ \\
$c_0$, $F_0$ & Center and side length of the initial view $v_0$ \\
$d(x,\partial P_0)$ & Distance from voxel $x$ to the boundary of $P_0$ \\
$\tilde M$ & Proposed prediction, including each proposal for $M_0$ \\
\end{tabular}
\\
\cmidrule[\heavyrulewidth]{1-1}\cmidrule[\heavyrulewidth]{3-3}
\end{tabular}
\end{minipage}

\section{Extended Related Work}
\label{sec:suppl_related}

\myparagraph{Interactive segmentation tools.}
Interactive segmentation tools turn a few clicks into a prediction, but they leave the annotator to decide where this prediction is still wrong~\cite{olabarriaga2001interaction,marinov2024interactive}.
RITM~\cite{sofiiuk2022ritm} and SAM~\cite{kirillov2023sam} feed the previous prediction back into the model, so that each later click corrects this prediction instead of segmenting the target again.
Medical tools bring this loop to 3D volumes: MedSAM2~\cite{ma2025medsam2} propagates a 2D prediction from a prompted slice through the volume, whereas SAM-Med3D~\cite{wang2023sammed3d}, SegVol~\cite{du2024segvol}, and nnInteractive~\cite{isensee2025nninteractive} predict a 3D mask directly from each prompt.
In nnInteractive, a prompt on any axial, sagittal, or coronal slice updates the whole 3D mask~\cite{isensee2025nninteractive}, so one correction can change the prediction beyond the part that was wrong~\cite{olabarriaga2001interaction}.
Annotators therefore evaluate the updated 3D mask on screen before the next correction~\cite{olabarriaga2001interaction,ma2025medsam2}.
Throughout this loop, the tool draws the prediction, and the annotator supplies only the decision of where to place a positive or a negative click.
\method keeps the interactive segmentation model $\fseg$ unchanged and lets an MLLM make this decision, so it runs with either nnInteractive or MedSAM2 as $\fseg$ (\cref{tab:suppl_backend}).

\myparagraph{Simulated annotators.}
Interactive segmentation tools are trained with simulated annotators that make this click decision from the ground truth~\cite{sofiiuk2022ritm,kirillov2023sam,isensee2025nninteractive}, and benchmarks evaluate them with the same simulators~\cite{marinov2024rethinking}.
The common rule places the next click inside an error region between the prediction and the ground truth: a positive click where the target is missing, and a negative click where the prediction is extra~\cite{sofiiuk2022ritm,kirillov2023sam,marinov2024rethinking}.
MLLM agents for interactive segmentation are trained on the same simulators: SegAgent~\cite{zhu2025segagent}, MedSAM-Agent~\cite{liu2026medsamagent}, and IBISAgent~\cite{jiang2026ibisagent} fine-tune MLLMs on click trajectories that a rule-based simulator derives from ground-truth masks.
MedSAM-Agent and IBISAgent then apply reinforcement learning with rewards computed against the ground truth.
\method keeps the click rule of these simulators but replaces the ground truth with the judgment of a general-purpose MLLM on tri-planes.
Once the MLLM confirms a missing or an extra part, the harness samples candidate points on the matching side of the prediction boundary (\mainref{eq:candidates}), and the MLLM picks one of them.
\method thus places every click without ground truth, unlike a simulated annotator, and needs no training of the MLLM, unlike these agents.
Because \method needs no ground truth, it can also record click trajectories on unannotated volumes, where a simulated annotator cannot run.
Each MLLM call in such a trajectory is a self-contained pair of images and one decision, since no history passes between calls.
Such trajectories could thus serve as training data for a smaller MLLM that runs locally, enabling privacy-preserving deployment of such agents.

\section{Theory of Tri-Planes}
\label{sec:suppl_theory}

This section supports the choice of tri-planes in \mainref{sec:triview}.
Once the target is located, each judgment of the agent is local, since an error region can be recognized from its surroundings and a click creates new error regions only where it changes the prediction (\cref{sec:suppl_rem_verify}).
A call therefore needs one or a few views rather than the whole volume.
These few views must thus rarely miss an error.
For randomly placed views and slices, tri-planes are far more likely to show a set of errors in the worst case than the same number of parallel slices (\cref{sec:suppl_detect}).
Parallel slices can catch up only with more images, so we then count the pixels that tri-planes and slice stacks need to show the three planes through the view center in the same detail (\cref{sec:suppl_proof,sec:suppl_minimal,sec:suppl_aniso}).

\subsection{Error Detection by Tri-Planes}
\label{sec:suppl_detect}

A view \emph{shows} an error when one of its three images contains this voxel at the image resolution, and it \emph{misses} a set of errors when it shows none of them.
We establish a bound on the probability that randomly placed tri-planes miss a set of errors, and compare this probability with that of parallel slices using the same number of images.

\begin{proposition}[Error detection by tri-planes]
\label{prop:suppl_detect}
Let $U$ be a box of voxels, and let $E\subseteq U$ be a nonempty set of errors, connected or not, with volume fraction $\varepsilon=|E|/|U|$.
Suppose that the centers of $k$ views are drawn independently and uniformly from the voxels of $U$, that each of $3k$ parallel slices lies in a layer drawn independently and uniformly along its axis, and that every plane shows its whole cross-section of $U$ at voxel resolution.
Then:
(a)~all $k$ views miss $E$ with probability at most $\bigl(1-\varepsilon^{1/3}\bigr)^{3k}$;
(b)~the probability that all $k$ views miss $E$ is at most the probability that $3k$ parallel slices along an axis drawn uniformly at random all miss $E$, with equality only when $E$ occupies the same fraction of layers along each axis; and
(c)~if $E$ consists of whole layers along one axis, $3k$ parallel slices along this axis all miss $E$ with probability $(1-\varepsilon)^{3k}$, whereas every view shows $E$.
\end{proposition}

\begin{proof}
Let $a_x$, $a_y$, and $a_z$ be the fractions of layers along the three axes that contain a voxel of $E$.
A view misses $E$ only when all three of its layers miss $E$, and the three coordinates of a uniform center are independent, so:
\begin{equation*}
\Pr(\text{one view misses }E)=(1-a_x)(1-a_y)(1-a_z).
\end{equation*}
(a)~Since $E$ lies inside the product of its occupied layers, $\varepsilon\le a_xa_ya_z$, and the inequality of arithmetic and geometric means gives $(a_xa_ya_z)^{1/3}+[(1-a_x)(1-a_y)(1-a_z)]^{1/3}\le1$.
One view therefore misses $E$ with probability at most $(1-\varepsilon^{1/3})^3$, and the independence of the centers gives:
\begin{equation}
\Pr(\text{all }k\text{ views miss }E)\le\bigl(1-\varepsilon^{1/3}\bigr)^{3k}.
\label{eq:suppl_random_tri}
\end{equation}
(b)~A slice along axis $\ell\in\{x,y,z\}$ misses $E$ with probability $1-a_\ell$, so $3k$ slices along a uniformly drawn axis all miss $E$ with probability $\frac13\sum_\ell(1-a_\ell)^{3k}$.
By the inequality of arithmetic and geometric means, this average is at least $\bigl[\prod_\ell(1-a_\ell)^{3k}\bigr]^{1/3}=\prod_\ell(1-a_\ell)^k$, the probability that all $k$ views miss $E$, with equality only when $a_x=a_y=a_z$.
(c)~Let the axis be $z$.
Then $a_z=\varepsilon$, so each axial slice misses $E$ with probability $1-\varepsilon$, whereas $a_x=1$ means that every sagittal plane contains a voxel of $E$, so every view shows $E$.
\end{proof}

\noindent Parts (a) and (c) compare the worst cases: the $k$ views show every $E$ with probability at least $1-(1-\varepsilon^{1/3})^{3k}$, whereas $3k$ parallel slices along a fixed axis show an $E$ that fills whole layers along this axis only with probability $1-(1-\varepsilon)^{3k}$.
For small $\varepsilon$, the first probability is far larger, since $\varepsilon^{1/3}\gg\varepsilon$.
In the worst case, tri-planes are thus far more likely than the same number of parallel slices to show a set of errors.
Part (b) holds for every $E$: when the orientation of the errors is unknown, tri-planes miss them no more often than parallel slices along an axis chosen at random.
The proposition draws views at random to compare the two layouts alone, whereas the harness centers each view on a chosen point.
For a view centered at the centroid of a component, \cref{lem:suppl_centroid} shows that all three of its planes cut that component.

\subsection{Pixel Budget of Tri-Planes and Slice Stacks}
\label{sec:suppl_proof}

Part (c) of \cref{prop:suppl_detect} shows that parallel slices can miss errors that lie in the gaps between them, and a stack can close these gaps only by adding slices.
Accordingly, we count the pixels needed for an observation to sample the three orthogonal planes through the view center.

\myparagraph{Sampling positions and coverage.}
Let $F>0$ denote the side length of the cubic view $Q=[0,F]^3$ in local coordinates, centered at $c=(c_x,c_y,c_z)=(F/2,F/2,F/2)$.
The axial, coronal, and sagittal planes within $Q$ are given by:
\begin{equation*}
\begin{aligned}
\mathcal A&=\{(x,y,c_z):x,y\in[0,F]\},\\
\mathcal C&=\{(x,c_y,z):x,z\in[0,F]\},\\
\mathcal S&=\{(c_x,y,z):y,z\in[0,F]\}.
\end{aligned}
\end{equation*}
Let $\Pi=\mathcal A\cup\mathcal C\cup\mathcal S$ denote the target region that the observation must sample.

A rendered pixel samples the intensity $V(p)$ at a position $p\in Q$; resampling allows this position to lie between voxel centers.
Let $\Omega\subset Q$ denote the nonempty finite set of distinct sampling positions.
The pixel budget $B$ counts every rendered pixel, including pixels that repeat a position, so an observation within budget $B$ has $|\Omega|\le B$.

We measure how well $\Omega$ covers $\Pi$ by the distance between a target point $u\in\Pi$ and a sample $p\in\Omega$ in the $\ell_\infty$ norm:
\begin{equation*}
\|u-p\|_\infty=\max\{|u_x-p_x|,\ |u_y-p_y|,\ |u_z-p_z|\}.
\end{equation*}
The coverage radius is the largest distance from a target point to its nearest sample:
\begin{equation}
\delta(\Omega)=\max_{u\in\Pi}\min_{p\in\Omega}\|u-p\|_\infty.
\end{equation}
A small $\delta(\Omega)$ thus means that every point on the three planes lies close to a sample.

We say that an observation achieves sampling rate $r$ if every target point lies within $F/(2r)$ of a sample along each coordinate:
\begin{equation*}
\delta(\Omega)\le\frac{F}{2r}.
\end{equation*}
For example, a uniform $r\times r$ grid on one plane places its samples $F/r$ apart along both in-plane axes, with the outermost samples $F/(2r)$ from the boundary.
Every point on this plane is then within $F/(2r)$ of a sample along both in-plane axes and at distance zero along the normal axis.

Within radius $\rho$, each sample covers a cube of side length $2\rho$, and this cube meets a plane through the sample in a square of the same side length.
The coverage radius $\delta(\Omega)$ is thus the smallest $\rho$ at which these cubes cover $\Pi$.
A cube marks only the positions near a sample; it does not assume that the intensity inside the cube is constant or known.
Below, all sampling rates and image side lengths are positive integers.

\myparagraph{Input layouts: tri-planes and slice stack.}
Tri-planes sample each target plane on a uniform $n\times n$ grid.
Let $G_n$ denote the centers of the $n$ equal intervals of $[0,F]$:
\begin{equation*}
G_n=\left\{\left(i-\tfrac12\right)\frac Fn:i=1,\ldots,n\right\}.
\end{equation*}
Placing the grid $G_n\times G_n$ on the two in-plane coordinates of each plane gives three images with $3n^2$ pixels in total.
Let $\Omega_{\mathrm{tri}}$ denote the resulting set of sampling positions.
Every point on $\Pi$ lies within $F/(2n)$ of a sample in its plane, and a boundary point (\eg, $(0,0,c_z)$) reaches this distance, giving:
\begin{equation*}
\delta(\Omega_{\mathrm{tri}})=\frac{F}{2n}.
\end{equation*}

A slice stack along the $z$-axis instead renders $N_{\mathrm{s}}$ parallel slices at depths $\zeta_1,\ldots,\zeta_{N_{\mathrm{s}}}\in[0,F]$.
All slices sample the same grid $X\times Y$, where $X,Y\subset[0,F]$ with $|X|=|Y|=m$.
The stack thus samples the positions:
\begin{equation*}
\Omega_{\mathrm{stack}}=\{(\xi,\eta,\zeta_k):\xi\in X,\ \eta\in Y,\ k=1,\ldots,N_{\mathrm{s}}\},
\end{equation*}
which requires $N_{\mathrm{s}}m^2$ rendered pixels.
The slice depths $\{\zeta_k\}$ and the grid coordinates $X,Y$ may be spaced unevenly.
Stacks along the $x$- and $y$-axes follow by permuting the coordinates.

\myparagraph{One-dimensional covering.}
The lower bound on the pixels of a slice stack rests on covering an interval.
\begin{lemma}[Covering]
\label{lem:cover}
Let $H\subset[0,F]$ be a finite set that covers $[0,F]$ within radius $\rho>0$.
Then $|H|\ge F/(2\rho)$.
\end{lemma}
\begin{proof}
Each point in $H$ covers an interval of length at most $2\rho$, so the $|H|$ intervals have a total length of at most $2\rho|H|$.
Since they must cover $[0,F]$, we have:
\begin{equation*}
2\rho|H|\ge F,\qquad\text{and thus}\qquad |H|\ge\frac{F}{2\rho}.
\end{equation*}
\end{proof}

\begin{remark}[Pixel budget]
\label{rem:suppl_budget}
Tri-planes achieve sampling rate $r$ using $3r^2$ rendered pixels, whereas any slice stack achieving sampling rate $r$ requires at least $r^3$ rendered pixels.
\end{remark}
\begin{proof}
Set $\rho=F/(2r)$.
Since $\delta(\Omega_{\mathrm{tri}})=F/(2n)$, the sampling rate of tri-planes equals the number of pixels $n$ along each side of their images.
Tri-planes with three $r\times r$ images thus reach $\delta(\Omega_{\mathrm{tri}})=\rho$ using $3r^2$ pixels.

For a slice stack along $z$, consider the coronal plane $\mathcal C$, which spans $[0,F]$ along both $x$ and $z$.
To achieve sampling rate $r$, every point $(x,c_y,z)\in\mathcal C$ must lie within $\rho$ of some sample $(\xi_i,\eta_j,\zeta_k)\in\Omega_{\mathrm{stack}}$, which implies:
\begin{equation*}
|x-\xi_i|\le\rho\qquad\text{and}\qquad |z-\zeta_k|\le\rho.
\end{equation*}
Because $x$ and $z$ vary independently over $[0,F]$, both $X$ and $\{\zeta_1,\ldots,\zeta_{N_{\mathrm{s}}}\}$ must cover $[0,F]$ within radius $\rho$.
Applying \cref{lem:cover} to each axis gives:
\begin{equation*}
m=|X|\ge\frac{F}{2\rho}=r,\qquad N_{\mathrm{s}}\ge\frac{F}{2\rho}=r.
\end{equation*}
The stack therefore needs at least $N_{\mathrm{s}}m^2\ge r^3$ rendered pixels.
The same argument holds for stacks along $x$ or $y$, with a target plane that contains the stacking axis.
\end{proof}

\begin{corollary}[Fixed budget]
\label{cor:suppl_budget}
Given a budget of $B\ge3$ pixels, the maximum integer sampling rates achievable by tri-planes and a slice stack satisfy:
\begin{equation*}
r_{\mathrm{tri}}(B)=\left\lfloor\sqrt{B/3}\right\rfloor,
\qquad r_{\mathrm{stack}}(B)\le B^{1/3}.
\end{equation*}
\end{corollary}
\begin{proof}
For tri-planes, the largest integer $n$ with $3n^2\le B$ gives the sampling rate $n=\lfloor\sqrt{B/3}\rfloor$.
For a slice stack, \cref{rem:suppl_budget} requires $r^3\le B$, so $r\le B^{1/3}$.
\end{proof}

\myparagraph{Sufficient detail.}
Under the same pixel budget, tri-planes show the three planes through the view center in finer detail than a slice stack, because their samples lie on these planes rather than on a 3D grid.
Since every pixel lies on those planes, the pixel count of tri-planes scales quadratically with the sampling rate $r$.
By contrast, each $z$-slice of a stack intersects the coronal and sagittal planes along a single line, so achieving rate $r$ on those planes requires at least $r$ slices of $r\times r$ pixels---most falling off the three planes.
Thus, given $B$ pixels, the sampling rate scales as $\sqrt{B}$ for tri-planes versus at most $\sqrt[3]{B}$ for a stack; for instance, $3\cdot512^2$ pixels yield a rate of 512 for tri-planes but at most 92 for a stack.
Alternatively, retaining $512\times512$ slices under this budget permits only three slices, which can miss errors in the inter-slice gaps (\cref{prop:suppl_detect}, part (c)).

\subsection{Near-Minimality of Tri-Planes}
\label{sec:suppl_minimal}

Among \emph{all} layouts that sample the three planes through the view center, and not only slice stacks, tri-planes use nearly the fewest pixels.
A sample near the line where two planes meet can cover points on both planes, whereas points far from these lines need different samples on each plane.
Counting the samples that these separated regions need gives a lower bound for any layout.

\begin{proposition}[Near-minimality]
Any observation achieving sampling rate $r$ requires at least $3r^2-12r$ distinct sampling positions, and hence at least $3r^2-12r$ rendered pixels.
\end{proposition}
\begin{proof}
Set $\rho=F/(2r)$.
The coronal and sagittal planes intersect the axial plane $\mathcal A$ along the lines $y=c_y$ and $x=c_x$, respectively.
Removing a strip of half-width $2\rho$ around each of these lines leaves the region:
\begin{equation*}
\mathcal A_{\mathrm{far}}=\{(x,y,c_z)\in\mathcal A:|x-c_x|>2\rho,\ |y-c_y|>2\rho\}.
\end{equation*}
Each excluded strip has width at most $4\rho$ and length $F$.
Subtracting the areas of both strips removes their overlap at the center twice, so it gives a lower bound on the remaining area:
\begin{equation*}
\operatorname{area}(\mathcal A_{\mathrm{far}})\ge F^2-8\rho F.
\end{equation*}
We define the separated regions $\mathcal C_{\mathrm{far}}\subset\mathcal C$ and $\mathcal S_{\mathrm{far}}\subset\mathcal S$ of the other two planes in the same way.

No sample $p\in Q$ lies within $\rho$ of points in two different separated regions.
For example, every point in $\mathcal A_{\mathrm{far}}$ has $|y-c_y|>2\rho$ and every point on $\mathcal C$ has $y=c_y$, so by the triangle inequality no sample lies within $\rho$ of both along the $y$-axis.
The same argument holds for every pair among $\mathcal A_{\mathrm{far}}$, $\mathcal C_{\mathrm{far}}$, and $\mathcal S_{\mathrm{far}}$.

A sample covers an area of at most $(2\rho)^2=4\rho^2$ on a target plane, so covering $\mathcal A_{\mathrm{far}}$ alone takes at least
\begin{equation*}
\frac{\operatorname{area}(\mathcal A_{\mathrm{far}})}{4\rho^2}\ge\frac{F^2-8\rho F}{4\rho^2}=r^2-4r
\end{equation*}
distinct samples.
Since no sample serves two of the three regions, summing over them gives:
\begin{equation*}
|\Omega|\ge3(r^2-4r)=3r^2-12r.
\end{equation*}
\end{proof}

Since tri-planes use $3r^2$ pixels, no layout can save more than $12r$ pixels, \ie, a fraction of at most $4/r$, which goes to zero as $r$ grows.
Equivalently, at a fixed pixel budget $B$, no layout reaches a sampling rate above:
\begin{equation*}
r\le2+\sqrt{B/3+4}.
\end{equation*}
At the budget of tri-planes with $512\times 512$ images ($B=3\times 512^2$), the best possible layout reaches a sampling rate of at most $514.004$, against $512$ for tri-planes and at most $92.304$ for a slice stack.

\subsection{Anisotropic Voxel Spacing}
\label{sec:suppl_aniso}

The covering argument also applies to anisotropic sampling rates $(r_x,r_y,r_z)\in\mathbb{Z}_+^3$, which a scan needs when its voxel spacing differs between axes.
An observation achieves sampling rates $(r_x,r_y,r_z)$ if every target point $u\in\Pi$ has a sample $p\in\Omega$ satisfying:
\begin{equation*}
|u_x-p_x|\le\frac{F}{2r_x},\qquad
|u_y-p_y|\le\frac{F}{2r_y},\qquad
|u_z-p_z|\le\frac{F}{2r_z}.
\end{equation*}
Each sample then covers a box rather than a cube, and each plane is rendered on a rectangular grid.

\begin{proposition}[Anisotropic sampling]
Tri-planes achieve sampling rates $(r_x,r_y,r_z)$ with $r_xr_y+r_xr_z+r_yr_z$ rendered pixels, whereas any parallel slice stack on a shared Cartesian grid requires at least $r_xr_yr_z$ pixels.
\end{proposition}
\begin{proof}
Uniform grids of $r_x\times r_y$, $r_x\times r_z$, and $r_y\times r_z$ samples on the axial, coronal, and sagittal planes meet the bound along each coordinate with $r_xr_y+r_xr_z+r_yr_z$ pixels.

For a slice stack along $z$, applying \cref{lem:cover} along each axis shows that covering the coronal plane requires $|X|\ge r_x$ and $N_{\mathrm{s}}\ge r_z$, and covering the sagittal plane requires $|Y|\ge r_y$.
The stack thus needs:
\begin{equation*}
N_{\mathrm{s}}|X||Y|\ge r_zr_xr_y.
\end{equation*}
Stacks along $x$ or $y$ follow by symmetry.
\end{proof}

The lower bound of a slice stack exceeds the pixels of tri-planes by the factor:
\begin{equation*}
\frac{r_xr_yr_z}{r_xr_y+r_xr_z+r_yr_z}
=\left(\frac1{r_x}+\frac1{r_y}+\frac1{r_z}\right)^{-1}
\ge\frac{\min(r_x,r_y,r_z)}3,
\end{equation*}
which equals the isotropic factor $r/3$ when $r_x=r_y=r_z=r$ and exceeds 1, so that tri-planes save pixels, whenever $1/r_x+1/r_y+1/r_z<1$.

\section{Agent Details}
\label{sec:suppl_agent}

This section details the harness that \mainref{sec:harness} summarizes, and \cref{sec:suppl_prompts} gives the prompts of all MLLM calls (\ref{prompt:seed_start} to \ref{prompt:search_seed}).
\cref{tab:suppl_loop} follows the steps of \mainref{fig:workflow} and lists what the agent does after each result.
The three subsections follow the three parts of \mainref{sec:harness}: initialization (\cref{sec:suppl_init}), click selection from candidates (\cref{sec:suppl_click}), and multi-view verification with the acceptance rule and the call budget (\cref{sec:suppl_verify}).
\cref{sec:suppl_verify_theory} then gives the theory behind multi-view verification.

\begin{table}[!htb]
\centering
\newcommand{\stepmark}[1]{\smash{\circled{#1}}}
\newcommand{\byword}[1]{\textcolor{black!55}{#1}}
\newcommand{\nextpass}{\textcolor[HTML]{0A8F2D}{View passes}}
\newcommand{\nextstop}{\textcolor[HTML]{E5484D}{Stop}}
\caption{\textbf{Outcomes of each step in \mainref{fig:workflow}.} Green marks a pass, and red (Stop) means that the agent stops without accepting the prediction. Each ``Judge again'' uses one of the two retries of the view, and the agent stops when no retry is left.}
\label{tab:suppl_loop}
\footnotesize
\setlength{\tabcolsep}{4pt}
\renewcommand{\arraystretch}{1.15}
\begin{tabularx}{\linewidth}{@{}>{\raggedright\arraybackslash}p{0.165\linewidth}>{\raggedright\arraybackslash}p{0.075\linewidth}>{\raggedright\arraybackslash}X>{\raggedright\arraybackslash}p{0.235\linewidth}@{}}
\toprule
\textbf{Step} & \textbf{By} & \textbf{Result} & \textbf{Next} \\
\midrule
(b) Judge the view & \byword{MLLM} & Accept & \nextpass \\
 & & Error with a box & Step \stepmark{1} \\
 & & Unsure & Judge again \\
 & & Wrong structure on the working, initial, or centroid view & \nextstop \\
 & & Wrong structure on any other view & Judge again, as if unsure \\
\midrule
\stepmark{1} Check the box & \byword{Harness} & Overlay covers less than $5\%$ of an extra-part box or more than $50\%$ of a missing-part box & \nextpass \\
 & & Otherwise & Confirm the error \\
\midrule
\stepmark{1} Confirm the error & \byword{MLLM} & Withdrawn & \nextpass \\
 & & Extra part covers at least $80\%$ of a large cross-section of the prediction & \nextstop \\
 & & Extra part covers less than $10\%$ of the prediction's cross-section on that plane & \nextpass \\
 & & Three corrections already committed & \nextstop \\
 & & Otherwise & Step \stepmark{2} \\
\midrule
\stepmark{2} Place candidates & \byword{Harness} & No valid candidate & Judge again \\
 & & Up to five points on the required side of the prediction boundary (\mainref{eq:candidates}) & Step \stepmark{3} \\
\midrule
\stepmark{3} Pick a point & \byword{MLLM} & Too little budget left to pick a point and verify the corrected prediction & \nextstop \\
 & & Declines to pick, or returns an invalid number & Judge again \\
 & & One point & Step \stepmark{4} \\
\midrule
\stepmark{4} Check the proposal & \byword{Both} & Valid large correction that the remaining budget cannot verify & \nextstop \\
 & & Rejected by the basic checks, the target check, or the before-and-after comparison & Undo the click and return to \stepmark{3} without this point; judge again once no point is left \\
 & & Passes all checks & Commit it and rebuild the views \\
\midrule
(c) After the last view & \byword{Harness} & Every earlier click still holds & \textcolor[HTML]{0A8F2D}{Accept the prediction} \\
 & & An earlier click no longer holds & \nextstop \\
\bottomrule
\end{tabularx}
\end{table}

\subsection{Initialization}
\label{sec:suppl_init}

\myparagraph{Initial view.}
The harness centers the initial view on one component $P_0$ of the localization prior $P$.
Among the components that hold at least $10\%$ of the voxels of the largest component, it takes the one with the highest average probability, so that a small but confident fragment cannot decide where the view goes.
The view center $c_0$ is the deepest point of $P_0$, \ie, the voxel of $P_0$ with the largest distance $d$ to its boundary:
\begin{equation}
c_0=\arg\max_{x\in P_0}d(x,\partial P_0).
\label{eq:suppl_init}
\end{equation}
Unlike the center of a bounding box, $c_0$ always lies inside $P_0$, even for a curved structure.
The side length $F_0$ is the smallest of the presets $64$, $128$, and $256$\,mm at which $P_0$ fills at most $65\%$ of the view along each axis.
This fill is measured by the smallest box that is centered at $c_0$ and encloses $P_0$ (\cref{fig:suppl_hparam}b tests other fill values).
When no preset is large enough, $F_0$ is $256$\,mm.

\myparagraph{Initial prediction.}
On this view, $\fseg$ produces two proposals: one grown from $P_0$, which $\fseg$ receives as an initial segmentation, and one grown from a single point.
For the point proposal, the harness tries up to five points, the first at $c_0$ and the other four inside the bounding box of $P_0$.
It keeps the proposal with the highest Dice against $P_0$, without calling the MLLM.
The MLLM then compares the two proposals on the same view in three sampled votes, each of which chooses one proposal or rejects both.
It sees only the two proposals and not $P$, so that a wrong prior cannot bias its choice.
For organs, muscles, and vessels whose $P_0$ has at least 100 voxels, the selected proposal $\tilde M$ must also pass an overlap check with $P_0$:
\begin{equation}
\frac{|\tilde M\cap P_0|}{\min(|\tilde M|,|P_0|)}\ge0.05.
\label{eq:suppl_prior_overlap}
\end{equation}
Dividing by the smaller size lets a proposal larger than $P_0$ pass as long as it covers $P_0$.
This check and the uncovered-prior view (\cref{tab:suppl_views}) apply only to these three ROI types, because for the others a difference from $P_0$ does not mean that a proposal is wrong.
For bones, $\floc$ often localizes a neighboring vertebra, and tumors and lymph nodes are often thin or split into several pieces.

The selected proposal becomes $M_0$ once it passes the checks in \cref{sec:suppl_click}.
When the MLLM selects the proposal from $P_0$ but this proposal fails the checks, three sampled votes choose among the sampled points themselves, drawn as numbered rings on the tri-planes of $v_0$.
The MLLM sees only the points, not their proposals, and the proposal from the chosen point then goes through the same checks.
When the votes reject both proposals, the harness moves to a view with another preset side length.
Initialization tries at most six proposals, and if none of them is committed, the agent continues with the ROI search below.

\myparagraph{ROI search without a localization prior.}
The ROI search runs when $\floc$ finds nothing or when initialization ends without a committed prediction, so that a failed localization does not end the annotation.
It restarts the loop with an empty prediction and a new call budget and does not use $P$.
The MLLM first places a rough point on the target in 16 axial slices that span the whole scan, and then refines this point in 16 axial slices within a $96$\,mm window around it.
On a $128$\,mm view centered at the refined point, the MLLM places the first positive click directly, without candidates.
The proposal from this click must pass the checks in \cref{sec:suppl_click}; when it fails, the MLLM places another click on the same view, within the limit of six trials.
The loop then continues from the committed prediction.
If the MLLM finds no target in the whole-scan slices, or no proposal is committed within the six trials, the agent stops without accepting a prediction.

\subsection{Click Selection from Candidates}
\label{sec:suppl_click}

\myparagraph{Confirming an error (step~\circled{1}).}
For a box that the overlay does not contradict (the box check in \cref{tab:suppl_loop}), the harness renders a view centered on the box, with a crosshair and the measured size of the boxed region (\ref{prompt:verify}).
The MLLM must name the tissue at the crosshair before it confirms or withdraws the error, and the prompt tells it to withdraw when in doubt.
The agent therefore edits only errors that the MLLM can confirm.

\myparagraph{Checking a proposal (step~\circled{4}).}
The harness commits a proposal only after it passes the basic checks and the MLLM confirms it.
The basic checks reject bad proposals, \eg, a proposal that is empty, leaves the prediction unchanged, breaks a current or past click, or covers more than $20\%$ of the volume.
The MLLM then confirms, with at least medium confidence, that the proposal lies on the correct target.
We call a correction large when it changes more than $20\%$ of the union of the two predictions.
A correction is also large when it adds or removes a piece that holds at least $5\%$ of the changed voxels and has its centroid at least $F_0/2$ from $c_0$ along an axis.
For a large correction, the MLLM also compares the predictions before and after the click and checks that the click fixes the marked error without spreading into wrong regions.

\subsection{Multi-View Verification}
\label{sec:suppl_verify}

\myparagraph{Verification views.}
\cref{tab:suppl_views} lists the views that the agent checks, which depend on the current prediction, the initialization, and the interaction history.
Verification normally uses three coverage views (initial, centroid, and last change or surface), and five coverage views after a large correction or when the prediction has an extra component that is far away, small, or large.
An extra component is a 6-connected component that does not contain the first click.
It is far away when its centroid lies at least $F_0/2$ from $c_0$ along an axis, small when it has fewer than $10\%$ of the voxels of the largest component, and large when it holds at least $5\%$ of the voxels of the prediction.
In both this rule and the rule for a large correction, $F_0$ and $c_0$ belong to the first initial view and stay fixed even when initialization moves to another preset; after the ROI search, $c_0$ is the searched location and $F_0=128$\,mm.

The uncovered-prior and overview views are added to the coverage views, and a view that lands within two voxels of an earlier verification view is dropped, since it would repeat that check.
When a correction is committed, the agent rebuilds the views and keeps only the verdicts of initial and centroid views whose images did not change, so that every decision concerns the latest prediction.
A ``wrong-structure'' verdict stops the agent only on the working, initial, and centroid views, since these views show the target structure best.
The uncovered-prior view checks only whether the outlined tissue belongs to the target, and it rejects nothing by itself.

\begin{table}[!htb]
\centering
\caption{\textbf{Views that the agent judges.} The first two come before verification, and the others are verification views.}
\label{tab:suppl_views}
\footnotesize
\setlength{\tabcolsep}{4pt}
\renewcommand{\arraystretch}{1.15}
\begin{tabularx}{\linewidth}{@{}l>{\raggedright\arraybackslash}X>{\raggedright\arraybackslash}X@{}}
\toprule
\textbf{View} & \textbf{Location} & \textbf{When used} \\
\midrule
Working & Location of the most recent positive click, or $c_0$ & Until a large correction has been committed \\
Negative click & Location of the negative click & After a negative correction made during verification, until a large correction has been committed \\
\midrule
Initial & Initial view $v_0$ & Always \\
Centroid & Centroid of the component of the prediction that contains the first click & Unless it lies within two voxels of the initial view \\
Last change / surface & Centroid of the largest added or removed component, or the last click; before the first correction, the centroid of the largest cross-section of the prediction through a surface point, moved to the nearest voxel of the prediction & Unless it lies within two voxels of an earlier verification view \\
Added and removed regions & Separate views centered on the largest added and largest removed components & With five coverage views, after a change of that type \\
Separate piece & Centroid of an extra component & With five coverage views, when such a component exists \\
Uncovered prior & Centroid of the largest component of $P_0\setminus M_t$, with the uncovered prior outlined in blue & Organs, muscles, and vessels, when at least $30\%$ of $P_0$ is uncovered and this component contains at least 100 voxels \\
Overview & Centroid of the whole prediction, using the next larger preset side length & When a larger preset is available \\
\bottomrule
\end{tabularx}
\end{table}

\myparagraph{Acceptance rule.}
The MLLM judges the views of $\mathcal{V}(M_t)$ one at a time in a fixed order.
The first view in which the MLLM confirms an error starts a correction at once, and the later views are not judged on $M_t$.
The MLLM confirms an error when it marks the error with a box and keeps the box through the box check and the confirmation view (\cref{tab:suppl_loop}).
After a small correction, the MLLM first judges the working view once more, or the negative-click view if the correction was negative and made during verification; the harness then rebuilds the views.
After a large correction, the harness rebuilds the views at once.
Let $\chi_t=1$ if every view in $\mathcal{V}(M_t)$ has passed on $M_t$, counting the verdicts kept for the initial and centroid views, and $\chi_t=0$ otherwise.
Let $\mathcal{K}^+_t$ and $\mathcal{K}^-_t$ denote the voxel positions of the positive and negative clicks up to step $t$.
The agent accepts the prediction at the first step $T_{\mathrm{acc}}$ at which every verification view passes and no later click has undone an earlier one:
\begin{equation}
T_{\mathrm{acc}}=\min\bigl\{t:\ \chi_t=1,\ \mathcal{K}^+_t\subseteq M_t,\ \mathcal{K}^-_t\cap M_t=\emptyset\bigr\}.
\label{eq:suppl_stop}
\end{equation}
The two click conditions require every positive click to stay inside the prediction and every negative click to stay outside it.
If $T_{\mathrm{acc}}<\infty$, the agent stops at $T=T_{\mathrm{acc}}$ and outputs $M=M_T$; otherwise, it stops without accepting the prediction in one of the cases of \cref{tab:suppl_loop}.

\myparagraph{Views after a correction.}
By \mainref{rem:verify}, a new error region after a correction must intersect the region $\Delta_t$ that the correction changes, which can be large, split into several pieces, or far from the click.
The harness centers the last-change view on the largest added or removed piece of $\Delta_t$.
After a large correction, it adds separate views on the largest added and the largest removed pieces, and the MLLM compares the predictions before and after the click (\cref{sec:suppl_click}).
The smaller pieces of $\Delta_t$ are checked wherever the planes of other views pass through them.
Likewise, an error region created by an earlier click intersects the change made by that click, and the next round centers views on this change.
If that round stops earlier at an error in another view, later rounds check this change wherever their views pass through it.
A view whose planes miss $\Delta_t$ has the same tri-planes after the correction, so its earlier verdict still applies.
The harness keeps such verdicts only for the initial and centroid views whose images did not change.
The MLLM judges the other rebuilt views again, since their centers depend on the updated prediction.

\myparagraph{Call budget and stopping.}
Each run of the loop (including the rerun for the ROI search) allows up to 26 MLLM calls and three committed corrections (\cref{fig:suppl_hparam}a tests other budget limits).
Since the rerun starts with a new budget, an ROI that goes to the ROI search can use more than 26 MLLM calls in total.
Before each click, the harness reserves enough of this budget to verify the resulting prediction, so the agent stops with the current prediction instead of committing a correction that it cannot verify.
Whenever the agent stops without accepting the prediction, it records the reason and any views it could not confirm, and it outputs the current prediction when one exists.

\section{Theory of Multi-View Verification}
\label{sec:suppl_verify_theory}

This section gives the theory behind multi-view verification (\mainref{sec:harness}); as in \cref{sec:suppl_theory}, the symbols that the proofs define are local to them.

\subsection{Where New Error Regions Can Appear}
\label{sec:suppl_rem_verify}

We prove \mainref{rem:verify}.
A voxel is wrong in a prediction $M$ when $M$ and $M^\ast$ disagree on it.
Let $d(x,\partial M^\ast)$ denote the distance from a voxel $x$ to the boundary of $M^\ast$, and let $E(M)$ denote the wrong voxels of $M$ that lie farther than a distance $\tau$ from this boundary:
\begin{equation*}
E(M)=\{x\in M\triangle M^\ast : d(x,\partial M^\ast)>\tau\}.
\end{equation*}
An error region of $M$ is a connected component of $E(M)$ under 26-connectivity whose volume is at least a fraction $\gamma$ of the volume of $M^\ast$.
The set $\Delta_t=M_{t+1}\triangle M_t$ holds the voxels that step $t$ changes, and it is empty unless step $t$ commits a click.

\begin{proof}
A voxel outside $\Delta_t$ lies either in both $M_t$ and $M_{t+1}$ or in neither, so it is wrong in $M_{t+1}$ exactly when it is wrong in $M_t$.
Its distance to the boundary of $M^\ast$ does not depend on the prediction, so $E(M_{t+1})$ and $E(M_t)$ contain the same voxels outside $\Delta_t$.
Let $C$ be an error region of $M_{t+1}$ that does not intersect $\Delta_t$.
Then $C\subseteq E(M_t)$, and since $C$ is connected, it lies inside one connected component $C'$ of $E(M_t)$.
The volume of $C'$ is at least that of $C$, hence at least a fraction $\gamma$ of the volume of $M^\ast$, so $C'$ is an error region of $M_t$ that contains $C$.
For the second part, the renderer $R$ draws each of the tri-planes of a view from $V$ and from the cross-section of the prediction on that plane.
The volume $V$ stays fixed, and $M_{t+1}$ and $M_t$ agree on every plane that contains no voxel of $\Delta_t$.
For any view $v$ whose three planes miss $\Delta_t$, the harness therefore renders the same three images for $M_t$ and $M_{t+1}$, \ie, $R(V,M_{t+1};v)=R(V,M_t;v)$.
\end{proof}

\subsection{Views Centered at a Centroid}
\label{sec:suppl_centroid}

The last-change view and most other verification views are centered at the centroid of a connected component, such as a piece of the prediction or the largest region that the last click changed.
The lemma below shows that such a view always cuts its component, so the harness can use the raw centroid without moving it onto the prediction.

\begin{lemma}[Planes through a centroid]
\label{lem:suppl_centroid}
For a view centered at the centroid of a connected component, each of its three planes passes through that component, even when the center lies outside it.
\end{lemma}

\begin{proof}
Let $\Gamma$ be a connected component of a voxel set, under any connectivity in which neighboring voxels differ by at most one layer along each axis, such as 6-, 18-, or 26-connectivity.
We treat the axial axis; the coronal and sagittal axes follow in the same way.
Let $z(w)$ denote the axial layer of a voxel $w$, and let $z_{\min}$ and $z_{\max}$ be the first and last layers that $\Gamma$ occupies.
Since $\Gamma$ is connected, a chain of neighboring voxels in $\Gamma$ leads from a voxel in layer $z_{\min}$ to a voxel in layer $z_{\max}$.
The layer changes by at most one from each voxel of the chain to the next, so the chain passes through every layer in between, and $\Gamma$ occupies every layer from $z_{\min}$ to $z_{\max}$.
The axial coordinate of the centroid,
\begin{equation*}
\bar z=\frac{1}{|\Gamma|}\sum_{w\in \Gamma}z(w),
\end{equation*}
is an average of values in $[z_{\min},z_{\max}]$, so it lies in this interval as well.
The renderer places the axial plane at the layer nearest to $\bar z$, which lies between $\lfloor\bar z\rfloor$ and $\lceil\bar z\rceil$; both lie in $[z_{\min},z_{\max}]$ because $z_{\min}$ and $z_{\max}$ are integers.
The axial plane therefore lies in a layer that $\Gamma$ occupies, so it meets $\Gamma$.
\end{proof}

\noindent The center itself need not lie in $\Gamma$: for a ring-shaped or curved component, the centroid can fall in the hole while each plane still cuts the ring.
The guarantee concerns the planes in 3D, not the cropped images.
When $\Gamma$ extends beyond the view, its voxels on a plane can lie outside the rendered image, so the lemma does not imply that $\Gamma$ appears in every image of the tri-planes.
The lemma covers the centroid, last-change, added- and removed-region, separate-piece, and uncovered-prior views.
The overview view falls outside it, since its center is the centroid of the whole prediction and can lie between its pieces.
The surface view is moved to the nearest voxel of the prediction, and the initial view is centered at the deepest point of $P_0$, so their planes meet the prediction and $P_0$, respectively, by construction.

\subsection{Chance of Missing an Error Region}
\label{sec:suppl_coverage}

We bound the chance that several views all miss the error regions of a prediction, because an error region can lie off the planes of every view that passes.
The harness chooses its views by fixed rules, so this chance comes from what the agent does not know: we treat the ground-truth mask $M^\ast$ as random given the history $\mathcal{H}$ of everything that the agent has observed.
A view \emph{shows} an error region when one of its three images contains a voxel of this region at the image resolution, whether or not the MLLM notices it; the judgment of the MLLM enters in \cref{cor:suppl_cost}.

\begin{proposition}[Chance of missing an error region]
\label{prop:suppl_coverage}
Let the current prediction contain an error region, and suppose that at least $k$ of the views judged on it each show an error region with probability at least $\pcov\in(0,1]$, given $\mathcal{H}$ and given that the views before it showed none.
Then all views judged on it miss the error regions with probability at most $(1-\pcov)^k$.
\end{proposition}

\begin{proof}
Let $J$ be the number of views, and let $A_j$ be the event that none of the first $j$ views shows an error region, with $A_0$ certain.
Each factor $\Pr(A_j\mid A_{j-1},\mathcal{H})$ is at most $1$, and by assumption it is at most $1-\pcov$ for at least $k$ of the views, so the chain rule gives:
\begin{equation*}
\Pr(A_{J}\mid\mathcal{H})=\prod_{j=1}^{J}\Pr(A_j\mid A_{j-1},\mathcal{H})\le(1-\pcov)^k.
\end{equation*}
\end{proof}

\noindent The condition asks each of these $k$ views to keep a chance of showing an error region after the earlier views have missed it.
More views alone do not meet the condition, since views close to each other can leave the same region unseen.
A view with the same images as a view that the MLLM has already judged, including a view whose verdict the harness keeps from an earlier round, adds no new chance and does not count toward $k$.
The harness therefore spreads its views over different parts of the prediction, the change, and the prior.
\cref{sec:suppl_abl_harness} measures $\pcov$ for the views of our harness.

\subsection{Outcome and Cost of an Idealized Run}
\label{sec:suppl_idealized}

We count judgments of views, \ie, MLLM calls that judge the working view or a verification view, and let $K$ be the largest number of views in $\mathcal{V}(M_t)$.
Let $R_1,\dots,R_{N_{\mathrm{err}}}$ denote the error regions of $M_0$ (\cref{sec:suppl_rem_verify}).
Since $M^\ast$ is random given the history (\cref{sec:suppl_coverage}), so is $N_{\mathrm{err}}$.
We consider an idealized run that no rule ends early, \ie, we ignore the call budget, the cap of three corrections, and the other stops without acceptance in \cref{tab:suppl_loop}.
\mainref{cor:verify} in the main paper states the first part of the corollary below.

\begin{corollary}[Reaching a prediction without error regions]
\label{cor:suppl_cost}
Assume that (A1)~the MLLM judges every view correctly, \ie, it confirms an error in a view exactly when the view shows an error region, and it is never unsure; (A2)~in every round, at least $k$ of the verification views meet the condition of \cref{prop:suppl_coverage} for a lower bound $\pcov$ that holds in all rounds; and (A3)~each error that the MLLM confirms leads to a committed correction after which every error region lies inside an error region of the prediction before it, and at least one of $R_1,\dots,R_{N_{\mathrm{err}}}$ that contained an error region before the correction contains none after it.
In an idealized run, the agent then accepts a prediction without error regions within $N_{\mathrm{err}}$ corrections with probability at least $1-\mathbb{E}[N_{\mathrm{err}}](1-\pcov)^k$, and the run makes at most $(N_{\mathrm{err}}+1)(K+1)$ judgments in every case.
\end{corollary}

\begin{proof}
We call the judgments between two consecutive corrections a round, and likewise the judgments before the first correction and after the last one.
By (A1), a round on a prediction with error regions ends in a correction when one of its views shows an error region, and in a wrong acceptance otherwise; by \cref{prop:suppl_coverage} and (A2), the second case has probability at most $(1-\pcov)^k$.
By (A3), every error region after a correction lies inside an error region before it.
Since $R_1,\dots,R_{N_{\mathrm{err}}}$ are disjoint, induction shows that every error region of a later prediction lies inside one of them and that a region $R_i$ without error regions never gets one back.
Every correction clears at least one more $R_i$, so at most $N_{\mathrm{err}}$ rounds see a prediction with error regions.
Given the history before it, each of these rounds ends in a wrong acceptance with probability at most $(1-\pcov)^k$, so summing over the rounds bounds the chance of a wrong acceptance by $(1-\pcov)^k$ times the expected number of these rounds, which is at most $\mathbb{E}[N_{\mathrm{err}}]$.
Without a wrong acceptance, every $R_i$ is cleared, so no error region remains, and every view of the last round passes by (A1).
The click conditions of \cref{eq:suppl_stop} also hold, since the basic checks keep every earlier click on its side of each committed prediction (\cref{sec:suppl_click}).
The agent therefore accepts a prediction without error regions.
For the cost, each round starts with at most one judgment of the working view, namely the first judgment of the run or the re-check after a small correction, and then judges at most $K$ verification views, since the first error that the MLLM confirms ends it.
The run has at most $N_{\mathrm{err}}+1$ rounds, so it makes at most $(N_{\mathrm{err}}+1)(K+1)$ judgments.
\end{proof}

\noindent With $\pcov=1$, \ie, when the views of every round show a remaining error region, the agent accepts a prediction without error regions with certainty, which is the deterministic case.
In our harness, $K\le7$, because a round checks at most five coverage views plus the uncovered-prior and overview views.
An idealized run therefore makes at most $8(N_{\mathrm{err}}+1)$ judgments, regardless of the size of the target.
Besides these judgments, each correction needs a bounded number of other decisions: confirming the error, picking a candidate, and checking the proposal.
The total number of decisions is thus also at most linear in $N_{\mathrm{err}}$.
A real run may stop without accepting a prediction if fixing its error regions takes more than the budget of 26 MLLM calls or the cap of three corrections (\cref{sec:suppl_verify}).
We expect this case to be rare, since most ROIs finish within about ten calls (\cref{fig:suppl_hparam}a).

\myparagraph{When the assumptions fail.}
If (A1) or (A3) fails, a run can go two ways.
It can cost more: an unsure verdict costs up to two retries, a rejected proposal leads to further candidates and a new judgment of the view, and a correction that fixes nothing adds a round.
It can also end too early with a wrong prediction, when no view shows a remaining error region or the MLLM misses one that a view shows.
Suppose that, given the history, the MLLM confirms an error in a view that shows an error region with probability at least $s\in(0,1]$, every error that it confirms lies in an error region, and the correction that follows breaks (A3) with probability at most $\eta$.
The argument of \cref{cor:suppl_cost} then gives:
\begin{equation*}
\Pr(\text{the agent accepts a prediction without error regions})\ge1-\mathbb{E}[N_{\mathrm{err}}]\bigl[(1-s\,\pcov)^k+\eta\bigr].
\end{equation*}
More views that meet the condition of \cref{prop:suppl_coverage} shrink the first term, and a segmentation model that keeps earlier corrections keeps the second term small.
The false accepts in our experiments count the runs that end too early with a clearly wrong prediction (\mainref{tab:ablation}).

\section{Benchmark Composition}
\label{sec:suppl_bench}

\myparagraph{Design principles.}
Our benchmark favors breadth: many ROI types, datasets, and imaging modalities, with a few cases from each source.
We sample at most 15 cases from each dataset and keep every valid ROI that appears in the volume within a reasonable field of view.
CVPR-BiomedSegFM~\cite{ma2026medsegfm} splits SegRap2023 into a contrast-enhanced subset (14 volumes) and a non-contrast subset (10 volumes), so this limit applies to each subset.
To cover more anatomical structures and scanner protocols, we also include held-out cases from datasets used to train $\floc$ or $\fseg$.

\myparagraph{Training exposure groups.}
We group the dataset families by which of $\floc$ and $\fseg$ have seen them in training.
The localization model $\floc$ was trained on the text-guided training split of CVPR-BiomedSegFM~\cite{ma2026medsegfm}, which spans 68 datasets (24 CT, 28 MR, 1 PET, 4 ultrasound, and 11 microscopy).
For $\fseg$, we use the training and held-out lists in Tables A1 and A2 of~\cite{isensee2025nninteractive}, which name 120 training datasets.
The two conditions give four exposure groups: C1 (both), C2 ($\fseg$ only), C3 ($\floc$ only), and C4 (neither).
For the four lesion datasets in C4 (MSWAL, DOLCHID, Pituitary-MRI, and UT-EndoMRI), we also checked that no training dataset of $\fseg$ annotates the same lesion types.

\myparagraph{Benchmark statistics.}
The benchmark starts from 706 candidate ROIs in 262 volumes from 35 datasets.
We exclude 16 of the 87 ROIs from TotalSegmentator MRI because their ground-truth masks are unusable (left and right labels swapped, or the ROI cut off by the field of view).
This exclusion leaves 690 ROIs for evaluation (\cref{fig:suppl_bench,tab:suppl_bench}).
By training exposure, they split into C1 (288), C2 (45), C3 (192), and C4 (165); by modality, into CT (368), MR (319), and PET (3); by size, into 260 small and 430 large; and by ROI type, into 398 organs, 222 lesions, 34 vessels, 10 bones, 15 lymph nodes, and 11 muscles.

\begin{figure}[!htb]
\centering
\includegraphics[width=0.75\linewidth]{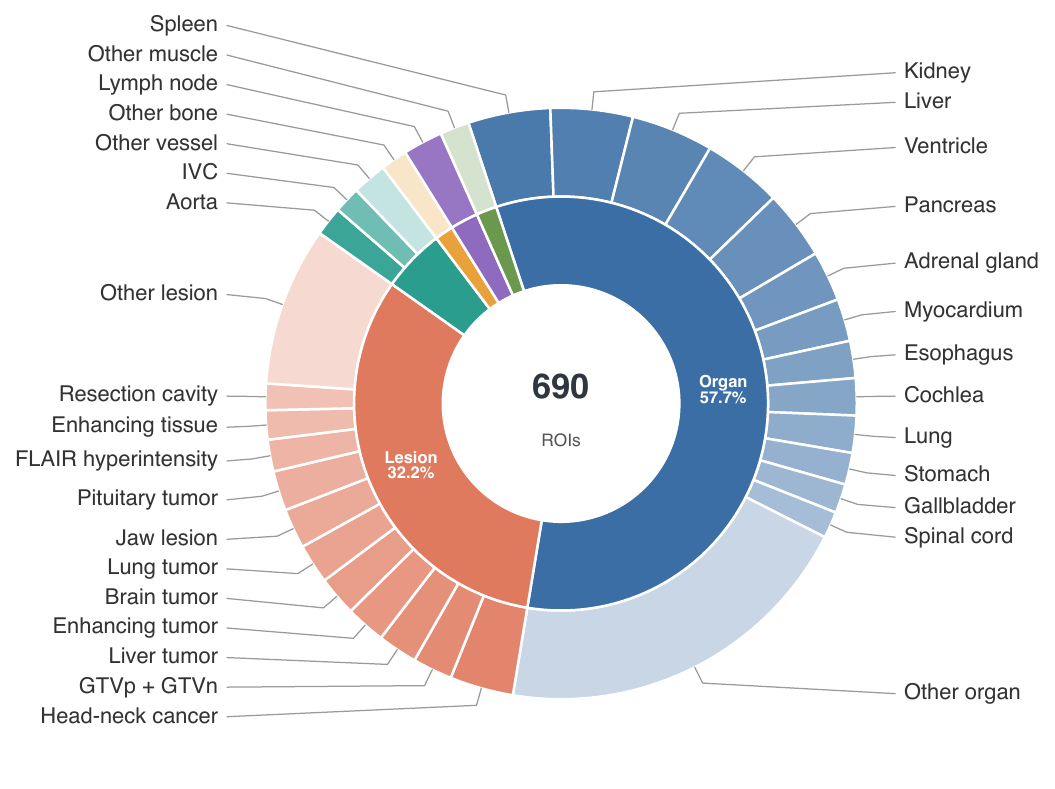}
\caption{\textbf{Distribution of the evaluated ROIs in the benchmark.}
Inner ring: ROI types; outer ring: the ROIs of each type, with left and right counterparts merged and ROIs with fewer than 10 instances grouped as ``other''.
IVC: inferior vena cava; GTVp + GTVn: primary and nodal gross tumor volume.}
\label{fig:suppl_bench}
\end{figure}

\begin{table}[!htb]
\centering
\caption{\textbf{Data sources of the benchmark.}
The sources are grouped by which of $\floc$ and $\fseg$ have seen them in training: C1, both (288 ROIs); C2, $\fseg$ only (45); C3, $\floc$ only (192); and C4, neither (165).
Vol.: volumes.
For MSD Prostate / PROSTATEx, CVPR-BiomedSegFM~\cite{ma2026medsegfm} releases the masks under the class name ``transition zone'', but they cover the whole gland.}
\label{tab:suppl_bench}
\footnotesize
\setlength{\tabcolsep}{4pt}
\renewcommand{\arraystretch}{1.0}
\begin{tabular}{@{}lllllrr@{}}
\toprule
\textbf{Group} & \textbf{Source} & \textbf{Mod.} & \textbf{Region} & \textbf{Annotated ROIs} & \textbf{Vol.} & \textbf{ROIs} \\
\midrule
\multirow{18}{*}{C1} & AMOS~\cite{ji2022amos} & CT & abdomen & abdominal organs and vessels & 3 & 44 \\
 & AMOS-MRI~\cite{ji2022amos} & MR & abdomen & abdominal organs and vessels & 3 & 39 \\
 & AbdomenAtlas~\cite{li2024abdomenatlas} & CT & abdomen & abdominal organs and vessels & 3 & 34 \\
 & TotalSegmentator MRI~\cite{dantonoli2025totalsegmri} & MR & whole body & organs, muscles, bones, vessels & 3 & 71 \\
 & CHAOS~\cite{kavur2021chaos} & MR & abdomen & liver, kidneys, spleen (T1, T2) & 4 & 16 \\
 & BraTS~\cite{baid2021brats,menze2015brats,bakas2017brats,karargyris2023medperf} & MR & brain & glioma subregions (T1n, T1c, T2w, T2f) & 12 & 41 \\
 & crossMoDA~\cite{dorent2023crossmoda} & MR & brain & vestibular schwannoma, cochlea & 3 & 12 \\
 & WMH~\cite{kuijf2019wmh} & MR & brain & white matter hyperintensities & 3 & 3 \\
 & ACDC~\cite{bernard2018acdc} & MR & heart & LV, RV, myocardium & 3 & 5 \\
 & MSD Heart~\cite{antonelli2022msd} & MR & heart & left atrium & 2 & 2 \\
 & ATM'22~\cite{zhang2023atm22} & CT & thorax & airway tree & 3 & 3 \\
 & AVT~\cite{radl2022avt} & CT & thorax--abdomen & aortic vessel tree & 3 & 3 \\
 & COVID-19-20~\cite{roth2022covid1920} & CT & thorax & COVID-19 lesions & 3 & 3 \\
 & COVID-19-CT-Seg~\cite{ma2021covidbench} & CT & thorax & COVID-19 lesions & 1 & 1 \\
 & NSCLC-Radiomics~\cite{aerts2014radiomics} & CT & thorax & left, right lung & 1 & 2 \\
 & NCI-ISBI 2013~\cite{bloch2015nciisbi} & MR & pelvis & prostate (whole gland) & 3 & 3 \\
 & MSD Prostate / PROSTATEx~\cite{antonelli2022msd,armato2018prostatex} & MR & pelvis & prostate (whole gland) & 3 & 3 \\
 & autoPET~\cite{gatidis2022autopet} & PET & whole body & FDG-avid lesions & 3 & 3 \\
\midrule
C2 & WORD~\cite{luo2022word} & CT & abdomen & liver, pancreas, spleen & 15 & 45 \\
\midrule
\multirow{8}{*}{C3} & HaN-Seg~\cite{podobnik2023hanseg} & CT & head \& neck & organs at risk & 4 & 116 \\
 & SegRap2023~\cite{luo2025segrap} & CT & head \& neck & primary GTV (contrast, non-contrast) & 24 & 24 \\
 & HNTS-MRG 2024~\cite{wahid2025hntsmrg} & MR & head \& neck & primary and nodal GTV (T2) & 15 & 15 \\
 & LNQ 2023~\cite{dorent2025lnq,idris2024lnqdata} & CT & thorax & mediastinal lymph nodes & 15 & 15 \\
 & LCTSC~\cite{yang2018lctsc} & CT & thorax & thoracic organs at risk & 3 & 15 \\
 & Adrenal-ACC-Ki67~\cite{moawad2023adrenalacc,ahmed2020adrenalki67} & CT & abdomen & adrenocortical carcinoma & 3 & 3 \\
 & ISLES 2022~\cite{hernandez2022isles} & MR & brain & ischemic stroke lesions & 3 & 3 \\
 & QIN-PROSTATE~\cite{fedorov2018qinprostate} & MR & pelvis & prostate & 1 & 1 \\
\midrule
\multirow{8}{*}{C4} & MS-CMRSeg~\cite{zhuang2019mvmm,wu2023minimizing,gao2023bayeseg} & MR & heart & LV, RV, myocardium (LGE) & 15 & 45 \\
 & RHUH-GBM~\cite{cepeda2023rhuhgbm} & MR & brain & enhancing tumor, whole tumor & 15 & 30 \\
 & CRLM~\cite{simpson2024crlm} & CT & abdomen & colorectal liver metastases & 15 & 15 \\
 & RIDER Lung CT~\cite{zhao2009rider,wee2020riderlabels} & CT & thorax & primary lung tumor (GTV) & 15 & 15 \\
 & MSWAL~\cite{wu2025mswal} & CT & abdomen & gallstones, kidney stones & 15 & 15 \\
 & DOLCHID~\cite{huang2026dolchid} & CT & jaw & odontogenic lesions & 15 & 15 \\
 & Pituitary-MRI~\cite{pandit2025pituitary} & MR & sellar region & pituitary neuroendocrine tumor & 15 & 15 \\
 & UT-EndoMRI~\cite{liang2025utendomri} & MR & pelvis & endometriomas, pelvic cysts & 15 & 15 \\
\midrule
\multicolumn{5}{@{}l}{Total: 35 sources; CT 368, MR 319, PET 3} & 262 & 690 \\
\bottomrule
\end{tabular}
\end{table}

\myparagraph{Data splits and leakage audit.}
For any dataset that either model trained on, we sample only from its validation or test cases, whereas datasets absent from both training lists (\eg, RHUH-GBM and CRLM) may contribute cases from any split.
We take the datasets of C1 and C3 from CVPR-BiomedSegFM, drawing from its 3D validation split (held out from $\floc$ training by task filename) and excluding all FLARE22 training cases~\cite{ma2024flare22}.
We take the datasets of C2 and C4 from their original public releases, including the collections of The Cancer Imaging Archive~\cite{clark2013tcia}.
We also remove every volume or series from any patient whose scans appear in the public training records of either model.
We exclude HCC-TACE entirely, since it has no independent official validation or test split and six of its candidate cases overlap with the training set of $\floc$.

\myparagraph{Prompt formulation.}
The prompt of the agent follows the template \texttt{\{target\} in \{region\} \{modality\}}.
For $\floc$, we follow its own prompt convention: CVPR-BiomedSegFM ROIs use the training prompt of their dataset, other ROIs of a known class use the first training prompt of this class, and ROIs without a matching class use the template of the agent (\eg, odontogenic lesions in DOLCHID~\cite{huang2026dolchid}).

\myparagraph{Morphological complexity.}
For the shape analysis of multi-view verification (\cref{sec:suppl_abl_harness}), we measure two quantities of each ground-truth ROI $M^\ast$: its physical volume $\operatorname{Vol}(M^\ast)$ and its surface area $\operatorname{Area}(\partial M^\ast)$, counted on voxel faces.
We rank ROIs by their sphericity $\psi=\pi^{1/3}[6\operatorname{Vol}(M^\ast)]^{2/3}/\operatorname{Area}(\partial M^\ast)$, which divides the surface area of a sphere with the same volume by the surface area of the ROI.
The sphericity $\psi$ equals 1 for a sphere and decreases as the ROI becomes more elongated, hollow, or fragmented.
We call an ROI complex when $\psi<0.386$ and compact otherwise, a threshold that depends only on the ground-truth shapes: it marks about the least spherical third of the 630 ROIs outside the four C4 lesion datasets, which follow the same threshold.
Of all 690 ROIs, 203 are complex; of the 190-ROI subset, 58 are complex and 132 compact.
Complex ROIs are mostly tubular, ring-shaped, or branching structures (\eg, aorta, myocardium, and airway), whereas compact ROIs are rounder structures (\eg, liver, spleen, and lymph nodes).
The two groups have similar median equivalent spherical diameters (4.2 and 4.3\,cm), so the split separates shape rather than size.

\myparagraph{Development set.}
We developed prompts and thresholds on a pool of cases from the validation and test splits of TotalSegmentator~v2~\cite{wasserthal2023totalsegmentator}, unused CVPR-BiomedSegFM validation cases, WORD, MS-CMRSeg, RHUH-GBM, and FLARE22.
From this pool, we fixed a development set of 74 ROIs (\cref{tab:suppl_dev}): 29 for quick iteration and 45 more for wider coverage.
The development pool shares no cases with the benchmark, and for CVPR-BiomedSegFM sources it shares no subjects either.
\cref{tab:suppl_exposure} reports separately the C4 datasets that the pool does not include.

\begin{table}[!htb]
\centering
\caption{\textbf{Composition of the 74-ROI development set and of the benchmark.}
ROI types and sizes follow \mainref{sec:setup}.}
\label{tab:suppl_dev}
\footnotesize
\setlength{\tabcolsep}{4pt}
\renewcommand{\arraystretch}{1.15}
\begin{tabular}{@{}llccccccccc@{}}
\toprule
\multirow{2}{*}{\textbf{Source}} & \multirow{2}{*}{\textbf{Modality}} & \multirow{2}{*}{\textbf{ROIs}} & \multicolumn{4}{c}{\textbf{Anatomy}} & \multicolumn{2}{c}{\textbf{Lesion}} & \multicolumn{2}{c}{\textbf{ROI size}} \\
\cmidrule(lr){4-7}\cmidrule(lr){8-9}\cmidrule(l){10-11}
& & & \textbf{Organ} & \textbf{Muscle} & \textbf{Vessel} & \textbf{Bone} & \textbf{Lesion} & \textbf{Lymph node} & \textbf{Small} & \textbf{Large} \\
\midrule
TotalSegmentator~v2 & CT & 34 & 22 & 2 & 4 & 6 & 0 & 0 & 4 & 30 \\
CVPR-BiomedSegFM (27 datasets) & CT 11, MR 15, PET 1 & 27 & 11 & 0 & 1 & 2 & 12 & 1 & 7 & 20 \\
FLARE22 & CT & 4 & 4 & 0 & 0 & 0 & 0 & 0 & 0 & 4 \\
WORD & CT & 3 & 3 & 0 & 0 & 0 & 0 & 0 & 0 & 3 \\
MS-CMRSeg & MR & 4 & 4 & 0 & 0 & 0 & 0 & 0 & 0 & 4 \\
RHUH-GBM & MR & 2 & 0 & 0 & 0 & 0 & 2 & 0 & 1 & 1 \\
\midrule
Development set & CT 52, MR 21, PET 1 & 74 & 44 & 2 & 5 & 8 & 14 & 1 & 12 & 62 \\
Benchmark & CT 368, MR 319, PET 3 & 690 & 398 & 11 & 34 & 10 & 222 & 15 & 260 & 430 \\
\bottomrule
\end{tabular}
\end{table}

\myparagraph{Ablation subset.}
To limit the API cost, we run the ablations on a 190-ROI subset that covers every kind of ROI in the benchmark.
From the 630 ROIs outside the four C4 lesion datasets (MSWAL, DOLCHID, Pituitary-MRI, and UT-EndoMRI), the subset includes at least one ROI for each ROI name in each dataset.
It also includes four ROIs from each of the four lesion datasets.
Of the 190 ROIs, 69 have an equivalent diameter below 3\,cm.
As with \method in \mainref{tab:main}, every run falls back to the ROI search without a prior when $\floc$ returns an empty prediction (\cref{sec:suppl_init}), and an ROI without a final prediction scores 0.

\newpage %
\section{Experimental Setup}
\label{sec:suppl_setup}

This section gives the configuration of the MLLMs, of nnInteractive Agent, and of the other models in our experiments.

\myparagraph{MLLMs.}
Unless noted otherwise, \method runs every MLLM with the same prompts at its default reasoning effort; only \cref{tab:suppl_reasoning} uses other effort levels.
Among these MLLMs, DeepSeek-V4.1-Flash and Qwen3.8-Flash are open-weight models.
All runs of DeepSeek-V4.1-Flash use its 1017 version.

\myparagraph{nnInteractive Agent.}
We run nnInteractive Agent~\cite{isensee2026nninteractiveagent} from its public repository at commit \texttt{57b06cb}, using its unguided track and the public nnInteractive\_v1.0 weights.
GPT-6.1-Sol and GPT-5.6-Sol run in Codex CLI 0.160.0~\cite{openai2025codex} and Claude-Sonnet-5.5 runs in Claude Code 2.1.314~\cite{anthropic2025claudecode}, all at medium reasoning effort.
Each ROI runs alone in a bubblewrap sandbox, with a limit of 20 minutes and no limit on model calls, so that the agent can use as many turns as it needs within that time.
For \mainref{tab:codex}, nnInteractive Agent with GPT-5.6-Sol also runs on the 190-ROI subset with the localization prior: it receives the same cached mask of $\floc$ that \method starts from, is told that the mask may be wrong, and may load, correct, or discard it; all other settings stay the same.
Three of these runs reach the time limit and four end without a mask, and all seven score 0.

\myparagraph{Other models.}
BiomedParse~v2~\cite{zhao2025biomedparse,zhao2025boltzformer}, nnInteractive~\cite{isensee2025nninteractive}, the five end-to-end segmentation models, and MedSAM-Agent~\cite{liu2026medsamagent} all run from their public weights.
The oracle of \mainref{tab:main} starts from an empty prediction and simulates its clicks from the largest connected component of the ground-truth mask.
The ablations use two more models: SAT-Nano~\cite{zhao2025sat} (\cref{tab:suppl_prior}) at commit \texttt{5676be5} with its public Nano weights, and MedSAM2~\cite{ma2025medsam2} (\cref{tab:suppl_backend}) at commit \texttt{332f30d} with the public \texttt{MedSAM2\_latest} weights.

\section{Coding-Agent Evaluation}
\label{sec:suppl_codex}

This section gives the evaluation details of Codex in \mainref{tab:codex}, where Codex runs on the 190-ROI subset of \cref{sec:suppl_bench}.

\myparagraph{Observation conditions.}
Codex annotates each ROI under four conditions, which differ in what Codex sees and in whether its initial view comes from the localization prior.
In the volume condition, Codex reads the raw 3D volume with its own code, whereas in the axial and tri-planar conditions it sees rendered images of one axial slice or of tri-planes.
The fourth condition adds the localization prior to the tri-planar condition: its initial view comes from the prior and has the same center and side length as the initial view of \method.
Without the prior, the initial view shows the whole scan.

\myparagraph{Agent setup and access to $\fseg$.}
We run Codex CLI 0.160.0~\cite{openai2025codex} with GPT-6.1-Sol~\cite{openai2026gpt61sol} at medium reasoning effort.
For each ROI, a new session starts in an isolated sandbox that contains neither our code nor the ground-truth masks.
The task description gives Codex the target name and annotation guidelines that \method receives, and it states the scoring metric.
Codex controls the same $\fseg$ through a command-line tool: it places one positive or negative click at a time, gets the 3D prediction, and accepts or rejects it before the next click.
In the image-based conditions, it can also ask for a new view with another center or side length.
Each ROI allows up to 20 clicks, 60 views, and 20 minutes.

\myparagraph{Scoring and audit of the sessions.}
We score the final accepted prediction by 3D Dice over the whole volume, and we use the ground-truth masks only for this score.
An ROI scores 0 when Codex submits no prediction.
We checked every shell command that Codex ran: none made an external network call or tried to leave the sandbox.

\section{Extended Experiment Results}
\label{sec:suppl_ablation}

This section extends the experiments of the main paper.
Unless noted otherwise, they use the 190-ROI subset of \cref{sec:suppl_bench}, and the tables report Dice (\%) and bold the best result of each MLLM.
Small, Large, and FA follow \mainref{sec:setup}.
\mainref{tab:main} leaves out \method with Qwen3.8-Flash for space, so \cref{tab:suppl_qwen} gives its results on all 690 ROIs in the same columns.

\begin{table}[!htb]
\centering
\footnotesize
\setlength{\tabcolsep}{4pt}
\caption{\textbf{Performance of \method with Qwen3.8-Flash on the benchmark (690 ROIs).}
Columns as in \mainref{tab:main}.}
\label{tab:suppl_qwen}
\begin{tabular}{@{}l cc cccc cc cc@{}}
\toprule
& \multicolumn{2}{c}{\textbf{Overall}}
& \multicolumn{4}{c}{\textbf{Anatomy} (453)}
& \multicolumn{2}{c}{\textbf{Lesion} (237)}
& \multicolumn{2}{c}{\textbf{ROI size}} \\
\cmidrule(lr){2-3}\cmidrule(lr){4-7}\cmidrule(lr){8-9}\cmidrule(l){10-11}
\textbf{MLLM} & \textbf{Mean} & \textbf{Median}
& \textbf{Organ} & \textbf{Muscle} & \textbf{Vessel} & \textbf{Bone}
& \textbf{Lesion} & \textbf{Lymph node}
& \textbf{Small} & \textbf{Large} \\
\midrule
Qwen3.8-Flash~\cite{qiu2026qwen38next} & 64.9 (2.4) & 78.0 & 73.1 & 77.7 & 69.3 & 26.5 & 51.6 & 50.2 & 47.8 & 75.2 \\
\bottomrule
\end{tabular}
\end{table}

\subsection{Localization and Generalization}
\label{sec:suppl_abl_loc}

\begin{table}[!htb]
\centering
\footnotesize
\setlength{\tabcolsep}{4pt}
\renewcommand{\arraystretch}{1.15}
\caption{
\textbf{Ablation on the localization prior.}
Every run uses the ROI search fallback.
Seen and Unseen: the seen and unseen datasets of $\floc$ (\mainref{fig:prior}).
ROIs that ended in a runtime error were run again once; 7 ROIs with SAT-Nano and DeepSeek-V4.1-Flash still failed and are scored as they stand.
}
\label{tab:suppl_prior}
\begin{tabularx}{\linewidth}{@{}ll CCC CC Z{C}@{}}
\toprule
\multirow{3}{*}{\textbf{MLLM}} & \multirow{3}{*}{\textbf{Prior}}
& \multicolumn{5}{c}{\textbf{Dice}}
& \\
\cmidrule(lr){3-7}
& & \textbf{All} & \textbf{Seen} & \textbf{Unseen} & \textbf{Small} & \textbf{Large} & \\
& & (190) & (159) & (31) & (69) & (121) & \multirow{-3}{*}{\textbf{FA}$\downarrow$} \\
\midrule
& None & 25.4 & 23.7 & 34.2 & 12.8 & 32.6 & 65 \\
& SAT-Nano & 32.8 & 31.8 & 37.9 & 14.1 & 43.5 & 71 \\
\rowcolor{blue!5} \cellcolor{white}\multirow{-3}{*}{DeepSeek-V4.1-Flash} & BiomedParse & \textbf{63.8} & \textbf{68.2} & \textbf{41.0} & \textbf{43.0} & \textbf{75.7} & \textbf{23} \\
\midrule
& None & 44.2 & 42.8 & 50.8 & 22.6 & 56.5 & 29 \\
& SAT-Nano & 41.4 & 38.2 & \textbf{57.9} & 22.2 & 52.3 & 38 \\
\rowcolor{blue!5} \cellcolor{white}\multirow{-3}{*}{Gemini-3.8-Flash} & BiomedParse & \textbf{65.8} & \textbf{67.5} & 57.2 & \textbf{49.3} & \textbf{75.3} & \textbf{12} \\
\bottomrule
\end{tabularx}
\end{table}

\begin{table}[!htb]
\centering
\begin{minipage}[t]{0.53\linewidth}
\caption{
\textbf{Localization accuracy of the prior and of the ROI search.}
Each entry is the share of ROIs (\%) whose point falls inside the ground truth.
This point is either the first view center from $\floc$ or the last point of the MLLM's whole-scan ROI search.
A search without a point counts as a miss.
Seen and Unseen are as in \cref{tab:suppl_prior}.
The best share in each column is in bold.
}
\label{tab:suppl_search}
\end{minipage}\hfill
\begin{minipage}[t]{0.44\linewidth}
\caption{
\textbf{Ablation of click selection by ROI size.}
\xmark\ marks the removed design, and the shaded row of each MLLM is the full \method.
The number of ROIs is below each size.
}
\label{tab:suppl_click_size}
\end{minipage}
\\[4pt]
\footnotesize
\setlength{\tabcolsep}{3pt}
\renewcommand{\arraystretch}{1.15}
\begin{minipage}[t]{0.53\linewidth}
\centering
\begin{tabularx}{\linewidth}{@{}Y{l} CCC C Z{C}@{}}
\toprule
& \multicolumn{5}{c}{\textbf{Inside the ground truth}} \\
\cmidrule(l){2-6}
& \textbf{All} & \textbf{Seen} & \textbf{Unseen} & \textbf{Small} & \textbf{Large} \\
\multirow{-3}{*}{\textbf{Point}} & (190) & (159) & (31) & (69) & (121) \\
\midrule
DeepSeek-V4.1-Flash ROI search & 33 & 32 & 39 & 17 & 42 \\
Gemini-3.8-Flash ROI search & 59 & 57 & \textbf{74} & 33 & 74 \\
SAT-Nano view center & 34 & 34 & 32 & 9 & 48 \\
\rowcolor{blue!5} BiomedParse view center & \textbf{83} & \textbf{91} & 45 & \textbf{65} & \textbf{93} \\
\bottomrule
\end{tabularx}
\end{minipage}\hfill
\begin{minipage}[t]{0.44\linewidth}
\centering
\begin{tabularx}{\linewidth}{@{}lc CCCC Z{C}@{}}
\toprule
\multirow{3}{*}{\textbf{MLLM}} & \multirow{3}{*}{\makecell{\textbf{Click}\\\textbf{selection}}}
& \multicolumn{4}{c}{\textbf{Equivalent diameter (cm)}} & \\
\cmidrule(lr){3-6}
& & \textbf{$<$1.5} & \textbf{1.5--3} & \textbf{3--6} & \textbf{$\geq$6} & \textbf{All} \\
& & (28) & (41) & (60) & (61) & (190) \\
\midrule
& \xmark & 25.2 & \textbf{53.3} & 64.0 & 83.8 & 62.3 \\
\rowcolor{blue!5} \cellcolor{white}\multirow{-2}{*}{\shortstack[l]{DeepSeek-\\V4.1-Flash}} & \cmark & \textbf{31.7} & 50.7 & \textbf{64.5} & \textbf{86.7} & \textbf{63.8} \\
\midrule
& \xmark & 33.4 & 52.2 & 58.9 & 83.0 & 61.4 \\
\rowcolor{blue!5} \cellcolor{white}\multirow{-2}{*}{\shortstack[l]{Gemini-\\3.8-Flash}} & \cmark & \textbf{36.7} & \textbf{57.9} & \textbf{66.0} & \textbf{84.4} & \textbf{65.8} \\
\bottomrule
\end{tabularx}
\end{minipage}
\end{table}

\myparagraph{Localization prior and ROI search.}
The localization prior gives \method its starting point, and the ROI search recovers some of the targets that the localization misses (\cref{tab:suppl_prior,tab:suppl_search}).
The prior helps most on small targets, which appear on only a few of the hundreds of slices, and on the seen datasets of $\floc$: on both, $\floc$ places the initial view center inside the target far more often than the ROI search does.
On the unseen datasets of $\floc$, the ROI search of Gemini-3.8-Flash places more points inside the target than $\floc$, so the fallback recovers targets that $\floc$ misses there.
A prior that localizes the target too rarely loses most of its benefit: SAT-Nano places the initial view center inside the target about as often as the ROI search of DeepSeek-V4.1-Flash, so with either MLLM, the performance with SAT-Nano stays much closer to that without a prior than to that with $\floc$.
\method therefore gains directly from a more accurate localization model.

\begin{table}[!htb]
\centering
\footnotesize
\setlength{\tabcolsep}{4pt}
\renewcommand{\arraystretch}{1.15}
\caption{\textbf{Performance on the seen and unseen datasets of \boldmath$\floc$ (690 ROIs).}
Seen and Unseen: ROIs from the seen (480) and unseen (210) datasets of $\floc$; C4: datasets unseen by both $\floc$ and $\fseg$ (165); Not used in development: the C4 datasets that we never used while developing \method (90).
$\floc$ and $\floc{\rightarrow}\fseg$ (\mainref{fig:prior}) are shown as references; the best and second-best results of \method are bolded and underlined.}
\label{tab:suppl_exposure}
\newcolumntype{F}{>{\columncolor{white}[0pt][\tabcolsep]}p{0.14\linewidth}}
\newcolumntype{G}{>{\columncolor{white}[\tabcolsep][0pt]\centering\arraybackslash}p{0.2\linewidth}} %
\begin{tabularx}{\linewidth}{@{}F p{0.17\linewidth} CCCCG@{}}
\toprule
\textbf{Method} & \textbf{MLLM} & \textbf{All} & \textbf{Seen} & \textbf{Unseen} & \textbf{C4} & \textbf{Not used in development} \\
\midrule
\rowcolor{black!6} $\floc$~\cite{zhao2025biomedparse} & -- & 66.9 & 75.2 & 47.8 & 35.7 & 13.0 \\
\rowcolor{black!6} $\floc{\rightarrow}\fseg$ & -- & 65.6 & 72.2 & 50.7 & 39.7 & 16.1 \\
\midrule
\rowcolor{blue!5}  & Claude-Sonnet-5.5 & 63.2 & 67.6 & 53.2 & 43.0 & 25.2 \\
\rowcolor{blue!5} & GPT-5.6-Sol & 64.2 & 65.5 & 61.3 & 54.0 & 39.2 \\
\rowcolor{blue!5} & DeepSeek-V4.1-Flash & 64.7 & 69.0 & 55.0 & 45.3 & 24.1 \\
\rowcolor{blue!5} & Qwen3.8-Flash & 64.9 & 68.1 & 57.5 & 48.6 & 28.6 \\
\rowcolor{blue!5} & Gemini-3.8-Flash & 66.3 & 66.4 & 66.1 & 59.2 & 46.6 \\
\rowcolor{blue!5} & GPT-6.1-Sol & 68.8 & 68.2 & \underline{70.2} & \underline{64.4} & \underline{53.3} \\
\rowcolor{blue!5} & Claude-Opus-5.5 & \underline{69.2} & \textbf{70.7} & 65.6 & 58.7 & 45.6 \\
\rowcolor{blue!5} \multirow{-8}{*}{\method} & GPT-6-Astra & \textbf{70.0} & \underline{69.3} & \textbf{71.5} & \textbf{66.2} & \textbf{57.2} \\
\bottomrule
\end{tabularx}
\end{table}

\begin{table}[!htb]
\centering
\footnotesize
\setlength{\tabcolsep}{4pt}
\renewcommand{\arraystretch}{1.15}
\caption{\textbf{Performance on the five datasets whose targets have several separate lesions (40 ROIs).}
Lesions: median number of separate lesions in a ground-truth mask, with the range in brackets.
COVID-19 combines COVID-19-20 and COVID-19-CT-Seg. The first three MLLMs of \method are those that nnInteractive Agent also runs, and the best result in the All five row is bolded.}
\label{tab:suppl_multifocal}
\begin{tabularx}{\linewidth}{@{}lc r@{\ }l CCCCC CCC@{}}
\toprule
& & \multicolumn{2}{c}{} & \multicolumn{5}{c}{\textbf{\method}} & \multicolumn{3}{c}{\textbf{nnInteractive Agent}} \\
\cmidrule(lr){5-9}\cmidrule(l){10-12}
\textbf{Dataset} & \textbf{ROIs} & \multicolumn{2}{c}{\textbf{Lesions}}
& \makecell[b]{\textbf{GPT-5.6-}\\\textbf{Sol}} & \makecell[b]{\textbf{Claude-}\\\textbf{Sonnet-5.5}} & \makecell[b]{\textbf{GPT-6.1-}\\\textbf{Sol}} & \makecell[b]{\textbf{Gemini-}\\\textbf{3.8-Flash}} & \makecell[b]{\textbf{GPT-6-}\\\textbf{Astra}}
& \makecell[b]{\textbf{GPT-5.6-}\\\textbf{Sol}} & \makecell[b]{\textbf{Claude-}\\\textbf{Sonnet-5.5}} & \makecell[b]{\textbf{GPT-6.1-}\\\textbf{Sol}} \\
\midrule
WMH & 3 & 26 & [14, 29] & 58.7 & 56.7 & 61.6 & 34.1 & 61.1 & 54.8 & 55.7 & 51.9 \\
COVID-19 & 4 & 12.5 & [1, 34] & 42.6 & 40.0 & 41.8 & 45.5 & 42.6 & 32.7 & 31.9 & 49.7 \\
autoPET & 3 & 21 & [3, 76] & 35.9 & 33.1 & 34.7 & 36.0 & 36.0 & 4.4 & 26.1 & 4.4 \\
LNQ 2023 & 15 & 2 & [1, 5] & 47.9 & 48.4 & 51.6 & 48.1 & 50.9 & 17.8 & 24.0 & 36.3 \\
CRLM & 15 & 2 & [1, 8] & 55.8 & 34.6 & 51.7 & 38.0 & 55.9 & 38.1 & 34.0 & 50.3 \\
\midrule
All five & 40 & 2 & [1, 76] & 50.2 & 41.8 & 50.1 & 42.1 & \textbf{51.6} & 28.7 & 31.1 & 41.7 \\
\bottomrule
\end{tabularx}
\end{table}

\myparagraph{Generalization to unseen datasets.}
With every MLLM, \method outperforms $\floc$ and $\floc{\rightarrow}\fseg$ on the unseen datasets of $\floc$, on the datasets unseen by both $\floc$ and $\fseg$ (C4), and on the datasets never used while developing \method (\cref{tab:suppl_exposure}).
This result extends \mainref{fig:prior} from two MLLMs to all eight, and with frontier MLLMs, \method also outperforms $\floc$ over all ROIs.
Since $\fseg$ raises the Dice of $\floc$ only slightly on these datasets, the gain comes from the agent, which checks the initial prediction in tri-planes and corrects it through clicks.
The lead is largest on the datasets never used in development, so it does not stem from tuning the agent to its development data.

\myparagraph{Targets with several lesions.}
On the five datasets whose targets have several separate lesions, \method reaches a higher mean Dice with every MLLM than nnInteractive Agent with any MLLM (\cref{tab:suppl_multifocal}).
This lead holds although \method grows its initial prediction from the main component of the localization prior (\cref{sec:suppl_init}).

\subsection{Input Layout}
\label{sec:suppl_abl_layout}

\begin{table}[!htb]
\centering
\footnotesize
\setlength{\tabcolsep}{4pt}
\renewcommand{\arraystretch}{1.15}
\caption{
\textbf{Ablation on the input layout with DeepSeek-V4.1-Flash and Gemini-3.8-Flash.}
Slice stack: parallel axial slices around the view center, one image per slice except for ``3 in one image''; Tokens / ROI: total over all MLLM calls.
}
\label{tab:suppl_layout}
\newcolumntype{D}{>{\centering\arraybackslash\leavevmode}X}
\begin{tabularx}{\linewidth}{@{}ll c D DD D Z{D}@{}}
\toprule
\multirow{2}{*}{\textbf{MLLM}} & \multirow{2}{*}{\textbf{Layout}} & \multirow{2}{*}{\textbf{Slices / view}} & \multirow{2}{*}{\textbf{Calls / ROI}$\downarrow$}
& \multicolumn{2}{c}{\textbf{Tokens / ROI}}
& \multicolumn{2}{c}{\textbf{Dice}} \\
\cmidrule(lr){5-6}\cmidrule(l){7-8}
& & & & \textbf{Input}$\downarrow$ & \textbf{Reasoning}$\downarrow$ & \textbf{Mean} & \textbf{Median} \\
\midrule
& \multirow{2}{*}{Slice stack} & 16 & -- & 49.4k & 61.0k & 63.5 & 75.9 \\
& & 3 in one image & -- & 13.5k & 43.2k & \textbf{64.0} & 77.0 \\
\rowcolor{blue!5} \cellcolor{white}\multirow{-3}{*}{DeepSeek-V4.1-Flash} & Tri-planes & 3 & -- & \textbf{12.1k} & \textbf{32.8k} & 63.8 & \textbf{79.1} \\
\midrule
& \multirow{5}{*}{Slice stack} & 16 & 14.4 & 337.1k & 28.4k & 63.0 & 78.3 \\
& & 9 & 13.6 & 191.7k & 25.1k & 63.9 & 78.4 \\
& & 5 & 14.2 & 119.1k & 24.1k & 63.3 & 78.8 \\
& & 3 & 13.9 & 75.0k & 21.2k & 62.5 & 77.9 \\
& & 3 in one image & 13.0 & 32.9k & 18.9k & 63.9 & 79.0 \\
\rowcolor{blue!5} \cellcolor{white}\multirow{-6}{*}{Gemini-3.8-Flash} & Tri-planes & 3 & \textbf{11.7} & \textbf{28.1k} & \textbf{15.5k} & \textbf{65.8} & \textbf{79.3} \\
\bottomrule
\end{tabularx}
\end{table}

With both MLLMs, tri-planes are the most context-efficient input layout without loss in performance (\cref{tab:suppl_layout}).
With Gemini-3.8-Flash, a slice stack uses more input tokens per ROI as the number of slices grows, whereas its performance stays below that of tri-planes at every number of slices.
A stack of 16 slices uses $4.1\times$ as many input tokens per ROI as tri-planes with DeepSeek-V4.1-Flash and $12\times$ as many with Gemini-3.8-Flash.
It also uses nearly twice as many reasoning tokens with both MLLMs.
Placing three panels side by side in one image is an implementation choice that tri-planes share with the ``3 in one image'' stack (\mainref{sec:setup}).
With the same placement and the same number of panels, tri-planes need fewer input and reasoning tokens than three parallel axial slices with both MLLMs.
With Gemini-3.8-Flash, they also need fewer calls and perform better, since they show the view along three axes rather than one.

\subsection{Harness Design Ablation}
\label{sec:suppl_abl_harness}

\begin{table}[!htb]
\centering
\footnotesize
\setlength{\tabcolsep}{4pt}
\renewcommand{\arraystretch}{1.15}
\caption{
\textbf{Ablation of the harness with DeepSeek-V4.1-Flash and Gemini-3.8-Flash.}
Marks and columns as in \mainref{tab:ablation}.
}
\label{tab:suppl_ablation_deepseek}
\begin{tabularx}{\linewidth}{@{}lccc CC CC Z{C}@{}}
\toprule
\multirow{2}{*}{\textbf{MLLM}} & \multirow{2}{*}{\makecell{\textbf{Prior}\\\textbf{mask}}} & \multirow{2}{*}{\makecell{\textbf{Click}\\\textbf{selection}}} & \multirow{2}{*}{\makecell{\textbf{Multi-view}\\\textbf{verification}}}
& \multicolumn{4}{c}{\textbf{Dice}}
& \\
\cmidrule(lr){5-8}
& & &
& \textbf{Mean} & \textbf{Median} & \textbf{Small} & \textbf{Large} & \multirow{-2}{*}{\textbf{FA}$\downarrow$} \\
\midrule
& \xmark & \cmark & \cmark & 52.6 & 62.6 & 33.9 & 63.2 & 30 \\
& \cmark & \xmark & \cmark & 62.3 & 76.6 & 41.9 & 74.0 & \textbf{23} \\
& \cmark & \cmark & \xmark & 63.7 & 79.0 & 42.6 & \textbf{75.7} & 28 \\
\rowcolor{blue!5} \cellcolor{white}\multirow{-4}{*}{DeepSeek-V4.1-Flash} & \cmark & \cmark & \cmark & \textbf{63.8} & \textbf{79.1} & \textbf{43.0} & \textbf{75.7} & \textbf{23} \\
\midrule
& \xmark & \cmark & \cmark & 53.0 & 63.8 & 37.1 & 62.1 & 25 \\
& \cmark & \xmark & \cmark & 61.4 & 77.9 & 44.6 & 71.0 & 15 \\
& \cmark & \cmark & \xmark & 65.1 & 79.1 & 48.5 & 74.6 & 16 \\
\rowcolor{blue!5} \cellcolor{white}\multirow{-4}{*}{Gemini-3.8-Flash} & \cmark & \cmark & \cmark & \textbf{65.8} & \textbf{79.3} & \textbf{49.3} & \textbf{75.3} & \textbf{12} \\
\bottomrule
\end{tabularx}
\end{table}

\myparagraph{Harness designs with DeepSeek-V4.1-Flash.}
Each design of the harness helps DeepSeek-V4.1-Flash as it helps Gemini-3.8-Flash in \mainref{tab:ablation} (\cref{tab:suppl_ablation_deepseek}).
Removing the prior mask lowers the performance the most, since $\fseg$ then starts from a single click instead of the prior.
Multi-view verification lowers the false accepts while keeping the performance.
Click selection improves the performance of both MLLMs on the smallest ROIs, below 1.5\,cm (\cref{tab:suppl_click_size}), where a click that the MLLM places itself can fall outside the target when it is a few pixels off.

\begin{wraptable}[11]{r}{0.55\linewidth} %
\centering
\footnotesize
\setlength{\tabcolsep}{3pt}
\renewcommand{\arraystretch}{1.15}
\vspace{-14pt} %
\caption{\textbf{Ablation of multi-view verification by ROI shape.}
Complex: ROIs with $\psi<0.386$ (58); Compact: the others (132).
\xmark\ marks the removed design, and the shaded row of each MLLM is the full \method.}
\label{tab:suppl_shape}
\begin{tabularx}{\linewidth}{@{}lc CC C Z{C}@{}}
\toprule
\multirow{2}{*}{\textbf{MLLM}} & \multirow{2}{*}{\makecell{\textbf{Multi-view}\\\textbf{verification}}}
& \multicolumn{2}{c}{\makecell{\textbf{Accepted with}\\\textbf{Dice $<$50\%}$\downarrow$}} & \multicolumn{2}{c}{\textbf{Mean Dice}} \\
\cmidrule(lr){3-4}\cmidrule(l){5-6}
& & \textbf{Complex} & \textbf{Compact} & \textbf{Complex} & \textbf{Compact} \\
\midrule
& \xmark & 17 & 29 & 60.1 & \textbf{65.3} \\
\rowcolor{blue!5} \cellcolor{white}\multirow{-2}{*}{\shortstack[l]{DeepSeek-\\V4.1-Flash}} & \cmark & \textbf{12} & \textbf{23} & \textbf{60.6} & 65.2 \\
\midrule
& \xmark & 12 & 22 & 59.6 & 67.5 \\
\rowcolor{blue!5} \cellcolor{white}\multirow{-2}{*}{\shortstack[l]{Gemini-\\3.8-Flash}} & \cmark & \textbf{10} & \textbf{18} & \textbf{59.8} & \textbf{68.5} \\
\bottomrule
\end{tabularx}
\end{wraptable}
\myparagraph{Multi-view verification by ROI shape.}
With both MLLMs, multi-view verification lowers the number of accepted ROIs with a Dice below 50\% on both complex and compact ROIs (\cref{tab:suppl_shape}), so it catches wrong predictions on targets of either shape.
We group ROIs by their sphericity $\psi$ as in \cref{sec:suppl_bench}.
Complex ROIs ($\psi<0.386$) are mostly tubular, ring-shaped, or branching structures, such as the aorta, myocardium, and airway, whereas compact ROIs are rounder structures, such as the liver, spleen, and lymph nodes.

\begin{table}[!htb]
\vspace{8pt} %
\centering
\footnotesize
\setlength{\tabcolsep}{5pt}
\renewcommand{\arraystretch}{1.15}
\caption{\textbf{Coverage of the verification views.}
The shares are measured on the first view of each round, with $\tau=2$\,mm. Per round: among the rounds whose prediction has an error region, the share in which the view shows at least one. Per error region: share of the error regions in these rounds shown by the view. Brackets: 95\% bootstrap intervals over ROIs; parentheses: number of rounds or error regions.}
\label{tab:suppl_coverage}
\begin{tabular}{c cc cc}
\toprule
& \multicolumn{2}{c}{\textbf{Per round}} & \multicolumn{2}{c}{\textbf{Per error region}} \\
\cmidrule(lr){2-3}\cmidrule(l){4-5}
$\gamma$ & \textbf{Gemini-3.8-Flash} & \textbf{DeepSeek-V4.1-Flash} & \textbf{Gemini-3.8-Flash} & \textbf{DeepSeek-V4.1-Flash} \\
\midrule
5\% & 95.5 [90.5, 99.2] (111) & 95.8 [91.9, 99.1] (120) & 85.8 [78.8, 92.0] (169) & 82.2 [75.3, 88.7] (197) \\
10\% & 95.6 [89.9, 100.0] (90) & 93.8 [87.7, 98.9] (96) & 89.8 [83.2, 95.9] (108) & 83.7 [77.4, 89.9] (129) \\
25\% & 94.8 [87.9, 100.0] (58) & 92.5 [85.7, 98.4] (67) & 90.6 [83.3, 97.1] (64) & 85.7 [78.2, 93.0] (77) \\
\bottomrule
\end{tabular}
\end{table}

\myparagraph{Coverage of the verification views.}
The first verification view of each round already shows at least one remaining error region in the large majority of rounds, and most of the error regions on its own (\cref{tab:suppl_coverage}).
We measure this coverage on the runs of Gemini-3.8-Flash and DeepSeek-V4.1-Flash on the 190-ROI subset.
From their logged views, predictions, and ground truth, we compute whether a view shows an error region (\cref{sec:suppl_rem_verify}) without new MLLM calls.
We count views with the same center and side length as one view, since they show the same images.
We leave out a band of $\tau=2$\,mm along the boundary of $M^\ast$, so that thin boundary disagreements do not count.
We report three fractions $\gamma$; an error region that covers $10\%$ of the target lowers Dice by about five points on its own.
For the first view of each round, this measurement gives $\pcov$ in \cref{prop:suppl_coverage}.

\subsection{Model Choices and Hyperparameters}
\label{sec:suppl_abl_settings}

\myparagraph{Reasoning effort.}
\method keeps its performance at other reasoning efforts with either MLLM (\cref{tab:suppl_reasoning}).
With Gemini-3.8-Flash, high effort produces far more reasoning tokens per reply but does not improve the performance.
With DeepSeek-V4.1-Flash, low effort uses fewer reasoning tokens per reply and keeps the performance.

\begin{table}[!htb]
\centering
\footnotesize
\setlength{\tabcolsep}{4pt}
\renewcommand{\arraystretch}{1.15}
\caption{
\textbf{Ablation on the reasoning effort.}
Reas.: median number of reasoning tokens per model reply.
}
\label{tab:suppl_reasoning}
\begin{tabularx}{\linewidth}{@{}llc CC C Z{C}@{}}
\toprule
\multirow{2}{*}{\textbf{MLLM}} & \multirow{2}{*}{\textbf{Effort}} & \multirow{2}{*}{\textbf{Reas.}}
& \multicolumn{4}{c}{\textbf{Dice}} \\
\cmidrule(l){4-7}
& & & \textbf{Mean} & \textbf{Median} & \textbf{Small} & \textbf{Large} \\
\midrule
& Low & 2489 & \textbf{64.2} & 77.5 & 41.3 & \textbf{77.2} \\
\rowcolor{blue!5} \cellcolor{white}\multirow{-2}{*}{DeepSeek-V4.1-Flash} & Medium & 2751 & 63.8 & \textbf{79.1} & \textbf{43.0} & 75.7 \\
\midrule
& High & 4577 & 65.6 & 79.0 & 48.5 & \textbf{75.3} \\
\rowcolor{blue!5} \cellcolor{white}\multirow{-2}{*}{Gemini-3.8-Flash} & Medium & 1030 & \textbf{65.8} & \textbf{79.3} & \textbf{49.3} & \textbf{75.3} \\
\bottomrule
\end{tabularx}
\end{table}

\begin{table}[!htb]
\centering
\footnotesize
\setlength{\tabcolsep}{4pt}
\renewcommand{\arraystretch}{1.15}
\caption{
\textbf{Ablation on the interactive segmentation model \boldmath$\fseg$.}
\method runs with DeepSeek-V4.1-Flash, the same prompts, and the ROI search fallback.
Columns as in \mainref{tab:ablation}.
}
\label{tab:suppl_backend}
\begin{tabularx}{\linewidth}{@{}Y{l} CC CC Z{C}@{}}
\toprule
& \multicolumn{4}{c}{\textbf{Dice}}
& \\
\cmidrule(lr){2-5}
\multirow{-2}{*}{\textbf{$\fseg$}} & \textbf{Mean} & \textbf{Median} & \textbf{Small} & \textbf{Large} & \multirow{-2}{*}{\textbf{FA}$\downarrow$} \\
\midrule
MedSAM2~\cite{ma2025medsam2} & 61.6 & 74.5 & 39.1 & 74.5 & \textbf{18} \\
\rowcolor{blue!5} nnInteractive~\cite{isensee2025nninteractive} & \textbf{63.8} & \textbf{79.1} & \textbf{43.0} & \textbf{75.7} & 23 \\
\bottomrule
\end{tabularx}
\end{table}

\myparagraph{Interactive segmentation model.}
The harness exchanges only clicks and 3D predictions with $\fseg$, so it works with other interactive segmentation models.
MedSAM2~\cite{ma2025medsam2} segments a 3D volume by propagating 2D predictions across slices.
With MedSAM2 in place of nnInteractive, \method performs similarly without any change to the harness (\cref{tab:suppl_backend}).

\begin{figure}[!htb]
\centering
\begin{minipage}[t]{0.46\textwidth}
\centering
\small \textbf{(a)} Call budget
\end{minipage}\hfill
\begin{minipage}[t]{0.52\textwidth}
\centering
\small \textbf{(b)} Initial view size
\end{minipage}
\\[3pt]
\begin{minipage}[c]{0.46\textwidth}
\centering
\includegraphics{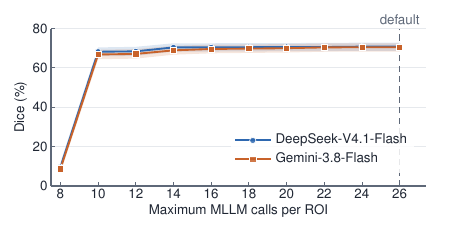}
\end{minipage}\hfill
\begin{minipage}[c]{0.52\textwidth}
\centering
\scriptsize
\setlength{\tabcolsep}{4pt}
\renewcommand{\arraystretch}{1.5}
\raisebox{5.8pt}[\height][\depth]{%
\begin{tabular}{@{}Y{l} c ccc c Z{c}@{}}
\toprule
& \multirow{2}{*}{\makecell{\textbf{Reruns}\\(\#)}}
& \multicolumn{3}{c}{\textbf{Dice}}
& \multirow{2}{*}{\makecell{\textbf{Mean Dice}\\\textbf{vs.\ default}}}
& \\
\cmidrule(lr){3-5}
\multirow{-2}{*}{\textbf{Prior fill}} & & \textbf{Mean} & \textbf{Small} & \textbf{Large} & & \multirow{-2}{*}{\textbf{FA}$\downarrow$} \\
\midrule
50\% & 25 & 63.6 & 41.7 & \textbf{76.1} & $-0.17$ [$-1.25$, $+0.64$] & \textbf{23} \\
\rowcolor{blue!5} 65\% (default) & -- & \textbf{63.8} & 43.0 & 75.7 & -- & \textbf{23} \\
80\% & 27 & 63.3 & \textbf{43.7} & 74.5 & $-0.50$ [$-1.80$, $+0.51$] & 25 \\
\bottomrule
\end{tabular}%
}
\end{minipage}
\caption{
\textbf{Ablation on the call budget and the initial view size.}
(a) Dice when the agent stops at each call budget, on the 598 ROIs where neither MLLM runs the ROI search; an ROI scores 0 when the budget runs out before the agent commits a prediction. Bands show 95\% confidence intervals.
(b) Initial view size with DeepSeek-V4.1-Flash.
Prior fill: the largest share of the initial view that the localization prior may fill along each axis. A higher fill therefore allows a smaller view (\cref{sec:suppl_init}).
Reruns: ROIs whose initial view changes at that fill; only these ROIs are run again. Mean Dice vs.\ default: the change in mean Dice from the default fill, with its 95\% confidence interval. Other columns as in \mainref{tab:ablation}.
}
\label{fig:suppl_hparam}
\end{figure}

\myparagraph{Hyperparameters.}
The performance of \method changes little around the default call budget (\cref{fig:suppl_hparam}a) and the default prior fill (\cref{fig:suppl_hparam}b).
Since most ROIs finish within about ten calls, the curve levels off after about 14 calls, well below the default limit of 26.
A rerun with a budget of 40 calls performs similarly, so the default budget does not limit \method.
Because $F_0$ takes one of three preset sizes, a different prior fill changes the initial view of about one in seven ROIs.
At the other two fill values, the performance stays close to its value at the default fill, and the confidence interval of each change covers zero.
The findings of the main paper therefore do not depend on finely tuned hyperparameters.

\section{Statistical Analysis}
\label{sec:suppl_stats}

This section gives the uncertainty of the results in \mainref{tab:main}.
We resample the 690 ROIs with replacement 4,000 times to obtain 95\% bootstrap intervals.
ROIs from the same scan share an image, and ROIs from the same dataset share an annotation protocol and similar difficulty.
We therefore also resample whole datasets 4,000 times, which gives wider and more conservative intervals.
A paired difference between two systems uses the same resamples for both.
We test each pair of systems with the Wilcoxon signed-rank test and correct the $p$-values over all pairs with Holm's method.

\begin{table}[!htb]
\centering
\footnotesize
\setlength{\tabcolsep}{4pt}
\caption{\textbf{Lead of \method over the two strongest baselines in \mainref{tab:main} (690 ROIs).}
$\Delta$: difference in mean Dice (points) with 95\% bootstrap intervals over ROIs and over whole datasets; $p$: Wilcoxon signed-rank test, Holm-corrected over all pairs of systems.
Leaving out any single dataset does not reverse any of these differences.}
\label{tab:suppl_pairs}
\begin{tabular}{@{}l cccc cccc@{}}
\toprule
& \multicolumn{4}{c}{\textbf{Against nnInteractive Agent (GPT-6.1-Sol)}} & \multicolumn{4}{c}{\textbf{Against VoxTell}} \\
\cmidrule(lr){2-5}\cmidrule(l){6-9}
\textbf{MLLM} & $\boldsymbol{\Delta}$ & \textbf{CI (ROI)} & \textbf{CI (dataset)} & $\boldsymbol{p}$ & $\boldsymbol{\Delta}$ & \textbf{CI (ROI)} & \textbf{CI (dataset)} & $\boldsymbol{p}$ \\
\midrule
GPT-6.1-Sol & +9.1 & [+6.8, +11.6] & [+3.6, +14.5] & $1.5{\times}10^{-19}$ & +10.4 & [+7.9, +12.9] & [+4.3, +18.7] & $2.1{\times}10^{-13}$ \\
Claude-Opus-5.5 & +9.5 & [+7.1, +11.9] & [+2.4, +15.4] & $3.6{\times}10^{-23}$ & +10.7 & [+8.3, +13.2] & [+4.4, +18.9] & $1.4{\times}10^{-15}$ \\
GPT-6-Astra & +10.3 & [+8.2, +12.5] & [+4.8, +15.5] & $3.9{\times}10^{-25}$ & +11.6 & [+9.4, +13.9] & [+5.4, +20.1] & $3.5{\times}10^{-17}$ \\
\bottomrule
\end{tabular}
\end{table}

All leads of \method over nnInteractive Agent with GPT-6.1-Sol and over VoxTell are significant, and leaving out any single dataset does not reverse any of them (\cref{tab:suppl_pairs}).

A repeated run of \method gives nearly the same performance.
We ran \method with Gemini-3.8-Flash a second time on all 690 ROIs through another API endpoint.
Its mean Dice changes by 0.3 points (95\% interval $[-0.8, +1.5]$), and 81\% of the ROIs change by less than one point.
The standard error that model randomness adds to the mean (0.42 points) is well below the standard error from sampling ROIs (1.16 points).

\clearpage %
\section{Qualitative Results}
\label{sec:suppl_qualitative}

\cref{fig:suppl_seg} compares the 3D predictions of \method with those of the baselines across ROI types and modalities, and \cref{fig:suppl_cases} shows how one click refines a 3D prediction on four more ROIs.

\begin{figure}[!htb]
\centering
\includegraphics[width=\linewidth]{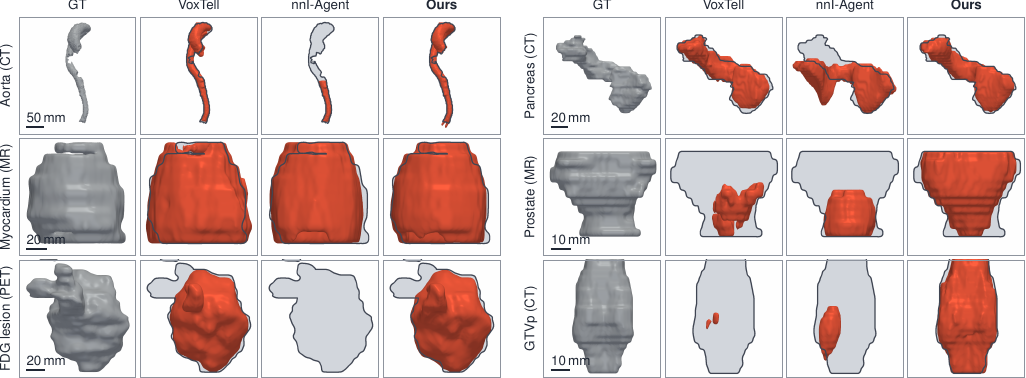}
\caption{\textbf{Comparison of 3D segmentation across ROI types and modalities.}
Gray is the ground truth and orange the 3D prediction; a panel without orange has no prediction in view.
nnI-Agent: nnInteractive Agent~\cite{isensee2026nninteractiveagent} with GPT-6.1-Sol~\cite{openai2026gpt61sol}; \method runs with Gemini-3.8-Flash~\cite{google2026gemini38flash}.}
\label{fig:suppl_seg}
\end{figure}

\begin{figure}[!htb]
\centering
\includegraphics[width=\linewidth]{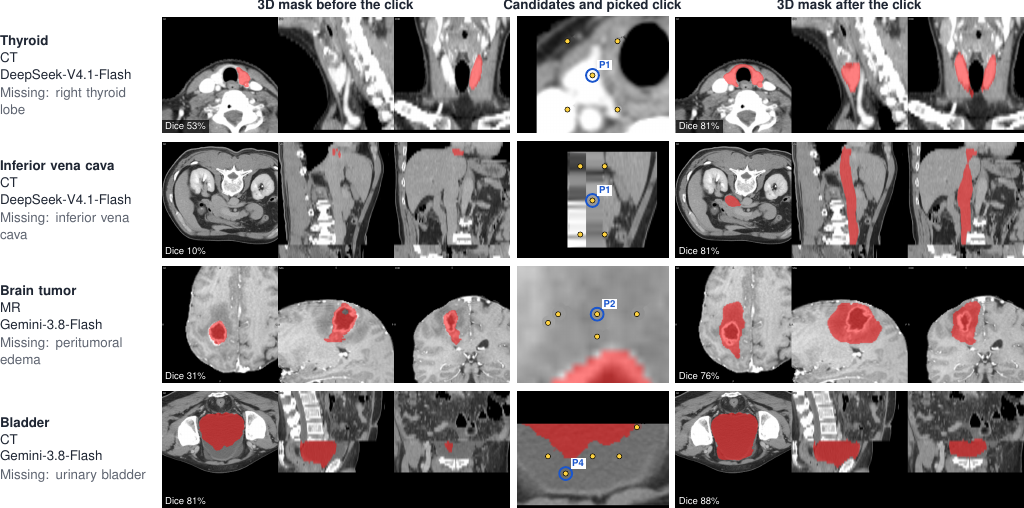}
\caption{\textbf{One click of \method on four ROIs.}
Each row shows the same view before and after the click, with the candidate clicks in between and a blue circle on the one that the MLLM picks.
The left column names the part that the MLLM confirms as missing, and the corner of each view gives the Dice of the 3D prediction.}
\label{fig:suppl_cases}
\end{figure}

  \definecolor{slotcolor}{RGB}{176,72,0}
\newcommand{\slot}[1]{\mbox{\textcolor{slotcolor}{$\langle${\itshape #1}$\rangle$}}}
\newcounter{prompt}\setcounter{prompt}{-1}
\renewcommand{\theprompt}{P\arabic{prompt}}
\newcommand{\promptlabel}[1]{\refstepcounter{prompt}\label{#1}}
\newtcolorbox{promptbox}[2]{
  enhanced, breakable,
  colback=white, colframe=black!28, boxrule=0.5pt, arc=2pt,
  colbacktitle=black!6, coltitle=black!88, titlerule=0.4pt, titlerule style={black!28},
  fonttitle=\sffamily\small, title={\textbf{#1}\\[1pt]{\footnotesize\color{black!62}#2}},
  toptitle=3pt, bottomtitle=3pt, left=8pt, right=8pt, top=6pt, bottom=6pt,
  fontupper=\sffamily\footnotesize, fontlower=\footnotesize,
  before upper={\raggedright\parindent=0pt\parskip=0pt},
  before lower={\parindent=0pt}, segmentation style={black!22, solid},
}
\makeatletter\def\zifour@opt{1}\def\zifour@altopt{3}\makeatother %
\newcommand{\pmono}{\fontencoding{T1}\fontfamily{zi4}\selectfont}
\newcommand{\id}[1]{{\pmono #1}}
\newcommand{\gap}{\par\vspace{5pt}}
\newcommand{\pitem}[1]{\par\noindent\hangindent=0.85em\hangafter=1 -\ #1\par}
\newcommand{\ans}[2][]{\par\noindent\hangindent=1.5em\hangafter=1{\pmono #2}%
  \if\relax\detokenize{#1}\relax\else\hspace{1em}{\color{black!58}#1}\fi\par}
\newcommand{\note}[1]{{\color{black!58}(#1)}}
\newenvironment{slots}{\begin{tabular}{@{}>{\raggedright\arraybackslash}p{0.17\linewidth}>{\raggedright\arraybackslash}p{0.80\linewidth}@{}}}{\end{tabular}}

\clearpage %
\section{MLLM Prompts}
\label{sec:suppl_prompts}

Every MLLM call in \method sends one decision prompt as the system message and the labeled images as the user message (``Image 1'', or ``stack 1, slice 3/16'' in a search stack).
Each call thus asks for exactly one visual judgment.
\cref{tab:suppl_prompts} maps each prompt to the step of the loop at which the harness calls it.
The boxes below give the text of each prompt in the default configuration.
All prompts stay the same for every MLLM and every benchmark dataset.
Text in angle brackets, such as \slot{target}, is a slot that the harness fills at run time.
The lower part of each box lists the values of these slots.
Most prompts share two passages: one describes the layout of the tri-planar panels, and the other gives the annotation standard for the ROI type.
We give their full text once in \ref{prompt:shared}, and the other prompts show only the corresponding slots, \slot{tri-planar panel layout} and \slot{ROI standard}.

\begin{table}[h]
\centering
\footnotesize
\setlength{\tabcolsep}{4pt}
\renewcommand{\arraystretch}{1.15}
\caption{\textbf{MLLM prompts of \method.}
Each prompt is listed with the step of the loop at which the harness calls it.}
\label{tab:suppl_prompts}
\resizebox{\linewidth}{!}{%
\begin{tabular}{@{}l l l l l@{}}
\toprule
\textbf{Prompt} & \textbf{Step of the loop} & \textbf{Described in} & \textbf{Images} & \textbf{Answers} \\
\midrule
\ref{prompt:seed_start} & Choosing the initial prediction & \cref{sec:suppl_agent} & two sets of tri-planes, one per proposal & mask start, point start, neither, unsure \\
\ref{prompt:seed_point} & Retry after the proposal from $P_0$ fails &\cref{sec:suppl_agent} & one set of tri-planes with numbered rings & one ring, none, unsure \\
\ref{prompt:review} & Verification view & \mainref{sec:harness} & one set of tri-planes & accept, error with box, wrong structure, unsure \\
\ref{prompt:verify} & Confirming an error (step~\circled{1}) & \cref{sec:suppl_agent} & one set of tri-planes centered on the box & confirm, withdraw \\
\ref{prompt:correction} & Click selection (step~\circled{3}) & \mainref{sec:harness} & one set of tri-planes with numbered rings & one ring, none, unsure \\
\ref{prompt:trial_seed}, \ref{prompt:trial_click} & Committing a proposal (step~\circled{4}) & \cref{sec:suppl_agent} & one set of tri-planes of the proposal & on target, wrong structure, unsure \\
\ref{prompt:repair} & Committing a large correction & \cref{sec:suppl_agent} & two or four sets of tri-planes, before and after & improved, not improved, unsure \\
\midrule
\ref{prompt:search} & ROI search &\cref{sec:suppl_agent} & 16 axial slices, one image each & one point, unsure \\
\ref{prompt:search_seed} & First click after the ROI search & \cref{sec:suppl_agent} & one set of tri-planes & one point, unsure \\
\bottomrule
\end{tabular}
}
\end{table}

\promptlabel{prompt:shared}
\begin{promptbox}{\theprompt\quad Shared blocks}{Inserted where a prompt shows the slot \emph{tri-planar panel layout} or \emph{ROI standard}. The ROI standard is fixed text selected by the ROI type (organ, bone, vessel, tumor, muscle, node, or a generic text for an unknown type); three of the seven are shown.}
\slot{tri-planar panel layout}

Image 1 is a tri-view: three panels side by side, AX (axial), SAG (sagittal) and COR (coronal). Each panel is one 2D slice of the scan, and the three slices cross at the panel centres. Letters at the panel edges mark the patient's right, left, front, back, head and foot (R, L, A, P, S, I), so you can tell left from right. Panels without letters have unknown orientation: judge structures by anatomy, not by side.
\gap
\slot{ROI standard} \note{organ}

The mask should cover the whole organ seen here, including the vessels and hilum inside its outline. It must not include neighbouring organs, the fat between them, or vessels outside the organ. An error is a missing pole, lobe or segment, or red crossing the fat plane into a neighbour.
\gap
\slot{ROI standard} \note{vessel}

The mask should follow the lumen of the named vessel along its course here. It must not follow a parallel vessel or the organ it enters. Thin side branches do not count. An error is a gap longer than the vessel is wide, or red following a different vessel.
\gap
\slot{ROI standard} \note{tumor}

Every lesion seen here is target tissue, as the name defines it (whole tumour, enhancing part, \dots). The mask must not include normal tissue around a lesion. Its edge is often blurred: a few millimetres either way is fine. An error is a lesion with no red, or red reaching well beyond a lesion into normal tissue.
\end{promptbox}

\promptlabel{prompt:seed_start}
\begin{promptbox}{\theprompt\quad Choosing the initial prediction}{Called three times with sampled answers, which decide by majority. Images: the proposal from $P_0$ (Image~1) and the point-based proposal (Image~2), each drawn on the tri-planes of the initial view.}
You are a skilled annotator choosing how a 3D segmentation should start.
\gap
\slot{tri-planar panel layout} \note{with ``Image 1 is'' replaced by ``Each image is''}

Red is a candidate first mask, shown where it crosses each slice. Both images are centred where a separate localizer placed the target; the localizer can be wrong.

Image 1: the mask the segmentation tool made from the localizer's own segmentation of the target (\slot{voxels 1} voxels).

Image 2: the mask the tool grew from one point placed inside the localizer's segmentation (\slot{voxels 2} voxels).
\gap
The target is: \slot{target}. Each panel is \slot{F} mm wide.

\slot{ROI standard}
\gap
Choose the better start: the mask that lies on the target's tissue and covers more of it. It does not have to be complete or tidy: missing parts, a rough outline and small overspill are fixed later. Image 1 keeps the localizer's outline, so it is often irregular or has extensions along the target; those are usually target tissue, and a smoother outline in Image 2 is not by itself a reason to prefer Image 2. Prefer Image 2 when Image 1 clearly crosses into a neighbouring structure or covers only a small part of the target.

Answer \id{no\_valid\_candidate} only when the red in both images sits on a different structure than the target: a rough mask that lies on the target is a valid start.
\gap
Answer with one JSON object:
\ans[start from Image 1]{\{"decision":"use\_localizer\_mask"\}}
\ans[start from Image 2]{\{"decision":"select\_seed\_candidate","candidate\_id":"\slot{point label}"\}}
\ans[both masks sit on a different structure than the target]{\{"decision":"no\_valid\_candidate"\}}
\ans[the images do not show enough to decide]{\{"decision":"uncertain"\}}
\tcblower
\begin{slots}
\slot{target} & The text goal $q$, \eg, ``left kidney in abdomen CT''. \\
\slot{F} & Side length of the view in mm. \\
\slot{voxels 1}, \slot{voxels 2} & Voxel counts of the two proposals. \\
\end{slots}
\end{promptbox}

\promptlabel{prompt:seed_point}
\begin{promptbox}{\theprompt\quad Choosing a point after the proposal from $P_0$ fails}{Called three times with sampled answers, which decide by majority, once the proposal from $P_0$ has been rejected (\cref{sec:suppl_agent}). Image: the tri-planes of the initial view, showing up to five numbered rings, plus a magnified crop when the rings lie close together.}
You are a skilled annotator choosing where a 3D segmentation should start.
\gap
\slot{tri-planar panel layout}
\gap
In one panel, yellow rings labelled \slot{labels} mark candidate points; each point is the centre of its ring.\slot{crop}
\gap
The target is: \slot{target}. Each panel is \slot{F} mm wide. The segmentation tool will grow a 3D mask outward from the one point you choose, so the point must lie on the target's tissue. Choose the ring whose centre is on the target. If several are, prefer one away from the target's edge and away from neighbouring organs and vessels.
\gap
Answer with one JSON object:
\ans[the label you chose]{\{"decision":"select\_seed\_candidate","candidate\_id":"\slot{first label}"\}}
\ans[no ring centre is on the target]{\{"decision":"no\_valid\_candidate"\}}
\ans[the image is too unclear to decide]{\{"decision":"uncertain"\}}
\tcblower
\begin{slots}
\slot{labels} & Labels of the rings, one per candidate point. \\
\slot{crop} & Empty, or, when the rings lie close together: `` The rings are close together, so Image 2 is a magnified crop of that part of Image 1, showing the same rings with the same labels.'' \\
\end{slots}
\end{promptbox}

\promptlabel{prompt:review}
\begin{promptbox}{\theprompt\quad Judging a verification view}{Called once per verification view in $\mathcal{V}(M_t)$, and at most twice more while the answer stays unsure. Image: the tri-planes of that view with $M_t$ as a red overlay; on the view of the uncovered prior, the uncovered part of $P_0$ is outlined in blue.}
You are a skilled annotator checking a segmentation.
\gap
\slot{tri-planar panel layout}
\gap
The red overlay is the current 3D mask of the target, shown where it crosses each slice. This view is centred \slot{placement}. Several such views are checked in turn, and the mask is accepted only when all of them pass.
\gap
The target is: \slot{target}. Each panel is \slot{F} mm wide.\slot{last change}

The target may consist of several pieces (tumours, nodes, a row of vertebrae): every piece in the view is target tissue. A target that names a side or a level (the left kidney, L3) is one piece: its counterpart and the levels above and below are not the target.
\gap
What counts as an error here:
\pitem{\id{fn}: tissue that is certainly part of the target, with no red on it.}
\pitem{\id{fp}: red on tissue that is certainly not the target (another organ, bone, vessel, fat, air), or red crossing the border into a neighbour.}
Not an error: a boundary off by a voxel or two; a slice near the edge of the structure that shows only a thin sliver or a small patch of red, which is how a 3D mask looks there; tissue you cannot confidently assign to the target.

\slot{ROI standard}

If you see a clear error, mark it with a coarse rectangle in the panel where it is clearest; the exact point is chosen from candidates inside the rectangle in the next step, and the claim is checked on a view centred there before anything changes. Judge only what these slices show. When in doubt, accept: a missed sliver costs less than a wrong correction.
\gap
Answer with one JSON object:
\ans[no clear error here]{\{"decision":"accept\_local","confidence":"medium|high"\}}
\ans{\{"decision":"localize\_error","error\_type":"fn|fp","panel":"axial|sagittal|coronal", "region":\{"x0":0,"y0":0,"x1":1000,"y1":1000\}\}}
\ans[the red as a whole follows another structure that you can name, and the target has no red]{\{"decision":"wrong\_structure"\}}
\ans[these slices do not show enough to decide]{\{"decision":"uncertain"\}}
\tcblower
\begin{slots}
\slot{placement} & By verification view: ``at the centre of the first view'' (initial view); ``at the centre of the mask'' (centroid); ``where the last change to the mask was made'', or ``where the last correction added tissue'' / ``removed tissue'' / ``changed the mask'' (last change); ``on the surface of the mask''; ``on a separate piece of the mask''; ``on the mask, one zoom level out, to judge the structure as a whole'' (overview). On the view of the uncovered prior: ``on the largest region that a separate localizer assigned to the target but the red mask does not cover, outlined in blue. Decide whether the blue region is target tissue the mask misses (then mark it as fn there) or another structure (then accept)''. \\[2pt]
\slot{last change} & Empty, or one sentence on the last change: `` The mask was just created from a separate localizer's segmentation (\slot{n} voxels).''; `` The mask was just created from the point you chose (\slot{n} voxels).''; `` The last correction added a positive (negative) point; the mask gained \slot{a} voxels and lost \slot{b}, to \slot{n} voxels.'' \\
\end{slots}
\end{promptbox}

\promptlabel{prompt:verify}
\begin{promptbox}{\theprompt\quad Confirming an error}{Called for each box that the overlay does not contradict (\cref{sec:suppl_agent}). Image: the tri-planes of a view centered on the box, with a crosshair at its center.}
You are checking a claim about a segmentation.
\gap
\slot{tri-planar panel layout}
\gap
The red overlay is the current 3D mask of the target, shown where it crosses each slice. The three slices meet at the crosshair, which sits on tissue that an earlier check marked as \slot{claim}.
\gap
The target is: \slot{target}. Each panel is \slot{F} mm wide.

\slot{ROI standard}
\gap
Measured on the marked slice: \slot{measurements}
\gap
First name the tissue at the crosshair, then decide:
\pitem{\id{confirm}: the claim is right. \slot{confirm rule}}
\pitem{\id{withdraw}: the claim is wrong, or these slices do not let you tell. A wrong correction costs more than a missed sliver, so withdraw when in doubt.}
\gap
Answer with one JSON object:
\ans{\{"decision":"confirm","tissue":"<the tissue at the crosshair>"\}}
\ans{\{"decision":"withdraw","tissue":"<the tissue at the crosshair>"\}}
\tcblower
\begin{slots}
\slot{claim} & fn: ``target tissue that the mask missed''; fp: ``red lying on tissue that is not the target''. \\[2pt]
\slot{confirm rule} & fn: ``The tissue at the crosshair is certainly part of the target and has no red.''; fp: ``The red at the crosshair certainly lies on something other than the target.'' \\[2pt]
\slot{measurements} & Up to six sentences computed in code: ``The box is \slot{w}$\times$\slot{h} mm and \slot{p} of it is red.''; fn only: ``Its nearest red voxel is \slot{d} mm away.''; ``The tissue in the box outside the red (fp: The red in the box) is \slot{z} standard deviations brighter (darker) than the mask's own tissue; \slot{p} of it falls within the mask's usual intensity range.''; fp only: ``The red in the box belongs to a piece holding \slot{p} of the mask.''; ``On this slice the mask's cross-section is \slot{p} of its largest cross-section.''; fn only, for organs, muscles and vessels: ``A separate localizer's own mask of the target has \slot{p} of its voxels outside the red mask; the marked region lies in (does not lie in) that uncovered part.'' \\
\end{slots}
\end{promptbox}

\promptlabel{prompt:correction}
\begin{promptbox}{\theprompt\quad Selecting a click}{Called after an error is confirmed. Image: the tri-planes, showing up to five numbered rings inside the box, yellow for a missing part and cyan for an extra part, plus a magnified crop when the rings lie close together.}
You are a skilled annotator choosing a correction point.
\gap
\slot{tri-planar panel layout}

Red is the current mask.
\gap
In one panel, \slot{colour} rings labelled \slot{labels} mark candidate points inside the rectangle you drew; each point is the centre of its ring.\slot{crop}
\gap
The target is: \slot{target}. Each panel is \slot{F} mm wide. You marked \slot{marked} there.
\gap
\slot{rule}
\gap
Answer with one JSON object:
\ans[the label you chose]{\{"decision":"select\_candidate","candidate\_id":"\slot{first label}"\}}
\ans[no ring centre fits]{\{"decision":"no\_valid\_candidate"\}}
\ans[the image is too unclear to decide]{\{"decision":"uncertain"\}}
\tcblower
\begin{slots}
\slot{colour}, \slot{marked} & fn: ``yellow'', ``missed target tissue (fn)''; fp: ``cyan'', ``red on non-target tissue (fp)''. \\[2pt]
\slot{rule} & fn: ``The chosen point will be a positive click: the segmentation tool grows the mask to include the tissue around it. Choose a ring whose centre is on target tissue that is not red. If several are, prefer one well inside the missed region.'' fp: ``The chosen point will be a negative click: the segmentation tool removes the red around it. Choose a ring whose centre is on red that covers non-target tissue. If several are, prefer one well inside the wrong red, away from the target.'' \\[2pt]
\slot{crop} & As in \ref{prompt:seed_point}. \\
\end{slots}
\end{promptbox}

\promptlabel{prompt:trial_seed}
\begin{promptbox}{\theprompt\quad Confirming the initial prediction}{Called on the chosen proposal after the basic checks pass. Image: the tri-planes of the initial view with the proposal as a red overlay.}
You are a skilled annotator checking a first segmentation attempt.
\gap
\slot{tri-planar panel layout}
\gap
\slot{source}
\gap
The target is: \slot{target}. Each panel is \slot{F} mm wide.

\slot{ROI standard}
\gap
Decide whether this mask is a usable start. It does not have to be complete: missing parts and small overspill are fixed later.
\pitem{\id{accept\_trial}: the red lies on the target's tissue and covers a solid part of it.}
\pitem{\id{wrong\_structure}: the red as a whole follows another structure that you can name (a neighbouring organ, a vessel, the counterpart), and the target itself has no red.}
\pitem{\id{uncertain}: you cannot tell, or the red is only a tiny speck of the target.}
\gap
Answer with one JSON object:
\ans{\{"decision":"accept\_trial","confidence":"medium|high"\}}
\ans{\{"decision":"wrong\_structure"\}}
\ans{\{"decision":"uncertain"\}}
\tcblower
\begin{slots}
\slot{source} & Proposal from $P_0$: ``The red overlay is a 3D mask that the segmentation tool made from a separate localizer's segmentation of the target, shown where it crosses each slice. The panels are centred where the localizer placed the target; the localizer can be wrong.'' Point-based proposal: ``The red overlay is a 3D mask that the segmentation tool grew from the point you chose, shown where it crosses each slice. The panels are centred where they were before: your point lies on the slice of the panel its ring was shown in, and the other two slices may cut the mask off-centre or miss it.'' \\
\end{slots}
\end{promptbox}

\promptlabel{prompt:trial_click}
\begin{promptbox}{\theprompt\quad Confirming a prediction after a click}{Called on every proposal from a click after the basic checks pass. Image: the tri-planes of the proposal as a red overlay.}
You are a skilled annotator checking a correction to a segmentation.
\gap
\slot{tri-planar panel layout}
\gap
The red overlay is the 3D mask after your correction point was added as a \slot{purpose}, shown where it crosses each slice. The mask changed from \slot{n before} to \slot{n after} voxels.
\gap
The target is: \slot{target}. Each panel is \slot{F} mm wide.

\slot{ROI standard}
\gap
Decide only whether the mask still follows the target. It may still be incomplete or spill a little; that is checked separately.
\pitem{\id{accept\_trial}: the red still lies on the target's tissue.}
\pitem{\id{wrong\_structure}: the red as a whole now follows another structure that you can name, and the target itself has no red.}
\pitem{\id{uncertain}: you cannot tell.}
\gap
Answer with one JSON object:
\ans{\{"decision":"accept\_trial","confidence":"medium|high"\}}
\ans{\{"decision":"wrong\_structure"\}}
\ans{\{"decision":"uncertain"\}}
\tcblower
\begin{slots}
\slot{purpose} & Positive click: ``positive point to fill a missed region''; negative click: ``negative point to remove overspill''. \\
\end{slots}
\end{promptbox}

\promptlabel{prompt:repair}
\begin{promptbox}{\theprompt\quad Comparing a prediction before and after a click}{Called after \ref{prompt:trial_click} accepts a proposal from a click that changes the prediction strongly, \eg, by adding a distant component. Images: the tri-planes of the same view before and after the click (Images~1 and 2) and, when available, a second before/after pair centered where the prediction changed most (Images~3 and 4).}
You are a skilled annotator judging whether a correction improved a segmentation.
\gap
\slot{tri-planar panel layout} \note{with ``Image 1 is'' replaced by ``Each image is''}

Red is the mask.

Image 1 and Image 2 show the same view before and after the correction: same centre, same slices, same field of view.\slot{second pair}
\gap
The target is: \slot{target}. Each panel is \slot{F} mm wide. You added a \slot{point}.

The mask gained \slot{a} voxels and lost \slot{b}, from \slot{n before} to \slot{n after} voxels; \slot{p} of it changed, across \slot{s} mm.\slot{retained} The change is checked before it is kept.

\slot{ROI standard}
\gap
\pitem{\id{accept\_repair}: the error you marked is fixed, and the change stays on the target: added red is target tissue, removed red was not.}
\pitem{\id{reject\_repair}: the error is not fixed, or the change spills onto other tissue or removes target tissue.}
\pitem{\id{uncertain}: the images do not show enough to decide.}
\gap
Answer with one JSON object:
\ans{\{"decision":"accept\_repair","confidence":"medium|high"\}}
\ans{\{"decision":"reject\_repair"\}}
\ans{\{"decision":"uncertain"\}}
\tcblower
\begin{slots}
\slot{second pair} & Empty, or `` Image 3 and Image 4 are a second before/after pair, centred where the correction changed the mask most.'' \\[2pt]
\slot{point} & Positive click: ``positive point to fill missed target tissue''; negative click: ``negative point to remove red on non-target tissue''. \\[2pt]
\slot{retained} & Empty, or `` \slot{p} of the previous mask is retained.'' \\[2pt]
\slot{s} & Largest extent of the changed region along the three axes. \\
\end{slots}
\end{promptbox}

\promptlabel{prompt:search}
\begin{promptbox}{\theprompt\quad Searching for the target without a localization prior}{Called when $\floc$ finds nothing, or when no initial prediction is committed from its prior. Round 1 shows 16 axial slices through the whole scan; round 2 shows 16 axial slices through a 96\,mm window around the round-1 point. Each image is preceded by its label, \eg, ``stack 1, slice 3/16''.}
You are a skilled annotator finding where a target lies in a 3D scan.
\gap
A view is a stack of 16 axial slices, one image per slice. The text before each image names its stack and slice (``stack 1, slice 3/16'') and each panel's corner reads AX k/16. The slices are in order from the foot end to the head end when the scan's orientation is known (else in slice order). Letters at the panel edges mark the patient's right, left, front and back (R, L, A, P), so you can tell left from right. Panels without letters have unknown orientation: judge structures by anatomy, not by side. There is no sagittal or coronal panel: name a panel by its slice, \id{"ax<k>"} (slice 3 is \id{"ax3"}).

\slot{round}
\gap
The target is: \slot{target}. An automatic localizer could not find it, so find it from the anatomy. The segmentation tool will grow a 3D mask outward from the point you give, so the point must lie on the target's tissue, well inside it, in the slice where it is clearest. Each slice is \slot{F} mm wide.

\slot{earlier attempts}
\gap
Answer with one JSON object:
\ans[the panel and the point's position in it: x 0..1000 from left to right, y 0..1000 from top to bottom]{\{"decision":"place\_point","panel":"ax<k>","x":500,"y":500\}}
\ans[the target is not visible in any slice]{\{"decision":"uncertain"\}}
\tcblower
\begin{slots}
\slot{round} & Round 1: ``The 16 slices are spaced evenly through the whole scan, from one end to the other, so the target may lie between two of them: pick the slice where it is clearest, or the slice nearest to it.'' Round 2: ``The 16 slices cover a 96 mm window around the point you chose on the whole scan, at finer spacing, so you can place the point precisely.'' \\[2pt]
\slot{F} & Round 1: the width of the scan; round 2: 96. \\[2pt]
\slot{earlier attempts} & Empty, or, when a point fell outside the scan and is asked again: ``Earlier attempts on this image, which you must not repeat:'' followed by one line per attempt, ``- the point at x \slot{x}, y \slot{y} in the \slot{panel} panel lies outside the scan''. \\
\end{slots}
\end{promptbox}

\promptlabel{prompt:search_seed}
\begin{promptbox}{\theprompt\quad Placing the first click after the search}{Called on a 128\,mm view centered at the point found by \ref{prompt:search}. The MLLM places the point itself, since no localization prior is left to sample candidate points from; the resulting prediction is then confirmed with \ref{prompt:trial_seed}.}
You are a skilled annotator choosing where a 3D segmentation should start.
\gap
\slot{tri-planar panel layout}
\gap
The panels are centred where your search through the whole scan placed the target. It can be wrong, so find the target from the anatomy.
\gap
The target is: \slot{target}. Each panel is \slot{F} mm wide. The segmentation tool will grow a 3D mask outward from the one point you give, so the point must lie on the target's tissue. Prefer a point well inside the target, away from its edge and away from neighbouring organs and vessels. Place it in the panel where the target is clearest.

\slot{earlier attempts}
\gap
Answer with one JSON object:
\ans[the panel and the point's position in it: x 0..1000 from left to right, y 0..1000 from top to bottom]{\{"decision":"place\_point","panel":"axial|sagittal|coronal","x":500,"y":500\}}
\ans[the target is not in the image or it is too unclear to decide]{\{"decision":"uncertain"\}}
\tcblower
\begin{slots}
\slot{earlier attempts} & Empty, or ``Earlier attempts on this image, which you must not repeat:'' followed by one line per earlier point, ``- the point at x \slot{x}, y \slot{y} in the \slot{panel} panel \slot{result}'', where \slot{result} is one of ``grew a mask on a different structure than the target'', ``grew a mask that could not be confirmed as the target'', ``produced no mask'', ``grew a mask over far more than the target'', ``lies outside the scan'', ``lies within 5 mm of a point that grew a different structure'', or ``was already tried''. \\
\end{slots}
\end{promptbox}

\fi
\end{document}